%% file: main.tex
\documentclass[preprint,12pt, 3p]{elsarticle}
\usepackage[utf8]{inputenc}
\usepackage[T1]{fontenc} 

\usepackage{graphicx}
\usepackage[
    colorlinks=true,
    linkcolor=blue,
    filecolor=magenta,      
    urlcolor=cyan,
]{hyperref} 

\usepackage{float} 
\usepackage{xargs} 

\usepackage{dingbat}
\usepackage{color,soul}

\usepackage{longtable}
\usepackage{lipsum}  
\usepackage{adjustbox}

\usepackage{lineno}
\usepackage{amssymb}
\usepackage[figuresright]{rotating}
\usepackage{lipsum}
\usepackage{mathtools}
\usepackage{cuted}

\usepackage[pdftex,dvipsnames]{xcolor}  

\colorlet{soultransparent}{red!0}
\sethlcolor{soultransparent}

\usepackage[colorinlistoftodos,prependcaption,textsize=small]{todonotes}
\newcommandx{\unsure}[2][1=]{\todo[linecolor=red,backgroundcolor=red!25,bordercolor=red,#1]{#2}}
\newcommandx{\change}[2][1=]{\todo[linecolor=blue,backgroundcolor=blue!25,bordercolor=blue,#1]{#2}}
\newcommandx{\info}[2][1=]{\todo[linecolor=OliveGreen,backgroundcolor=OliveGreen!25,bordercolor=OliveGreen,#1]{#2}}
\newcommandx{\improvement}[2][1=]{\todo[linecolor=Plum,backgroundcolor=Plum!25,bordercolor=Plum,#1]{#2}}
\newcommandx{\thiswillnotshow}[2][1=]{\todo[disable,#1]{#2}}

\usepackage{natbib}

\usepackage[tight,footnotesize]{subfigure}

\DeclareMathOperator*{\softmax}{soft\!\max}

\usepackage[automake]{glossaries}

\usepackage{glossary-mcols}
\setglossarystyle{mcolindex}
\makeglossaries

\input{glossaries.tex}

\begin{document}

\begin{frontmatter}


\title{Financial Time Series Forecasting with Deep Learning : A Systematic Literature Review: 2005-2019}

\author[addr]{Omer Berat Sezer}
\author[addr]{M. Ugur Gudelek}
\author[addr]{Ahmet Murat Ozbayoglu}

\address[addr]{Department of  Computer Engineering, TOBB University of Economics and Technology, Ankara, Turkey}

\begin{abstract}
\input{0_abstract.tex}
\end{abstract}

\begin{keyword}
deep learning \sep finance \sep computational intelligence \sep machine learning \sep time series forecasting \sep CNN \sep LSTM \sep RNN 
\end{keyword}

\end{frontmatter}

\input{1_introduction}

\input{2_ML_in_Finance}

\input{3_DL}

\input{4_Financial_Application}

\input{6_Snapshot_of_Field}

\input{7_Discussion}

\input{8_Conclusion_Futurework}

\renewcommand{\glsgroupskip}{}
\printglossaries


\clearpage

\bibliographystyle{unsrtnat_wo_doi}
\bibliography{bibdatabase}

\end{document}

%% file: glossaries.tex
\newacronym{ml}{ML}{Machine Learning}
\newacronym{dl}{DL}{Deep Learning}
\newacronym{ai}{AI}{Artificial Intelligence}
\newacronym[plural=CNNs]{cnn}{CNN}{Convolutional Neural Network}
\newacronym[plural=DBNs]{dbn}{DBN}{Deep Belief Network}
\newacronym{lstm}{LSTM}{Long-Short Term Memory}
\newacronym{crps}{CRPS}{Continuous Ranked Probability Score}
\newacronym[plural=RNNs]{rnn}{RNN}{Recurrent Neural Network}
\newacronym[plural=ANNs]{ann}{ANN}{Artificial Neural Network}
\newacronym[plural=DNNs]{dnn}{DNN}{Deep Neural Network}
\newacronym{mlp}{MLP}{Multilayer Perceptron}
\newacronym[plural=DMLPs]{dmlp}{DMLP}{Deep Multilayer Perceptron}
\newacronym[plural=AEs]{ae}{AE}{Autoencoder}
\newacronym{hci}{HCI}{Human-Computer Interaction}
\newacronym[plural=GPs]{gp}{GP}{Genetic Programming}
\newacronym[plural=GAs]{ga}{GA}{Genetic Algorithm}
\newacronym[plural=ECs]{ec}{EC}{Evolutionary Computation}
\newacronym[plural=MOEAs]{moea}{MOEA}{Multiobjective Evolutionary Algorithm}
\newacronym{relu}{ReLU}{Rectified Linear Unit}
\newacronym{sgd}{SGD}{Stochastic Gradient Descent}
\newacronym{adagrad}{AdaGrad}{Adaptive Gradient Algorithm}
\newacronym{rmsprop}{RMSProp}{Root Mean Square Propagation}
\newacronym{adam}{ADAM}{Adaptive Moment Estimation}
\newacronym{cd}{CD}{Contrastive Divergence}
\newacronym{kldivergence}{KL-Divergence}{Kullback Leibler Divergence}
\newacronym{ais}{AIS}{Annealed Importance Sampling}
\newacronym{rl}{RL}{Reinforcement learning}
\newacronym{dp}{DP}{Dynamic Programming}
\newacronym{mc}{MC}{Monte Carlo}
\newacronym{td}{TD}{Temporal Difference}
\newacronym{dwnn}{DWNN}{Deep and Wide Neural Network}
\newacronym{srnn}{SRNN}{Stacked Recurrent Neural Network}
\newacronym{arma}{ARMA}{Autoregressive Moving Average}
\newacronym{elm}{ELM}{Extreme Learning Machine}
\newacronym{gbt}{GBT}{Gradient Boosted Trees}
\newacronym{gan-fd}{GAN-FD}{GAN for minimizing Forecast error loss and Direction prediction loss}
\newacronym{rcnn}{RCNN}{Recurrent CNN}
\newacronym{ar}{AR}{Autoregressive}
\newacronym{rf}{RF}{Random Forest}
\newacronym{sp500}{S\&P500}{Standard’s \& Poor’s 500 Index}
\newacronym{nifty}{NIFTY}{National Stock Exchange of India}
\newacronym{sse}{SSE}{Shanghai Stock Exchange}
\newacronym{hsi}{HSI}{Hong Kong Hang Seng Index}
\newacronym{taiex}{TAIEX}{Taiwan Capitalization Weighted Stock Index}
\newacronym{dow30}{DOW30}{Dow Jones Industrial Average 30}
\newacronym{kospi}{KOSPI}{The Korea Composite Stock Price Index}
\newacronym{vxn}{VXN}{NASDAQ100 Volatility Index}
\newacronym{bovespa}{Bovespa}{Brazilian Stock Exchange}
\newacronym{omx}{OMX}{Stockholm Stock Exchange}
\newacronym{egarch}{EGARCH}{Exponential GARCH}
\newacronym{bi-lstm}{Bi-LSTM}{Bidirectional LSTM}
\newacronym{har}{HAR}{Heterogeneous Autoregressive Process}
\newacronym{gasvr}{GASVR}{\acrshort{ga} with a \acrshort{svr}}
\newacronym{flann}{FLANN}{Functional Link Neural network}
\newacronym{ea}{EA}{Evolutionary Algorithm}
\newacronym{lda}{LDA}{Latent Dirichlet Allocation}
\newacronym{cdbn}{CDBN}{Continuous-valued Deep Belief Networks}
\newacronym{crbm}{CRBM}{Continuous Restricted Boltzman machine}
\newacronym{pnn}{PNN}{Probabilistic Neural Network}
\newacronym{xgboost}{XGBoost}{eXtreme Gradient Boosting}
\newacronym{hmm}{HMM}{Hidden Markov Model}
\newacronym{tgru}{TGRU}{Two-stream GRU}
\newacronym{har-gasvr}{HAR-GASVR}{\acrshort{har} with a \acrshort{gasvr}}
\newacronym{rmdn}{RMDN}{Recurrent Mixture Density Network}
\newacronym{rmdn-garch}{RMDN-GARCH}{\acrshort{rmdn} with a \acrshort{garch}}
\newacronym{rw}{RW}{Random Walk}
\newacronym{mrs}{MRS}{Markov Regime Switching}
\newacronym{williamr}{William\%R}{Williams Percent Range}
\newacronym{pposc}{PPOSC}{Percentage Price Oscillator}
\newacronym{gml}{GML}{Generalized Linear Model}
\newacronym{dfnn}{DFNN}{Deep Feedforward Neural Network}
\newacronym{ffnn}{FFNN}{Feedforward Neural Network}
\newacronym{nn}{NN}{Neural Network}
\newacronym{aar}{AAR}{Annual Rate of Return}
\newacronym{ac}{AC}{Autocorrelation}
\newacronym{amex}{AMEX}{American Stock Exchange}
\newacronym{areturn}{AR}{Active Return}
\newacronym{arch}{ARCH}{Autoregressive Conditional Heteroscedasticity}
\newacronym{arima}{ARIMA}{Autoregressive Integrated Moving Average}
\newacronym{atr}{ATR}{Average True Range}
\newacronym{auc}{AUC}{Area Under the Curve}
\newacronym{auroc}{AUROC}{Area Under the Receiver Operating Characteristics}
\newacronym{ba}{BA}{Balanced Accuracy}
\newacronym{belm}{BELM}{Basic Extreme Learning Machine}
\newacronym{betc}{BETC}{Break Even Transaction Cost}
\newacronym{bist}{BIST}{Istanbul Stock Exchange Index}
\newacronym{bi-gru}{Bi-GRU}{Bidirectional Gated Recurrent Unit}
\newacronym{boll}{BOLL}{Bollinger Band}
\newacronym{bp}{BP}{Back Propagation}
\newacronym{bptt}{BPTT}{Back Propagation Through Time}
\newacronym{bse}{BSE}{Bombay Stock Exchange}
\newacronym{cagr}{CAGR}{Compound Annual Growth Rate}
\newacronym{car}{CAR}{Cumulative Abnormal Return}
\newacronym{cart}{CART}{Classification and Regression Trees}
\newacronym{cc}{CC}{Correlation Coefficient}
\newacronym{cci}{CCI}{Commodity Channel Index}
\newacronym{cdax}{CDAX}{German Stock Market Index Calculated by Deutsche Börse}
\newacronym{cdbn-fg}{CDBN-FG}{Fuzzy Granulation with Continuous-valued Deep Belief Networks}
\newacronym{cds}{CDS}{Credit Default Swaps}
\newacronym{cew}{CEW}{Emerging Markets Currency Index}
\newacronym{cgan}{CGAN}{Conditional \acrshort{gan}}
\newacronym{cme}{CME}{Chicago Mercantile Exchange}
\newacronym{coefficient}{coefficient}{}
\newacronym{crsp}{CRSP}{Center for Research in Security Prices}
\newacronym{cse}{CSE}{Colombo Stock Exchange}
\newacronym{csi}{CSI}{China Securities Index}
\newacronym{cwn}{cWN}{Conditional Wavenet}
\newacronym{da}{DA}{Direction Accuracy}
\newacronym{dax}{DAX}{The Deutscher Aktienindex}
\newacronym{dcnn}{DCNN}{Deep Convolutional Neural Network}
\newacronym{ddpg}{DDPG}{Deep Deterministic Policy Gradient}
\newacronym{de}{DE}{Differential Evolution} 
\newacronym{deep-fasp}{Deep-FASP}{The Financial Aspect and Sentiment Prediction task with Deep neural networks}
\newacronym{deepcnl}{DeepCNL}{Deep Co-investment Network Learning}
\newacronym{dffn}{DFFN}{Deep Feed Forward Network}
\newacronym{dgm}{DGM}{Deep Neural Generative Model}
\newacronym{djia}{DJIA}{Dow Jones Industrial Average}
\newacronym{dlr}{DLR}{Deep Learning Representation}
\newacronym{dmi}{DMI}{Directional Movement Index}
\newacronym{dof}{DOF}{Degrees of Freedom}
\newacronym{dpa}{DPA}{Direction Prediction Accuracy}
\newacronym{dql}{DQL}{Deep Q-Learning}
\newacronym{drl}{DRL}{Deep Reinforcement Learning}
\newacronym{drse}{DRSE}{Deep Random Subspace Ensembles}
\newacronym{dtw}{DTW}{Dynamic Time Warping}
\newacronym{ema}{EMA}{Exponential Moving Average}
\newacronym{emd2fnn}{EMD2FNN}{Empirical Mode Decomposition and Factorization Machine based Neural Network}
\newacronym{etf}{ETF}{Exchange-Traded Fund}
\newacronym{far}{FAR}{False Acceptance Rate}
\newacronym{fddr}{FDDR}{Fuzzy Deep Direct Reinforcement Learning}
\newacronym{fe-qar}{FE-QAR}{Fixed Effects Quantile VAR}
\newacronym{fiqa}{FiQA}{Financial Opinion Mining and Question Answering Challange}
\newacronym{fn}{FN}{False Negative}
\newacronym{fnn}{FNN}{Fully Connected Neural Network}
\newacronym{fcnn}{FCNN}{Fully Connected Neural Network}
\newacronym{fp}{FP}{False Positive}
\newacronym{fpe}{FPE}{Akaike’s Minimum Final Prediction Error}
\newacronym{fpga}{FPGA}{Field Programmable Gate Array}
\newacronym{frr}{FRR}{False Rejection Rate}
\newacronym{ftse}{FTSE}{London Financial Times Stock Exchange Index}
\newacronym{g-mean}{G-mean}{Geometric Mean}
\newacronym{gaf}{GAF}{Gramian Angular Field}
\newacronym{gan}{GAN}{Generative Adversarial Network}
\newacronym{garch}{GARCH}{Generalised Auto-Regressive Conditional Heteroscedasticity}
\newacronym{gbdt}{GBDT}{Gradient-Boosted-DecisionTrees}
\newacronym{glm}{GLM}{Generalized Linear Model}
\newacronym{gpu}{GPU}{Graphic Processing Unit}
\newacronym{gru}{GRU}{Gated-Recurrent Unit}
\newacronym{gspc}{GSPC}{S\&P500 Commodity Price Index}
\newacronym{han}{HAN}{Hybrid Attention Network}
\newacronym{hft}{HFT}{High Frequency Trading}
\newacronym{hit}{HIT}{Hit Rate}
\newacronym{hmrpso}{HMRPSO}{Modified Version of PSO}
\newacronym{hs}{HS}{China Shanghai Shenzhen Stock Index}
\newacronym{ibb}{IBB}{iShares Nasdaq Biotechnology ETF}
\newacronym{ic}{IC}{Information Coeffiencient}
\newacronym{ir}{IR}{Information Ratio}
\newacronym{ise100}{ISE100}{Istanbul Stock Exchange Index}
\newacronym{ixic}{IXIC}{NASDAQ Composite Index}
\newacronym{kelm}{KELM}{Kernel Extreme Learning Machine}
\newacronym{ks}{KS}{Kolmogorov–Smirnov}
\newacronym{lar}{LAR}{Linear Auto-regression Predictor}
\newacronym{lfm}{LFM}{Lookahead Factor Models}
\newacronym{lob}{LOB}{Limit Order Book Data}
\newacronym{lrnfis}{LRNFIS}{Locally Recurrent Neuro-fuzzy Information System}
\newacronym{ma}{MA}{Moving Average}
\newacronym{macd}{MACD}{Moving Average Convergence and Divergence}
\newacronym{mad}{MAD}{Mean Absolute Deviation}
\newacronym{madr}{MADR}{Moving Average Deviation Rate}
\newacronym{mae}{MAE}{Mean Absolute Error}
\newacronym{mam}{MAM}{Moving Average Mapping}
\newacronym{map}{MAP}{Maximum Absolute Percentage Error}
\newacronym{mape}{MAPE}{Mean Absolute Percentage Error}
\newacronym{mar}{MAR}{Mean Abnormal Return}
\newacronym{mase}{MASE}{Mean Standard Deviation}
\newacronym{mcc}{MCC}{Matthew Correlation Coefficient}
\newacronym{mda}{MDA}{Multilinear Discriminant Analysis}
\newacronym{mdd}{MDD}{Maximum Drawdown}
\newacronym{mdp}{MDP}{Markov Decision Process}
\newacronym{mfi}{MFI}{Money Flow Index}
\newacronym{mi}{MI}{Mutual Information}
\newacronym{modrl}{MODRL}{Multi-objective Deep Reinforcement Learning}
\newacronym{moe}{MoE}{Mixture of Experts}
\newacronym{mse}{MSE}{Mean Squared Error}
\newacronym{msfe}{MSFE}{Mean Squared Forecast Error}
\newacronym{mspe}{MSPE}{Mean Squared Prediction Error}
\newacronym{mtm}{MTM}{Momentum}
\newacronym{narmax}{NARMAX}{Nonlinear Autoregressive Moving Average model with exogenous inputs}
\newacronym{nasdaq}{NASDAQ}{National Association of Securities Dealers Automated Quotations}
\newacronym{nes}{NES}{Natural Evolution Strategies}
\newacronym{nikkei}{NIKKEI}{Tokyo Nikkei Index}
\newacronym{nlp}{NLP}{Natural Language Processing}
\newacronym{nmae}{NMAE}{Normalized Mean Absolute Error}
\newacronym{nmse}{NMSE}{Normalized Mean Square Error}
\newacronym{nymex}{NYMEX}{New York Mercantile Exchange}
\newacronym{nyse}{NYSE}{New York Stock Exchange}
\newacronym{obv}{OBV}{On Balance Volume}
\newacronym{ochl}{OCHL}{Open,Close,High, Low}
\newacronym{ochlv}{OCHLV}{Open,Close,High, Low, Volume}
\newacronym{pca}{PCA}{Principal Component Analysis}
\newacronym{pcc}{PCC}{Pearson’s Correlation Coefficient}
\newacronym{pcd}{PCD}{Percentage of Correct Direction}
\newacronym{plr}{PLR}{Piecewise Linear Representation}
\newacronym{pocid}{POCID}{Percentage of Change in Direction}
\newacronym{ppo}{PPO}{Proximal Policy Optimization}
\newacronym{profit}{PROFIT}{Average Annual Profit of the Model}
\newacronym{psn}{PSN}{Psi-Sigma Network}
\newacronym{pso}{PSO}{Particle Swarm Optimization}
\newacronym{r-sq}{R$^2$}{Squared correlation, Non-linear regression multiple correlation}
\newacronym{r1}{r1}{Correlation coefficient between actual value and prediction value}
\newacronym{r2}{r2}{Correlation coefficient between actual return and prediction return}
\newacronym{ra}{RA}{Rolling Average}
\newacronym{raf}{RAF}{Random Forests}
\newacronym{rbf}{RBF}{Radial Basis Function Neural Network}
\newacronym{rbm}{RBM}{Restricted Boltzmann Machine}
\newacronym{rceflann}{RCEFLANN}{Recurrent Computationally Efficient Functional Link Neural Network}
\newacronym{rci}{RCI}{Rank Correlation Index}
\newacronym{return}{RETURN}{Average Annual Returns of the Model}
\newacronym{rmse}{RMSE}{Root Mean Square Error}
\newacronym{rmsre}{RMSRE}{Root Mean Square Relative Error}
\newacronym{roa}{ROA}{Return on Assets}
\newacronym{roc}{ROC}{Price of Change}
\newacronym{rse}{RSE}{Relative Squared Error}
\newacronym{rsi}{RSI}{Relative Strength Index}
\newacronym{sae}{SAE}{Stacked Autoencoder}
\newacronym{sar}{SAR}{Parabolic Stop and Reverse}
\newacronym{sci}{SCI}{SSE Composite Index}
\newacronym{sd}{SD}{Standard Deviation (also referred as the Greek letter r)}
\newacronym{sdae}{SDAE}{Stacked Denoising Autoencoders}
\newacronym{sfm}{SFM}{State Frequency Memory}
\newacronym{si}{SI}{Stochastic Index}
\newacronym{slp}{SLP}{Single Layer Perceptron}
\newacronym{smape}{SMAPE}{Symmetric Mean Absolute Percentage Error}
\newacronym{som}{SOM}{Self-Organising Map}
\newacronym{sr}{SR}{Sharpe-ratio}
\newacronym{svd}{SVD}{Singular Value Decomposition}
\newacronym{svm}{SVM}{Support Vector Machine}
\newacronym{svr}{SVR}{Support Vector Regressor}
\newacronym{szse}{SZSE}{Shenzhen Stock Exchange Composite Index}
\newacronym{talib}{TALIB}{Technical Analysis Library Package}
\newacronym{tar}{TAR}{Threshold Autoregressive}
\newacronym{vec}{VEC}{Vector Error Correction model}
\newacronym{rhe}{RHE}{Recurrent Hybrid Elman}
\newacronym{tdnn}{TDNN}{Timedelay Neural Network}
\newacronym{theil-u}{THEIL-U}{Theil's inequality coefficient}
\newacronym{tn}{TN}{True Negative}
\newacronym{tp}{TP}{True Positive}
\newacronym{tr}{TR}{Total Return}
\newacronym{tse}{TSE}{Tokyo Stock Exchange}
\newacronym{tunindex}{TUNINDEX}{Tunisian Stock Market Index}
\newacronym{twse}{TWSE}{Taiwan Stock Exchange}
\newacronym{uwn}{uWN}{Unconditional WaveNet}
\newacronym{var}{VAR}{Vector Auto Regression}
\newacronym{vix}{VIX}{S\&P500 Volatility Index}
\newacronym{vr}{VR}{Variance Reduction}
\newacronym{vwl}{VWL}{WL Kernel-based Method}
\newacronym{vxd}{VXD}{Dow Jones Industrial Average Volatility Index}
\newacronym{wba}{WBA}{Weighted Balanced Accuracy}
\newacronym{weka}{WEKA}{Waikato Environment for Knowledge Analysis}
\newacronym{whr}{WHR}{Weighted Hit Rate}
\newacronym{wmtr}{WMTR}{Weighted Multichannel Time-series Regression}
\newacronym{wpr}{WPR}{William \% R}
\newacronym{wsurt}{WSURT}{Wilcoxon Sum-rank Test}
\newacronym{wt}{WT}{Wavelet Transforms}
\newacronym{true}{TRUE}{True Range of Price Movements}
\newacronym{nse}{NSE}{National Stock Exchange of India}
\newacronym{norm-rmse}{norm-RMSE}{Normalized \acrshort{rmse}}
\newacronym{taq}{TAQ}{Trade and Quote}
\newacronym{hr}{HR}{Hit Rate}
\newacronym{std}{STD}{Standard Deviation}
\newacronym{ise}{ISE}{Istanbul Stock Exchange Index}
\newacronym{gdax}{GDAX}{Global Digital Asset Exchange}
\newacronym{wti}{WTI}{West Texas Intermediate}
\newacronym{mm}{MM}{Markov Model}
\newacronym{hmae}{HMAE}{Heteroscedasticity Adjusted MAE}
\newacronym{hmse}{HMSE}{Heteroscedasticity Adjusted MSE}
\newacronym{spy}{SPY}{SPDR S\&P 500 ETF}
\newacronym{ssec}{SSEC}{Shanghai Stock Exchange Composite}
\newacronym{kse}{KSE}{Korea Stock Exchange}
\newacronym{ibovespa}{IBOVESPA}{Indice Bolsa de Valores de Sao Paulo}
\newacronym{dji}{DJI}{Dow Jones Index}
\newacronym{tfidf}{TF-IDF}{Term Frequency-Inverse Document Frequency}
\newacronym{lr}{LR}{Logistic Regression}
\newacronym{tema}{TEMA}{Triple Exponential Moving Average}
\newacronym{b-h}{B\&H}{Buy and Hold}
\newacronym{wcn}{WCN}{Wavenet Convolution Network} 
\newacronym{fhs}{FHS}{Firefly Harmony Search} 

\newacronym{manualsearch}{MS}{Manual Search}
\newacronym{gridsearch}{GS}{Grid Search}
\newacronym{randomsearch}{RS}{RandomSearch}
\newacronym{smbgo}{SMBGO}{Sequential Model-Based Global Optimization}
\newacronym{gpa}{GPA}{The Gaussian Process Approach} 
\newacronym{tspea}{TSPEA}{Tree-structured Parzen Estimator Approach}
\newacronym{fhso}{FHSO}{Firefly Harmony Search Optimization}

%% file: 0_abstract.tex
Financial time series forecasting is, without a doubt, the top choice of computational intelligence for finance researchers from both academia and financial industry due to its broad implementation areas and substantial impact. \gls{ml} researchers came up with various models and a vast number of studies have been published accordingly. As such, a significant amount of surveys exist covering \gls{ml} for financial time series forecasting studies. Lately, \gls{dl} models started appearing within the field, with results that significantly outperform traditional \gls{ml} counterparts. Even though there is a growing interest in developing models for financial time series forecasting research, there is a lack of review papers that were solely focused on \gls{dl} for finance. Hence, our motivation in this paper is to provide a comprehensive literature review on \gls{dl} studies for financial time series forecasting implementations. We not only categorized the studies according to their intended forecasting implementation areas, such as index, forex, commodity forecasting, but also grouped them based on their \gls{dl} model choices, such as \glspl{cnn}, \glspl{dbn}, \gls{lstm}. We also tried to envision the future for the field by highlighting the possible setbacks and opportunities, so the interested researchers can benefit.  

%% file: 1_introduction.tex
\section{Introduction}
\label{sec:introduction}

The finance industry has always been interested in successful prediction of financial time series data. Numerous studies have been published that were based on \gls{ml} models with relatively better performances compared to classical time series forecasting techniques. Meanwhile, the widespread application of automated electronic trading systems coupled with increasing demand for higher yields keeps forcing the researchers and practitioners to continue working on searching for better models. Hence, new publications and implementations keep pouring into finance and computational intelligence literature.

In the last few years, \gls{dl} started emerging strongly as the best performing predictor class within the \gls{ml} field in various implementation areas. Financial time series forecasting is no exception, as such, an increasing number of prediction models based on various \gls{dl} techniques were introduced in the appropriate conferences and journals in recent years. Despite the existence of the vast amount of survey papers covering financial time series forecasting and trading systems using traditional soft computing techniques, to the best of our knowledge, no reviews have been performed in literature for \gls{dl}. Hence, we decided to work on such a comprehensive study focusing on \gls{dl} implementations of financial time series forecasting. Our motivation is two-fold such that we not only aimed at providing the state-of-the-art snapshot of academic and industry perspectives of the developed \gls{dl} models but also pinpointing the important and distinctive characteristics of each studied model to prevent researchers and practitioners to make unsatisfactory choices during their system development phase. We also wanted to envision where the industry is heading by indicating possible future directions.  

Our fundamental motivation in this paper was to come up with answers for the following research questions:
\begin{itemize}
\item {Which \gls{dl} models are used for financial time series forecasting ?}
\item {How is the performance of \gls{dl} models compared with traditional \gls{ml} counterparts ?}
\item {What is the future direction for \gls{dl} research for financial time series forecasting ?}
\end{itemize}

Our focus was solely on \gls{dl} implementations for financial time series forecasting. For other \gls{dl} based financial applications such as risk assessment, portfolio management, etc., interested readers can check the recent survey paper \cite{Ozbayoglu_2019}. Since we singled out financial time series prediction studies in our survey, we omitted other time series forecasting studies that were not focused on financial data. Meanwhile, we included time-series research papers that had financial use cases or examples even though the papers themselves were not directly intended for financial time series forecasting. Also, we decided to include algorithmic trading papers that were based on financial forecasting, but ignore the ones that did not have a time series forecasting component. 

We reviewed journals and conferences for our survey, however, we also included Masters and PhD theses, book chapters, arXiv papers and noteworthy technical publications that came up in web searches. We decided to only include the articles in the English language. 

During our survey through the papers, we realized that most of the papers using the term ``deep learning" in their description were published in the last 5 years. However, we also encountered some older studies that implemented deep models; such as \glspl{rnn}, Jordan-Elman networks. However, at their time of publication, the term ``deep learning" was not in common usage. So, we decided to also include those papers. 

According to our findings, this will be one of the first comprehensive ``financial time series forecasting" survey papers focusing on  \gls{dl}.  A lot of \gls{ml} reviews for financial time series forecasting exist in the literature, meanwhile, we have not encountered any study on  \gls{dl}.  Hence, we wanted to fill this gap by analyzing the developed models and applications accordingly. We hope, as a result of this paper, the researchers and model developers will have a better idea of how they can implement \gls{dl} models for their studies.

We structured the rest of the paper as follows. Following this brief introduction, in Section~\ref{sec:ml_in_finance}, the existing surveys that are focused on \gls{ml} and soft computing studies for financial time series forecasting are mentioned. In Section~\ref{sec:dl}, we will cover the existing \gls{dl} models that are used, such as \gls{cnn}, \gls{lstm},  \gls{drl}. Section~\ref{sec:financial_application} will focus on the various financial time series forecasting implementation areas using \gls{dl}, namely stock forecasting, index forecasting, trend forecasting, commodity forecasting, volatility forecasting, foreign exchange forecasting, cryptocurrency forecasting.  In each subsection, the problem definition will be given, followed by the particular \gls{dl}  implementations. 
In Section~\ref{sec:snapshot}, overall statistical results about our findings will be presented including histograms about the yearly distribution of different subfields, models, publication types, etc. As a result,  the state-of-the-art snapshot for financial time series forecasting studies will be given through these statistics. At the same time, it will also show the areas that are already mature, compared against promising or new areas that still have room for improvement.
Section~\ref{sec:discussion} will provide discussions about what has been done through academic and industrial achievements and expectations through what might be needed in the future. The section will include highlights about the open areas that need further research. Finally, we will conclude in Section~\ref{sec:conclusions} by summarizing our findings.

%% file: 2_ML_in_Finance.tex
\section{Financial Time Series Forecasting with ML}
\label{sec:ml_in_finance}

Financial time series forecasting and associated applications have been studied extensively for many years. When \gls{ml} started gaining popularity, financial prediction applications based on soft computing models also became available accordingly. Even though our focus is particularly on \gls{dl} implementations of financial time series prediction studies, it will be beneficial to briefly mention about the existing surveys covering \gls{ml}-based financial time series forecasting studies in order to gain historical perspective.

In our study, we did not include any survey papers that were focused on specific financial application areas other than forecasting studies. However, we were faced with some review publications that included not only financial time-series studies but also other financial applications. We decided to include those papers in order to maintain the comprehensiveness of our coverage. 

Examples of these aforementioned publications are provided here. There were published books on stock market forecasting \cite{Aliev_2004}, trading system development \cite{Dymowa_2011}, practical examples of forex and market forecasting applications \cite{Kovalerchuk_2000} using \gls{ml} models like \glspl{ann}, \glspl{ec}, \gls{gp} and Agent-based models \cite{Brabazon_2008}. 

There were also some existing journal and conference surveys. Bahrammirzaee et. al. \cite{Bahrammirzaee_2010} surveyed financial prediction and planning studies along with other financial applications using various \gls{ai} techniques like \gls{ann}, Expert Systems, hybrid models. The authors of \cite{Zhang_2004} also compared \gls{ml} methods in different financial applications including stock market prediction studies. In \cite{Mochn_2007}, soft computing models for the market, forex prediction and trading systems were analyzed. Mullainathan and Spies \cite{Mullainathan_2017} surveyed the prediction process in general from an econometric perspective.

There were also a number of survey papers concentrated on a single particular \gls{ml} model. Even though these papers focused on one technique, the implementation areas generally spanned various financial applications including financial time series forecasting studies.  Among those soft computing methods, \gls{ec} and \gls{ann} had the most overall interest. 

For the \gls{ec} studies, Chen wrote a book on \glspl{ga} and \gls{gp} in Computational Finance \cite{Chen_2002s}. Later, \glspl{moea} were extensively surveyed on various financial applications including financial time series prediction \cite{Castillo_Tapia_2007, Ponsich_2013, Aguilar_Rivera_2015}. Meanwhile, Rada  reviewed \gls{ec} applications along with Expert Systems for financial investing models \cite{RADA_2008}.

For the \gls{ann} studies, Li and Ma reviewed implementations of \gls{ann} for stock price forecasting and some other financial applications \cite{Li_2010}. The authors of \cite{Tkac_2016} surveyed different implementations of \gls{ann} in financial applications including stock price forecasting.  Recently, Elmsili and Outtaj contained \gls{ann} applications in economics and management research including economic time series forecasting in their survey \cite{Elmsili_2018}.

There were also several text mining surveys focused on financial applications (which included financial time series forecasting). Mittermayer and Knolmayer  compared various text mining implementations that extract market response to news for prediction \cite{Mittermayer_2006}. The authors of \cite{Mitra_2012} focused on news analytics studies for prediction of abnormal returns for trading strategies in their survey. Nassirtoussi et. al. reviewed text mining studies for stock or forex market prediction \cite{Nassirtoussi_2014}. The authors of \cite{Kearney_2014} also surveyed text mining-based time series forecasting and trading strategies using textual sentiment. Similarly, Kumar and Ravi \cite{Kumar_2016} reviewed text mining studies for forex and stock market prediction. Lately, Xing et. al. \cite{Xing_2017} surveyed natural language-based financial forecasting studies.  

Finally, there were application-specific survey papers that focused on particular financial time series forecasting implementations. Among these studies, stock market forecasting had the most interest.  A number of surveys were published for stock market forecasting studies based on various soft computing methods at different times \cite{Vanstone_2003,Hajizadeh_2010,Nair_2014,Cavalcante_2016, Krollner_2010, Yoo, Preethi_2012, Atsalakis_2009}. Chatterjee et. al. \cite{Chatterjee_2000} and Katarya and Mahajan \cite{Katarya_2017} concentrated on \gls{ann}-based financial market prediction studies whereas Hu et. al. \cite{Hu_2015} focused on \gls{ec} implementations for stock forecasting and algorithmic trading models. In a different time series forecasting application, researchers surveyed forex prediction studies using \gls{ann} \cite{Huang_2004} and various other soft computing techniques \cite{Pradeepkumar_2018}.

Even though, many surveys exist for \gls{ml} implementations of financial time series forecasting, \gls{dl} has not been surveyed comprehensively so far despite the existence of various \gls{dl} implementations in recent years. Hence, this was our main motivation for the survey. At this point, we would like to cover the various \gls{dl} models used in financial time series forecasting studies.

%% file: 3_DL.tex
\section{Deep Learning}
\label{sec:dl}

\gls{dl} is a type of \gls{ann} that consists of multiple processing layers and enables high-level abstraction to model data.  The key advantage of \gls{dl} models is extracting the good features of input data automatically using a general-purpose learning procedure. Therefore, in the literature, \gls{dl} models are used in lots of applications: image, speech, video, audio reconstruction, natural language understanding (particularly topic classification), sentiment analysis, question answering and language translation \cite{LeCun2015}. The historical improvements on \gls{dl} models are surveyed in \cite{Schmidhuber_2015}. 

Financial time series forecasting has been very popular among \gls{ml} researchers for more than 40 years. The financial community got a new boost lately with the introduction of \gls{dl} models for financial prediction research and a lot of new publications appeared accordingly. The success of \gls{dl} over \gls{ml} models is the major attractive point for the finance researchers. With more financial time series data and different deep architectures, new \gls{dl} methods will be proposed. In our survey, we found that in the vast majority of the studies, \gls{dl} models were better than \gls{ml} counterparts.

In literature, there are different kinds of \gls{dl} models: \gls{dmlp}, \gls{rnn}, \gls{lstm}, \gls{cnn}, \glspl{rbm}, \gls{dbn}, \gls{ae}, and \gls{drl} \cite{LeCun2015,Schmidhuber_2015}. Throughout the literature, financial time series forecasting was mostly considered as a regression problem. However, there were also a significant number of studies, in particular trend prediction, that used classification models to tackle financial forecasting problems. In Section~\ref{sec:financial_application}, different \gls{dl} implementations are provided along with their model choices.

\subsection{Deep Multi Layer Perceptron (DMLP)}

\glspl{dmlp} is one of the first developed \glspl{ann}. The difference from shallow nets is that \gls{dmlp} contains more layers. Even though particular model architectures might have variations depending on different problem requirements, \gls{dmlp} models consist of mainly three layers: input, hidden and output. The number of neurons in each layer and the number of layers are the hyperparameters of the network. In general, each neuron in the hidden layers has input ($x$), weight ($w$) and bias ($b$) terms. In addition, each neuron has a nonlinear activation function which produces a cumulative output of the preceding neurons.  Equation~\ref{eq:neuron1} \cite{Goodfellow-et-al-2016} illustrates an output of a single neuron in the \gls{nn}. There are different types of nonlinear activation functions. Most commonly used nonlinear activation functions are: sigmoid (Equation~\ref{eq:sigmoid}) \cite{Cybenko_1989}, hyperbolic tangent (Equation~\ref{eq:tanh}) \cite{Kalman_1992}, \gls{relu} (Equation~\ref{eq:relu}) \cite{Nair_2010}, leaky-\gls{relu} (Equation~\ref{eq:relu_leaky}) \cite{Maas_2013}, swish (Equation~\ref{eq:swish}) \cite{Ramachandran_2017}, and softmax (Equation~\ref{eq:softmax}) \cite{Goodfellow-et-al-2016}. The comparison of the nonlinear activations are studied in \cite{Ramachandran_2017}.

\begin{equation}\label{eq:neuron1} 
y_i= \sigma (\sum\limits_{i} W_i x_i + b_i) 
\end{equation}

\begin{equation}\label{eq:sigmoid} 
\sigma(z)= \frac{1}{1+e^{-z}} 
\end{equation}

\begin{equation}\label{eq:tanh} 
\tanh(z)= \frac{e^{z}-e^{-z}}{e^{z}+e^{-z}} 
\end{equation}

\begin{equation}\label{eq:relu} 
R(z)= \max{(0,z)} 
\end{equation}

\begin{equation}\label{eq:relu_leaky} 
R(z)= 1(x<0)(\alpha x)+1(x>=0)(x)
\end{equation}

\begin{equation}\label{eq:swish} 
f(x) =x \sigma(\beta x) 
\end{equation}

\begin{equation}\label{eq:softmax} 
\softmax(z_i)= \frac{\exp{z_i}}{\sum\limits_{j} \exp{z_j}} 
\end{equation}

\begin{figure}[H]
\centering
\includegraphics[width=6in]{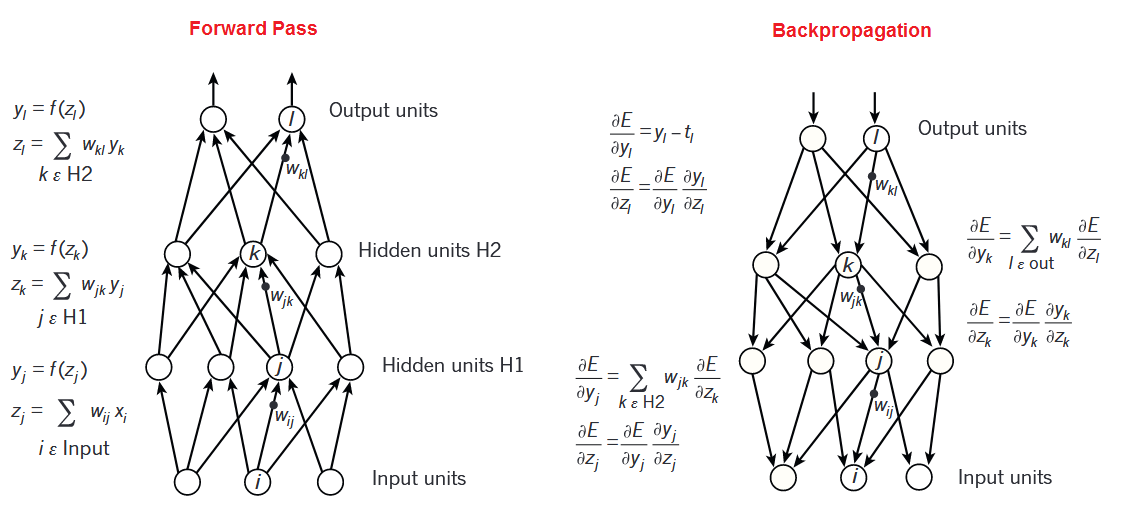}
\caption{Deep Multi Layer Neural Network Forward Pass and Backpropagation \cite{LeCun2015}}
\label{fig:forward-backward}
\end{figure}

\gls{dmlp}  models have been appearing in various application areas  \cite{Deng_2014_App, LeCun2015} . Using a  \gls{dmlp} model has advantages and disadvantages depending on the problem requirements. Through \gls{dmlp} models, problems such as regression and classification can be solved by modeling the input data \cite{Gardner_1998}. However, if the number of the input features is increased (e.g. image as input), the parameter size in the network will increase accordingly due to the fully connected nature of the model and it will jeopardize the computation performance and create storage problems. To overcome this issue, different types of \gls{dnn} methods are proposed (such as \gls{cnn}) \cite{LeCun2015}. With \gls{dmlp}, much more efficient classification and regression processes are performed. In Figure~\ref{fig:forward-backward}, a \gls{dmlp} model, layers, neurons in layers, weights between neurons are shown.

\gls{dmlp} learning stage is implemented through backpropagation. The amount of error in the neurons in the output layer is propagated back to the preceeding layers. Optimization algorithms are used to find the optimum parameters/variables of the \glspl{nn}. They are used to update the weights of the connections between the layers.  There are different optimization algorithms that are developed:  \gls{sgd}, \gls{sgd} with Momentum, \gls{adagrad}, \gls{rmsprop}, \gls{adam} \cite{Robbins_1951,Sutskever_2013,Duchi_2011,Tieleman_2012,Kingma_2014}. Gradient descent is an iterative method to find optimum parameters of the function that minimizes the cost function. \gls{sgd} is an algorithm that  randomly selects a few samples instead of the whole data set for each iteration \cite{Robbins_1951}. \gls{sgd} with Momentum remembers the update in each iteration that accelerates gradient descent method \cite{Sutskever_2013}. \gls{adagrad} is a modified \gls{sgd} that improves convergence performance over standard \gls{sgd} algorithm \cite{Duchi_2011}. \gls{rmsprop} is an optimization algorithm that provides the adaptation of the learning rate for each of the parameters. In \gls{rmsprop}, the learning rate is divided by a running average of the magnitudes of recent gradients for that weight \cite{Tieleman_2012}. \gls{adam} is updated version of \gls{rmsprop} that uses running averages of both the gradients and the second moments of the gradients. \gls{adam} combines advantages of the \gls{rmsprop} (works well in online and non-stationary settings) and \gls{adagrad} (works well with sparse gradients) \cite{Kingma_2014}.

As shown in Figure~\ref{fig:forward-backward}, the effect of the backpropagation is transferred to the previous layers. If the effect of \gls{sgd} is gradually lost when the effect reaches the early layers during backpropagation, this problem is called vanishing gradient problem in the literature \cite{Bengio_1994}. In this case, updates between the early layers become unavailable and the learning process stops. The high number of layers in the neural network and the increasing complexity cause the vanishing gradient problem.

The important issue in the \gls{dmlp} are the hyperparameters of the networks and method of tuning these hyperparameters. Hyperparameters are the variables of the network that affect the network architecture, and the performance of the networks. The number of hidden layers, the number of units in each layer, regularization techniques (dropout, L1, L2), network weight initialization (zero, random, He \cite{He_2015}, Xavier \cite{Glorot_2010}), activation functions (Sigmoid, \gls{relu}, hyperbolic tangent, etc.), learning rate, decay rate, momentum values, number of epochs, batch size (minibatch size), and optimization algorithms (\gls{sgd}, \gls{adagrad}, \gls{rmsprop}, \gls{adam}, etc.) are the hyperparameters of \gls{dmlp}. Choosing better hyperparameter values/variables for the network result in better performance. So, finding the best hyperparameters for the network is a significant issue. In literature, there are different methods to find best hyperparameters: \gls{manualsearch}, \gls{gridsearch}, \gls{randomsearch}, Bayesian Methods (\gls{smbgo}, \gls{gpa}, \gls{tspea}) \cite{Bergstra_2011, Bergstra_2012}.

\subsection{Recurrent Neural Network (RNN)}

\gls{rnn} is another type of \gls{dl} network that is used for time series or sequential data, such as language and speech. \glspl{rnn} are also used in traditional \gls{ml} models (\gls{bptt}, Jordan-Elman networks, etc.), however, the time lengths in such models are generally less than the models used in deep \gls{rnn} models. Deep \glspl{rnn} are preferred due to their ability to include longer time periods. Unlike \glspl{fnn}, \glspl{rnn} use internal memory to process incoming inputs. \glspl{rnn} are used in the analysis of time series data in various fields (handwriting recognition, speech recognition, etc. As stated in the literature, \glspl{rnn} are good at predicting the next character in the text, language translation applications, sequential data processing \cite{Deng_2014_App, LeCun2015}.

\begin{figure}[H]
\centering
\includegraphics[width=3in]{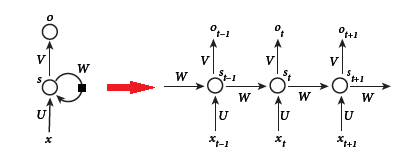}
\caption{RNN cell through time\cite{LeCun2015}}
\label{fig:rnn_fig2}
\end{figure}

\gls{rnn} model architecture consists of different number of layers and different type of units in each layer. The main difference between \gls{rnn} and  \gls{fnn}  is that each \gls{rnn} unit takes the current and previous input data at the same time. The output depends on the previous data in \gls{rnn} model. The \glspl{rnn} process input sequences one by one at any given time, during their operation. In the units on the hidden layer, they hold information about the history of the input in the ``state vector". When the output of the units in the hidden layer is divided into different discrete time steps, the \glspl{rnn} are converted into a \gls{dmlp} \cite{LeCun2015}. In Figure~\ref{fig:rnn_fig2}, the information flow in the \gls{rnn}'s hidden layer is divided into discrete times. The status of the node $S$ at different times of $t$ is shown as $s_t$, the input value $x$ at different times is $x_t$, and the output value $o$ at different times is shown as $o_t$. Parameter values ($U, W, V$) are always used in the same step.

\glspl{rnn} can be trained using the \gls{bptt} algorithm. Optimization algorithms (\gls{sgd}, \gls{rmsprop}, \gls{adam}) are used for weight adjustment process. With the \gls{bptt} learning method, the error change at any $t$ time is reflected in the input and weights of the previous $t$ times. The difficulty of training \gls{rnn} is due to the fact that the \gls{rnn} structure has a backward dependence over time. Therefore, \glspl{rnn} become very complex in terms of the learning period.  Although the main aim of using \gls{rnn} is to learn long-term dependencies, studies in the literature show that when knowledge is stored for long time periods, it is not easy to learn with \gls{rnn} (training difficulties on \gls{rnn}) \cite{Pascanu_2013}. In order to solve this particular problem,  \glspl{lstm} with different structures of \gls{ann} were developed \cite{LeCun2015}. Equations~\ref{eq:rnn1},~\ref{eq:rnn2} illustrate simpler \gls{rnn} formulations. Equation~\ref{eq:rnn3} shows the total error which is the sum of	each error at time step $t$\footnote{Richard Socher, CS224d: Deep Learning for Natural Language Processing, Lecture Notes}.

\begin{equation}\label{eq:rnn1} 
h_t= W f(h_{t-1}) + W^{(hx)}x_{[t]} 
\end{equation}

\begin{equation}\label{eq:rnn2} 
y_t= W^{(S)} f(h_t)
\end{equation}

\begin{equation}\label{eq:rnn3} 
\frac{\partial E}{\partial W}= \sum\limits_{t=1}^T \frac{\partial E_t}{\partial W}
\end{equation}

Hyperparameters of \gls{rnn} also define the network architecture and the performance of the network is affected by the parameter choices as was in  \gls{dmlp} case. The number of hidden layers, the number of units in each layer, regularization techniques, network weight initialization, activation functions, learning rate, momentum values, number of epochs, batch size (minibatch size), decay rate, optimization algorithms, model of  \gls{rnn} (Vanilla \gls{rnn}, \gls{gru}, \gls{lstm}), sequence length for  \gls{rnn} are the hyperparameters of \gls{rnn}. Finding the best hyperparameters for the network is a significant issue. In literature, there are different methods to find best hyperparameters: \gls{manualsearch}, \gls{gridsearch}, \gls{randomsearch}, Bayesian Methods (\gls{smbgo}, \gls{gpa}, \gls{tspea}) \cite{Bergstra_2011, Bergstra_2012}.

\subsection{Long Short Term Memory (LSTM)}

\gls{lstm} \cite{hochreiter1997lstm} is a type of \gls{rnn} where the network can remember both short term and long term values. \gls{lstm} networks are the preferred choice of many \gls{dl} model developers when tackling complex problems like automatic speech recognition, and handwritten character recognition. \gls{lstm}  models are mostly used with time-series data. It is used in different applications such as \gls{nlp}, language modeling, language translation, speech recognition, sentiment analysis, predictive analysis, financial time series analysis, etc. \cite{Wu_2016, Greff_2016}. With attention modules and  \gls{ae} structures, \gls{lstm} networks can be more successful on time series data analysis such as language translation \cite{Wu_2016}.

\gls{lstm} networks consist of \gls{lstm} units. Each \gls{lstm} unit merges to form an \gls{lstm} layer. An \gls{lstm} unit is composed of cells having input gate, output gate and forget gate. Three gates regulate the information flow. With these features,  each cell remembers the desired values over arbitrary time intervals. Equations~\ref{eq:lstm1}-\ref{eq:lstm5} show the form of the forward pass of the \gls{lstm} unit \cite{hochreiter1997lstm} ($x_t$: input vector to the \gls{lstm} unit, $f_t$: forget gate's activation vector, $i_t$: input gate's activation vector, $o_t$: output gate's activation vector, $h_t$: output vector of the \gls{lstm} unit, $c_t$: cell state vector, $\sigma_g$: sigmoid function, $\sigma_c$ , $\sigma_h$: hyperbolic tangent function, $*$: element-wise (Hadamard) product, $W$ , $U$: weight matrices that need to be learned, $b$: bias vector parameters that need to be learned) \cite{Greff_2016}.

\begin{equation}\label{eq:lstm1} 
f_t= \sigma_g (W_f x_t + U_f h_{t-1} + b_f) 
\end{equation}

\begin{equation}\label{eq:lstm2} 
i_t= \sigma_g (W_i x_t + U_i h_{t-1} + b_i) 
\end{equation}

\begin{equation}\label{eq:lstm3} 
o_t= \sigma_g (W_o x_t + U_o h_{t-1} + b_o)
\end{equation}

\begin{equation}\label{eq:lstm4} 
c_t= f_t * c_{t-1} + i_t * \sigma_c (W_c x_t + U_c h_{t-1} + b_c)
\end{equation}

\begin{equation}\label{eq:lstm5} 
h_t= o_t * \sigma_h (c_t)
\end{equation}

\gls{lstm} is a specialized version of \gls{rnn}. Therefore, the weight updates and preferred optimization methods are the same. In addition, the hyperparameters of \gls{lstm} are just like \gls{rnn}: the number of hidden layers, the number of units in each layer, network weight initialization, activation functions, learning rate, momentum values, the number of epochs, batch size (minibatch size), decay rate, optimization algorithms, sequence length for \gls{lstm}, gradient clipping , gradient normalization, and dropout\cite{Reimers_2017,Greff_2016}. In order to find the best hyperparameters of \gls{lstm}, the hyperparameter optimization methods that are used for \gls{rnn}  are also applicable to \gls{lstm} \cite{Bergstra_2011, Bergstra_2012}.

\subsection{Convolutional Neural Networks (CNNs)}

\gls{cnn} is a type of \gls{dnn} that consists of convolutional layers that are based on the convolutional operation. Meanwhile, \gls{cnn} is the most common model that is frequently used for vision or image processing based classification problems (image classification, object detection, image segmentation, etc.) \cite{Ji_2012, Szegedy_2013, Long_2015}. The advantage of the usage of \gls{cnn} is the number of parameters when comparing the vanilla \gls{dl} models such as \gls{dmlp}. Filtering with kernel window function gives an advantage of image processing to \gls{cnn} architectures with fewer parameters that are beneficial for computing and storage. In \gls{cnn} architectures, there are different layers: convolutional, max-pooling, dropout and fully connected \gls{mlp} layer. The convolutional layer consists of the convolution (filtering) operation. Basic convolution operation is shown in Equation~\ref{eq:convolution}  ($t$ denotes time, $s$ denotes feature map, $w$ denotes kernel, $x$ denotes input, $a$ denotes variable). In addition, the convolution operation is implemented on two-dimensional images. Equation~\ref{eq:convolution2d} shows the convolution operation of two-dimensional image ($I$ denotes input image, $K$ denotes the kernel, $m$ and $n$ denote the dimension of images, $i$ and $j$ denote variables). Besides, consecutive convolutional and max-pooling layers construct the deep network.  Equation~\ref{eq:cnn} provides the details about the \gls{nn} architecture ($W$ denotes weights, $x$ denotes input, $b$ denotes bias, $z$ denotes the output of neurons). At the end of the network, the softmax function is used to get the output. Equation~\ref{eq:softmax_simple} and \ref{eq:softmax_eq} illustrate the softmax function ($y$ denotes output) \cite{Goodfellow-et-al-2016}.

\begin{equation}\label{eq:convolution} 
s(t) = (x*w)(t) = \sum\limits_{a=-\infty}^\infty x(a)w(t-a) 
\end{equation}

\begin{equation}\label{eq:convolution2d} 
S(i,j) =  (I * K)(i, j) = \sum\limits_{m}\sum\limits_{n} I(m, n)K(i-m, j-n).
\end{equation}

\begin{equation}\label{eq:cnn} 
z_i = \sum\limits_{j} W_i,_j x_j +b_i.
\end{equation}

\begin{equation}\label{eq:softmax_simple} 
y = \softmax(z)
\end{equation}

\begin{equation}\label{eq:softmax_eq} 
\softmax(z_i) = \frac{\exp(z_i)}{\sum\limits_{j} \exp(z_j)}
\end{equation}

The backpropagation process is used for model learning of \gls{cnn}. Most commonly used optimization algorithms (\gls{sgd}, \gls{rmsprop}) are used to find optimum parameters of \gls{cnn}. Hyperparameters of \gls{cnn} are similar to other \gls{dl} model hyperparameters: the number of hidden layers, the number of units in each layer, network weight initialization, activation functions, learning rate, momentum values, the number of epochs, batch size (minibatch size), decay rate, optimization algorithms, dropout, kernel size, and filter size. In order to find the best hyperparameters of \gls{cnn}, usual search algorithms are used: \gls{manualsearch}, \gls{gridsearch}, \gls{randomsearch}, and Bayesian Methods. \cite{Bergstra_2011, Bergstra_2012}.

\subsection{Restricted Boltzmann Machines (RBMs)}

\gls{rbm} is a productive stochastic \gls{ann} that can learn probability distribution on the input set \cite{Qiu2014}. \glspl{rbm} are mostly used for unsupervised learning \cite{Hrasko_2015}. \glspl{rbm} are used in applications such as dimension reduction, classification, feature learning, collaborative filtering \cite{Salakhutdinov_2007}. The advantage of the \glspl{rbm} is to find hidden patterns with an unsupervised method. The disadvantage of \glspl{rbm} is its difficult training process. ``\glspl{rbm} are tricky because although there are good estimators of the log-likelihood gradient, there are no known cheap ways of estimating the log-likelihood itself" \cite{Bengio_2012}.

\begin{figure}[H]
\centering
\includegraphics[width= 3.7in]{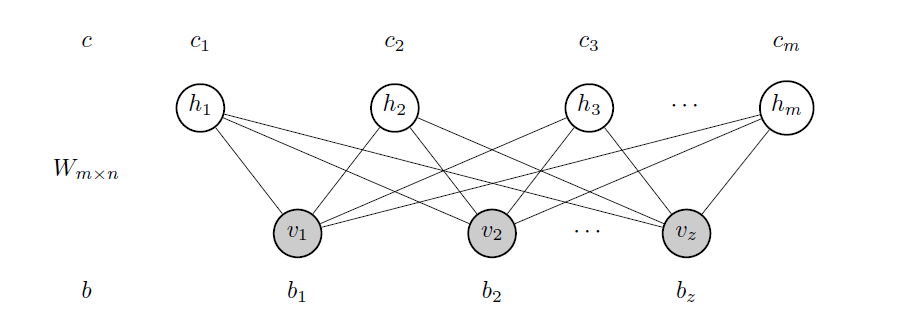}
\caption{RBM Visible and Hidden Layers \cite{Qiu2014}}
\label{fig:rbm_fig}
\end{figure}

\gls{rbm} is a two-layer,  bipartite, and undirected graphical model that consists of two layers; visible and hidden layers (Figure~\ref{fig:rbm_fig}).  The layers are not connected among themselves. Each cell is a computational point that processes the input and makes stochastic decisions about whether this nerve node will transmit the input. Inputs are multiplied by specific weights, certain threshold values (bias) are added to input values, then calculated values are passed through an activation function. In  reconstruction stage, the results in the outputs re-enter the network as the input, then they exit from the visible layer as the output. The values of the previous input and the values after the processes are compared. The purpose of the comparison is to reduce the difference. 

Equation~\ref{eq:rbm1} illustrates the probabilistic semantics for an \gls{rbm} by using its energy function ($P$ denotes the probabilistic semantics for an \gls{rbm}, $Z$ denotes the partition function, $E$ denotes the energy function, $h$ denotes hidden units, $v$ denotes visible units).Equation~\ref{eq:rbm_partition} illustrates the partition function or the normalizing constant. Equation~\ref{eq:rbm2} shows the energy of a configuration (in matrix notation) of the standard type of \gls{rbm} that has binary-valued hidden and visible units ($a$ denotes bias weights (offsets) for the visible units,  $b$ denotes bias weights for the hidden units, $W$ denotes matrix weight of the connection between hidden and visible units, $T$ denotes the transpose of matrix, $v$ denotes visible units, $h$ denotes hidden units) \cite{mohamed2009deep,lee2009convolutional}.

\begin{equation}\label{eq:rbm1} 
P(v,h)=\frac{1}{Z} \exp({-E(v,h)})
\end{equation}

\begin{equation}\label{eq:rbm_partition} 
Z=\sum\limits_{v} \sum\limits_{h} \exp({-E(v,h)})
\end{equation}

\begin{equation}\label{eq:rbm2} 
E(v,h)=-a^T v -b^T h - v^T Wh
\end{equation}

The learning is performed multiple times on the network \cite{Qiu2014}. The training of \glspl{rbm} is implemented through minimizing the negative log-likelihood of the model and data.
 \gls{cd} algorithm is used for the stochastic approximation algorithm which replaces the model expectation for an estimation using Gibbs Sampling with a limited number of iterations \cite{Hrasko_2015}. In the \gls{cd} algorithm, the \gls{kldivergence} algorithm is used to measure the distance between its reconstructed probability distribution and the original probability distribution of the input \cite{Van_2009}.

Momentum, learning rate, weight-cost (decay rate), batch size (minibatch size), regularization method, the number of epochs, the number of layers, initialization of weights, size of visible units, size of hidden units, type of activation units (sigmoid, softmax, \gls{relu}, Gaussian units, etc.), loss function, and optimization algorithms are the hyperparameters of \glspl{rbm}. Similar to the other deep networks, the hyperparameters are searched with \gls{manualsearch}, \gls{gridsearch}, \gls{randomsearch}, and bayesian methods (Gaussian process). In addition to these, \gls{ais} is used to estimate the partition function. \gls{cd} algorithm is also used for the optimization of \glspl{rbm} \cite{Bergstra_2011, Bergstra_2012, Yao_2016, Carreira_2005}.

\subsection{Deep Belief Networks (DBNs)}

\begin{figure}[H]
\centering
\includegraphics[width=2.4in]{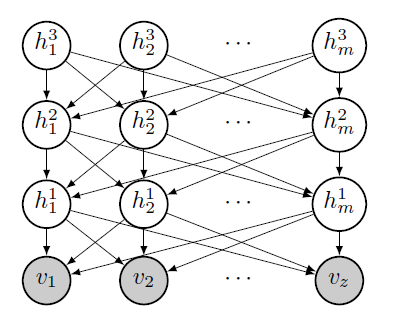}
\caption{Deep Belief Network \cite{Qiu2014} }
\label{fig:deep_belief_fig}
\end{figure}

\gls{dbn} is a type of deep \gls{ann} and consists of a stack of \gls{rbm} networks (Figure~\ref{fig:deep_belief_fig}). \gls{dbn} is a probabilistic generative model that consists of latent variables. In \gls{dbn}, there is no link between units in each layer. \glspl{dbn} are used to find discriminate independent features in the input set using unsupervised learning \cite{mohamed2009deep}. The ability to encode the higher-order network structures and fast inference are the advantages of the DBNs \cite{Tamilselvan_2013}. \glspl{dbn} have disadvantages of training like \glspl{rbm} which is mentioned in the \gls{rbm} section, (\glspl{dbn} are composed of \glspl{rbm}).

When \gls{dbn} is trained on the training set in an unsupervised manner, it can learn to reconstruct the input set in a probabilistic way. Then the layers on the network begin to detect discriminating features in the input. After this learning step, supervised learning is carried out to perform the classification \cite{Hinton2006}. Equation~\ref{eq:dbn1} illustrates the probability of generating a visible vector ($W$: matrix weight of connection between hidden unit $h$ and visible unit $v$, $p(h|W)$: the prior distribution over hidden vectors) \cite{mohamed2009deep}.

\begin{equation}\label{eq:dbn1} 
p(v)=\sum\limits_{h} p(h|W) p(v|h,W)
\end{equation}

\gls{dbn} training process can be divided into two steps: stacked \gls{rbm} learning and backpropagation learning. In stacked \gls{rbm} learning, iterative \gls{cd} algorithm is used \cite{Hrasko_2015}. In backpropagation learning, optimization algorithms (\gls{sgd}, \gls{rmsprop}, \gls{adam}) are used to train network \cite{Tamilselvan_2013}.  \glspl{dbn}' hyperparameters are similar to RBMs' hyperparameters. Momentum, learning rate, weight-cost (decay rate), regularization method, batch size (minibatch size), the number of epochs, the number of layers, initialization of weights, the number of \gls{rbm} stacks, size of visible units in \glspl{rbm}' layers, size of hidden units in \glspl{rbm}' layer, type of units (sigmoid, softmax, rectified, Gaussian units, etc.), network weight initialization, and optimization algorithms are the hyperparameters of DBNs. Similar to the other deep networks, the hyperparameters are searched with \gls{manualsearch}, \gls{gridsearch}, \gls{randomsearch}, and Bayesian methods. \gls{cd} algorithm is also used for the optimization of \glspl{dbn} \cite{Bergstra_2011, Bergstra_2012, Yao_2016, Carreira_2005}.

\subsection{Autoencoders (AEs)}

\gls{ae} networks are \gls{ann} types that are used as unsupervised learning models. In addition, \gls{ae} networks are commonly used in \gls{dl} models, wherein they remap the inputs (features) such that the inputs are more representative for classification. In other words, \gls{ae} networks perform an unsupervised feature learning process, which fits very well with the \gls{dl} theme. A representation of a data set is learned by reducing the dimensionality with \glspl{ae}. \glspl{ae} are similar to \glspl{ffnn}' architecture. They consist of an input layer, an output layer and one or more hidden layers that connect them together. The number of nodes in the input layer and the number of nodes in the output layer are equal to each other in \glspl{ae}, and they have a symmetrical structure. The most notable advantages of \glspl{ae} are dimensionality reduction and feature learning. Meanwhile, reducing dimensionality and feature extraction in \glspl{ae} cause some drawbacks. Focusing on minimizing the loss of data relationship in encoding of \gls{ae} cause the loss of some significant data relationships. Hence, this may be considered as a drawback of \glspl{ae}\cite{Meng_2017}.

In general, \glspl{ae} contain two components: encoder and decoder. The input $x\in [0,1]^d$ is converted through function $f(x)$ ($W_1$ denotes a weight matrix of encoder, $b_1$ denotes a bias vector of encoder, $\sigma_1$ element-wise sigmoid activation function of encoder). Output $h$ is the encoded part of  \glspl{ae} (code), latent variables, or latent representation. The inverse of function $f(x)$, called function $g(h)$, produces the reconstruction of output $r$ ($W_2$ denotes a weight matrix of decoder, $b_2$ denotes a bias vector of decoder, $\sigma_2$ element-wise sigmoid activation function of decoder). Equations~\ref{eq:autoencoder1} and ~\ref{eq:autoencoder2} illustrate the simple AE process \cite{Vincent_2008}. Equation~\ref{eq:autoencoder3} shows the loss function of the \gls{ae}, the \gls{mse}. In the literature, \glspl{ae} have been used for feature extraction and dimensionality reduction \cite{Goodfellow-et-al-2016,Vincent_2008}.

\begin{equation}\label{eq:autoencoder1} 
h = f(x) = {\sigma_1}({W_1}x+b_1)
\end{equation}

\begin{equation}\label{eq:autoencoder2} 
r = g(h) = {\sigma_2}({W_2}h+b_2)
\end{equation}

\begin{equation}\label{eq:autoencoder3} 
L(x,r) = {||x-r||}^2 
\end{equation}

\glspl{ae} are a specialized version of \glspl{ffnn}. The backpropagation learning is used for the update of the weights in the network\cite{Goodfellow-et-al-2016}. Optimization algorithms (\gls{sgd}, \gls{rmsprop}, \gls{adam}) are used for the learning process of \glspl{ae}. \gls{mse} is used as a loss function in \glspl{ae}. In addition, recirculation algorithms may also be used for the training of the \glspl{ae} \cite{Goodfellow-et-al-2016}. \glspl{ae}' hyperparameters are similar to \gls{dl} hyperparameters. Learning rate, weight-cost (decay rate), dropout fraction, batch size (minibatch size), the number of epochs, the number of layers, the number of nodes in each encoder layers, type of activation functions, number of nodes in each decoder layers, network weight initialization, optimization algorithms, and the number of nodes in the code layer (size of latent representation) are the hyperparameters of \glspl{ae}. Similar to the other deep networks, the hyperparameters are searched with \gls{manualsearch}, \gls{gridsearch}, \gls{randomsearch}, and Bayesian methods \cite{Bergstra_2011, Bergstra_2012}.

\subsection{Deep Reinforcement Learning (DRL)}

\gls{rl} is a type of learning method that differs from supervised and unsupervised learning models. It does not need a preliminary data set which is labeled or clustered before. \gls{rl} is an ML approach inspired by learning action/behavior, which deals with what actions should be taken by subjects to achieve the highest reward in an environment. There are different application areas that are used: game theory, control theory, multi-agent systems, operations research, robotics, information theory, managing investment portfolio, simulation-based optimization, playing Atari games, and statistics \cite{sutton1998introduction}. Some of the advantages of using \gls{rl} for control problems are that an agent can be easily re-trained to adapt to changes in the environment and that the system is continually improved while training is constantly performed. An \gls{rl} agent learns by interacting with its surroundings and observing the results of these interactions. This learning method mimics the basic way of how people learn. 

\gls{rl} is mainly based on \gls{mdp}. \gls{mdp} is used to formalize the \gls{rl} environment. \gls{mdp} consists of five tuples: state (finite set of states), action (finite set of actions), reward function (scalar feedback signal), state transition probability matrix ($p(s',r|s,a)$, $s'$ denotes next state, $r$ denotes reward function, $s$ denotes state, $a$ denotes action), discount factor ($\gamma$, present value of future rewards). The aim of the agent is to maximize the cumulative reward. The return ($G_t$) is the total discounted reward. Equation~\ref{eq:rl_return} illustrates the total return ($G_t$ denotes total discounted reward, $R$ denotes rewards, $t$ denotes time, $k$ denotes variable in time).

\begin{equation}\label{eq:rl_return} 
G_t = R_{t+1} + \gamma R_{t+2}+ \gamma^2 R_{t+3} + ... = \sum\limits_{k=0}^\infty  \gamma^k R_{t+k+1}
\end{equation}

The value function is the prediction of the future values. It informs about how good is state/action. Equation~\ref{eq:value} illustrates the formulation of the value function ($v(s)$ denotes the value function, $E[.]$ denotes the expectation function, $G_t$ denotes the total discounted reward, $s$ denotes the given state, $R$ denotes the rewards, $S$ denotes the set of states, $t$ denotes time). 

\begin{equation}\label{eq:value} 
v(s) = E[G_t | S_t=s] = E[R_{t+1}+\gamma v(S_{t+1}) | S_t=s] 
\end{equation}

Policy ($\pi$) is the agent's behavior strategy. It is like a map from state to action.  There are two types of value functions to express the actions in the policy: state-value function ($v_{\pi}(s)$), action-value function ($q_{\pi}(s,a)$). The state-value function (Equation~\ref{eq:state-value}) is the expected return of starting from $s$ to following policy $\pi$ ($E_{\pi}[.]$ denotes expectation function). The action-value function (Equation~\ref{eq:action-value}) is the expected return of starting from $s$, taking action a to following policy $\pi$ ($A$ denotes the set of actions, $a$ denotes the given action).

\begin{equation}\label{eq:state-value} 
v_{\pi}(s) = E_{\pi}[G_t | S_t =s] = E_{\pi}[ \sum\limits_{k=0}^\infty  \gamma^k R_{t+k+1} | S_t =s]
\end{equation}

\begin{equation}\label{eq:action-value} 
q_{\pi}(s,a) = E_{\pi}[G_t | S_t =s, A_t=a] 
\end{equation}

The optimal state-value function (Equation~\ref{eq:optimal-state-value}) is the maximum value function over all policies. The optimal action-value function (Equation~\ref{eq:optimal-action-value}) is the maximum action-value function over all policies.

\begin{equation}\label{eq:optimal-state-value} 
v_{*}(s) = \max( v_{\pi}(s) )
\end{equation}

\begin{equation}\label{eq:optimal-action-value} 
q_{*}(s,a) = \max( q_{\pi}(s,a) )
\end{equation}

The \gls{rl} solutions and methods in the literature are too broad to review in this paper. So, we summarized the important issues of \gls{rl}, important \gls{rl} solutions and methods. \gls{rl} methods are mainly divided into two sections: Model-based methods and model-free methods. The model-based method uses a model that is known by the agent before, value/policy and experience. The experience can be real (sample from the environment) or simulated (sample from the model). Model-based methods are mostly used in the application of robotics, and control algorithms \cite{Nguyen_2011}. Model-free methods are mainly divided into two groups: Value-based and policy-based methods. In value-based methods, a policy is produced directly from the value function (e.g. epsilon-greedy). In policy-based methods, the policy is parametrized directly. In value-based methods, there are three main solutions for \gls{mdp} problems: \gls{dp},  \gls{mc}, and  \gls{td}. 

In \gls{dp} method, problems are solved with optimal substructure and overlapping subproblems. The full model is known and it is used for planning in \gls{mdp}. There are two iterations (learning algorithms) in \gls{dp}: policy iteration and value iteration. \gls{mc} method learns experience directly by running an episode of game/simulation. \gls{mc} is a type of model-free method that does not need \gls{mdp} transitions/rewards. It collects states, returns and it gets mean of returns for the value function. \gls{td} is also a model-free method that learns the experience directly by running the episode. In addition, \gls{td} learns incomplete episodes like the \gls{dp} method by using bootstrapping. \gls{td}  method combines \gls{mc} and \gls{dp} methods. SARSA (state, action, reward, state, action; $S_t$, $A_t$, $R_t$, $S_{t+1}$, $A_{t+1}$) is a type of \gls{td} control algorithm. Q-value (action-value function) is updated with the agent actions. It is an on-policy learning model that learns from actions according to the current policy $\pi$. Equation~\ref{eq:SARSA} illustrates the update of the action-value function in SARSA algorithm ($S_t$ denotes current state, $A_t$ denotes current action, $t$ denotes time, $R$ denotes reward, $\alpha$ denotes learning rate, $\gamma$ denotes discount factor). Q-learning is another \gls{td} control algorithm.  It is an off-policy learning model that learns from different actions that do not need the policy $\pi$ at all. Equation~\ref{eq:QLearning} illustrates the update of the action-value function in Q-Learning algorithm (The whole algorithms can be reached in \cite{sutton1998introduction}, $a'$ denotes action).

\begin{equation}\label{eq:SARSA} 
Q(S_t, A_t) = Q(S_t, A_t) + \alpha [R(t+1) + \gamma  Q(S_{t+1},A_{t+1}) -Q(S_t, A_t)]
\end{equation}

\begin{equation}\label{eq:QLearning} 
Q(S_t, A_t) = Q(S_t, A_t) + \alpha [R(t+1) + \gamma max_{a'} Q(S_{t+1},a') -Q(S_t, A_t)]
\end{equation}

In the value-based methods, a policy can be generated directly from the value function (e.g. using epsilon-greedy). The policy-based method uses the policy directly instead of using the value function. It has advantages and disadvantages over the value-based methods. The policy-based methods are more effective in high-dimensional or continuous action spaces, and have better convergence properties when compared against the value-based methods. It can also learn the stochastic policies. On the other hand, the policy-based method evaluates a policy that is typically inefficient and has high variance. It typically converges to a local rather than the global optimum. In the policy-based methods, there are also different solutions: Policy gradient, Reinforce (Monte-Carlo Policy Gradient), Actor-Critic \cite{sutton1998introduction} (Details of policy-based methods can be reached in \cite{sutton1998introduction}).

\gls{drl} methods contain \glspl{nn}. Therefore, \gls{drl} hyperparameters are similar to \gls{dl} hyperparameters. Learning rate, weight-cost (decay rate), dropout fraction, regularization method, batch size (minibatch size), the number of epochs, the number of layers, the number of nodes in each layer, type of activation functions, network weight initialization, optimization algorithms, discount factor, and the number of episodes are the hyperparameters of \gls{drl}. Similar to the other deep networks, the hyperparameters are searched with \gls{manualsearch}, \gls{gridsearch}, \gls{randomsearch} and bayesian methods \cite{Bergstra_2011, Bergstra_2012}.

%% file: 4_Financial_Application.tex
\section{Financial Time Series Forecasting}
\label{sec:financial_application}

The most widely studied financial application area is forecasting of a given financial time series, in particular asset price forecasting. Even though some variations exist, the main focus is on predicting the next movement of the underlying asset. More than half of the existing implementations of \gls{dl} were focused on this area. Even though there are several subtopics of this general problem including Stock price forecasting, Index prediction, forex price prediction, commodity (oil, gold, etc) price prediction, bond price forecasting, volatility forecasting, cryptocurrency price forecasting, the underlying dynamics are the same in all of these applications. 

The studies can also be clustered into two main groups based on their expected outputs: price prediction and price movement (trend) prediction. Even though price forecasting is basically a regression problem, in most of the financial time series forecasting applications, correct prediction of the price is not perceived as important as correctly identifying the directional movement. As a result, researchers consider trend prediction, i.e. forecasting which way the price will change, a more crucial study area compared with exact price prediction. In that sense, trend prediction becomes a classification problem. In some studies, only up or down movements are taken into consideration (2-class problem), whereas up, down or neutral movements (3-class problem) also exist.

\gls{lstm} and its variations along with some hybrid models dominate the financial time series forecasting domain. \gls{lstm}, by its nature utilizes the temporal characteristics of any time series signal, hence forecasting financial time series is a well-studied and successful implementation of \gls{lstm}. However, some researchers prefer to either extract appropriate features from the time series or transform the time series in such a way that, the resulting financial data becomes stationary from a temporal perspective, meaning even if we shuffle the data order, we will still be able to properly train the model and achieve successful out-of-sample test performance. For those implementations, \gls{cnn} and \gls{dfnn} were the most commonly chosen \gls{dl} models.

Various financial time series forecasting implementations using \gls{dl} models exist in literature. We will cover each of these aforementioned implementation areas in the following subsections. In this survey paper, we examined the papers using the following criteria:
\begin{itemize}
\item {First, we grouped the articles according to their subjects.}
\item {Then, we grouped the related papers according to their feature set.}
\item {Finally, we grouped each subgroup according to \gls{dl}  models/methods.}
\end{itemize}

For each implementation area, the related papers will be subgrouped and tabulated. Each table will have the following fields to provide the information about the implementation details for the papers within the group: Article (Art.) and Data Set are trivial, Period refers to the time period for training and testing. Feature Set lists the input features used in the study. Lag has the time length of the input vector (e.g. 30d means the input vector has a 30 day window) and horizon shows how far out into the future is predicted by the model. Some abbreviations are used for these two aforementioned fields: min is minutes, h is hours, d is days, w is weeks, m is months, y is years, s is steps, * is mixed. Method shows the \gls{dl} models that are used in the study. Performance criteria provides the evaluation metrics, and finally the Environment (Env.) lists the development framework/software/tools. Some column values might be empty, indicating there was no relevant information in the paper for the corresponding field.

\subsection{Stock Price Forecasting}

Price prediction of any given stock is the most studied financial application of all. We observed the same trend within the \gls{dl} implementations. Depending on the prediction time horizon, different input parameters are chosen varying from \gls{hft} and intraday price movements to daily, weekly or even monthly stock close prices. Also, technical, fundamental analysis, social media feeds, sentiment, etc. are among the different parameters that are used for the prediction models. 

\input{tables/table1xx_1.tex}

In this survey, first, we grouped the stock price forecasting articles according to their feature set such as studies using only the raw time series data (price data, \gls{ochlv}) for price prediction; studies using various other data and papers that used text mining techniques. Regarding the first group, the corresponding \gls{dl} models were directly implemented using the raw time series for price prediction. Table~\ref{table:stock_price_forecasting_1} tabulates the stock price forecasting papers that used only raw time series data in the literature. In Table~\ref{table:stock_price_forecasting_1}, different methods/models are also listed based on four sub-groups: \gls{dnn} (networks that are deep but without any given topology details) and \gls{lstm} models; multi models; hybrid models; novel methods.

\gls{dnn} and \gls{lstm} models were solely used in 3 papers. In \cite{Chong_2017}, \gls{dnn} and lagged stock returns were used to predict the stock prices in \gls{kospi}. Chen et. al. \cite{Chen_2015}, Dezsi and Nistor \cite{Dezsi_2016} applied the raw price data as the input to \gls{lstm} models. 

Meanwhile, there were some studies implementing multiple \gls{dl} models for performance comparison using only the raw price (\gls{ochlv}) data for forecasting. Among the noteworthy studies, the authors in  \cite{Samarawickrama_2017} compared \gls{rnn}, \gls{srnn}, \gls{lstm} and \gls{gru}. Hiransha et. al. \cite{M_2018} compared \gls{lstm}, \gls{rnn}, \gls{cnn}, \gls{mlp}, whereas in \cite{Selvin_2017} \gls{rnn}, \gls{lstm}, \gls{cnn}, \gls{arima} were preferred, Lee and Yoo \cite{Lee_2018} compared 3 \gls{rnn} models (\gls{srnn}, \gls{lstm}, \gls{gru}) for stock price prediction and then constructed a threshold based portfolio with selecting stocks according to the predictions and Li et. al. \cite{Li_2017} implemented \gls{dbn}. Finally, the authors of \cite{Chen_2018} compared 4 different \gls{ml} models (1 \gls{dl} model - \gls{ae} and \gls{rbm}), \gls{mlp}, \gls{rbf} and \gls{elm} for predicting the next price in 1-minute price data. They also compared the results with different sized datasets. The authors of \cite{Krauss_2017} used price data and \gls{dnn}, \gls{gbt}, \gls{rf} methods for the prediction of the stocks in the \gls{sp500}. In Chandra and Chan \cite{Chandra_2016}, co-operative neuro-evolution, \gls{rnn} (Elman network) and \gls{dfnn} were used for the prediction of stock prices in \gls{nasdaq} (ACI Worldwide, Staples, and Seagate).

Meanwhile, hybrid models were used in some of the papers. The author of \cite{Liu_2017} applied \gls{cnn}+\gls{lstm} in their studies. Heaton et. al. \cite{Heaton_2016} implemented smart indexing with \gls{ae}. The authors of \cite{Batres_2015} combined \gls{dbn} and \gls{mlp} to construct a stock portfolio by predicting each stock's monthly log-return and choosing the only stocks that were expected to perform better than the performance of the median stock.

In addition, some novel approaches were adapted in some of the studies. The author of \cite{Yuan_2018} proposed novel \gls{dwnn} which is combination of \gls{rnn} and \gls{cnn}. The author of \cite{Zhang_2017} implemented \gls{sfm} recurrent network in their studies.  

In another group of studies, some researchers again focused on \gls{lstm} based models. However, their input parameters came from various sources including the raw price data, technical and/or fundamental analysis, macroeconomic data, financial statements, news, investor sentiment, etc. Table~\ref{table:stock_price_forecasting_2} tabulates the stock price forecasting papers that used various data such as the raw price data, technical and/or fundamental analysis, macroeconomic data  in the literature.  In Table~\ref{table:stock_price_forecasting_2}, different methods/models are also listed based on five sub-groups: \gls{dnn} model; \gls{lstm} and \gls{rnn} models; multiple and hybrid models; \gls{cnn} model; novel methods.

\gls{dnn} models were used in some of the stock price forecasting papers within this group. In \cite{Abe_2018}, \gls{dnn} model and 25 fundamental features were used for the prediction of the Japan Index constituents. Feng et. al. \cite{Feng_2018} also used fundamental features and \gls{dnn} model for the prediction. \gls{dnn} model, macro economic data such as GDP, unemployment rate, inventories, etc. were used by the authors of \cite{Fan_2014} for the prediction of the U.S. low-level disaggregated macroeconomic time series.

\gls{lstm} and \gls{rnn} models were chosen in some of the studies. Kraus and Feuerriegel \cite{Kraus_2017} implemented \gls{lstm} with transfer learning using text mining through financial news and the stock market data. Similarly, the author of \cite{Minami_2018} used \gls{lstm} to predict the stock's next day price using corporate action events and macro-economic index.  Zhang and Tan \cite{Zhang_2018_a} implemented DeepStockRanker, an \gls{lstm} based model for stock ranking using 11 technical indicators. In another study \cite{Zhuge_2017}, the authors used the price time series and emotional data from text posts for predicting the stock opening price of the next day with \gls{lstm} network. Akita et. al. \cite{Akita_2016} used textual information and stock prices through Paragraph Vector + \gls{lstm} for forecasting the prices and the comparisons were provided with  different classifiers. Ozbayoglu \cite{Ozbayoglu_2007} used technical indicators along with the stock data on a Jordan-Elman network for price prediction.

There were also multiple and hybrid models that used mostly technical analysis features as their inputs to the \gls{dl} model. Several technical indicators were fed into \gls{lstm} and \gls{mlp} networks in \cite{Khare_2017} for predicting intraday price prediction. Recently, Zhou et. al. \cite{Zhou_2018} used \gls{gan-fd} model for stock price prediction and compared their model performances against \gls{arima}, \gls{ann} and \gls{svm}. The authors of \cite{Singh_2016} used several technical indicator features and time series data with \gls{pca} for dimensionality reduction cascaded with \gls{dnn} (2-layer \gls{ffnn}) for stock price prediction.  In \cite{Karaoglu_2017}, the authors used Market microstructures based trade indicators as inputs into \gls{rnn} with Graves \gls{lstm} detecting the buy-sell pressure of movements in \gls{bist} in order to perform the price prediction for intelligent stock trading.  In \cite{Zhou_2018_a}, next month's return was predicted and top to be performed portfolios were constructed. Good monthly returns were achieved with \gls{lstm} and \gls{lstm}-\gls{mlp} models. 

Meanwhile, in some of the papers, \gls{cnn}  models were preferred. The authors of \cite{Abroyan_2017} used 250 features: order details, etc for the prediction of the private brokerage company’s real data of risky transactions. They used \gls{cnn} and \gls{lstm} for stock price forecasting.  The authors of \cite{GooglePatent} used \gls{cnn} model, fundamental, technical and market data for the prediction.

Novel methods were also developed in some of the studies. In \cite{Tran_2017}, FI-2010 dataset: bid/ask and volume were used as the feature set for the forecast. In the study, they proposed \gls{wmtr}, \gls{mda}. The authors of \cite{Feng_2018_a} used 57 characteristic features such as Market equity, Market Beta, Industry momentum, Asset growth, etc. as inputs to a Fama-French n-factor model \gls{dl} for predicting monthly US equity returns in \gls{nyse}, \gls{amex}, or \gls{nasdaq}.

\input{tables/table1xx_2.tex}

There were a number of research papers that also used text mining techniques for the feature extraction, but used non-\gls{lstm} models for the stock price prediction. Table~\ref{table:stock_price_forecasting_3} tabulates the stock price forecasting papers that used text mining techniques.  In Table~\ref{table:stock_price_forecasting_3}, different methods/models are clustered into three sub-groups: \gls{cnn} and \gls{lstm} models; \gls{gru}, \gls{lstm}, and \gls{rnn} models; novel methods.

\gls{cnn} and \gls{lstm} models were adapted in some of the papers. In \cite{Ding_2015}, events were detected from Reuters and Bloomberg news through text mining and that information was used for the price prediction and stock trading through the \gls{cnn} model.  Vargas et. al. \cite{Vargas_2017} used text mining on \gls{sp500} index news from Reuters through a \gls{lstm}+\gls{cnn} hybrid model for price prediction and intraday directional movement estimation together. The authors of \cite{Lee_2017_b} used the financial news data and implemented word embedding with Word2vec along with MA and stochastic oscillator to create inputs for \gls{rcnn} for stock price prediction. The authors of \cite{Iwasaki_2018} also used sentiment analyses through text mining and word embeddings from analyst reports and used sentiment features as inputs to \gls{dfnn} model for stock price prediction. Then different portfolio selections were implemented based on the projected stock returns.

\gls{gru}, \gls{lstm}, and \gls{rnn} models were preferred in the next group of papers. Das et. al. \cite{Das_2018} implemented sentiment analysis on Twitter posts along with the stock data for price forecasting using \gls{rnn}. Similarly, the authors of \cite{Jiahong_Li_2017} used sentiment classification (neutral, positive, negative) for the stock open or close price prediction with various \gls{lstm} models. They compared their results with \gls{svm} and achieved higher overall performance. In \cite{Zhongshengz_2018}, text and price data were used for the prediction of the \gls{sci} prices.  

Some novel approaches were also found in some of the papers. The authors of \cite{Nascimento_2015} used word embeddings for extracting information from web pages and then combined with the stock price data for stock price prediction. They compared \gls{ar} model and \gls{rf} with and without news. The results showed embedding news information improved the performance. In \cite{Han_2018}, financial news and ACE2005 Chinese corpus were used. Different event-types on Chinese companies were classified based on a novel event-type pattern classification algorithm in \cite{Han_2018}, also next day stock price change was predicted using additional inputs.

\input{tables/table1xx_3.tex}

\subsection{Index Forecasting}

Instead of trying to forecast the price of a single stock, several researchers preferred to predict the stock market index. Indices generally are less volatile than individual stocks, since they are composed of multiple stocks from different sectors and are more indicative of the overall momentum and general state of the economy. 

In the literature, different stock market index data were used for the experiments. Most commonly used index data can be listed as follows: \gls{sp500}, \gls{csi}300, \gls{nifty}, \gls{nikkei}225, \gls{djia}, \gls{sse}180, \gls{hsi}, \gls{szse}, \gls{ftse}100, \gls{taiex}, \gls{bist}, \gls{nasdaq}, \gls{dow30}, \gls{kospi}, \gls{vix}, \gls{vxn}, \gls{bovespa}, \gls{omx}, \gls{nyse}. The authors of \cite{Bao_2017, Parida_2016, Fischer_2018, Widegren_2017, borovykh_2018, Althelaya_2018, Dingli_2017, Rout_2017, Jeong_2019, Baek_2018, Hansson_2017, Elliot_2017, Ding_2015} used \gls{sp500} as their dataset. The authors of  \cite{Bao_2017, Parida_2016, Li_2017a, Namini_2018, Hsieh_2011}  used \gls{nikkei} as their dataset. \gls{kospi} was used in \cite{Li_2017a, Jeong_2019, Baek_2018}. \gls{djia} was used as the dataset in \cite{Bao_2017, Namini_2018, Hsieh_2011, Zhang_2015, Bekiros_2013}. Besides, the authors of \cite{Bao_2017, Li_2017a, Hsieh_2011, Jeong_2019} used \gls{hsi} as the dataset in their studies. \gls{szse} is used in studies of  \cite{Pang_2018, Li_2017a, Deng_2017, Yang_2017}.

In addition, in the literature, there were different methods for the prediction of the index data. While some of the studies used only the raw time series data, some others used various other data such as technical indicators, index data, social media feeds, news from Reuters, Bloomberg, the statistical features of data (standard deviation, skewness, kurtosis, omega ratio, fund alpha).  In this survey, first, we grouped the index forecasting articles according to their feature set such as studies using only the raw time series data (price/index data, \gls{ochlv}); then in the second group we clustered the studies using various other data. Table~\ref{table:index_forecasting_1} tabulates the index forecasting papers using only the raw time series data. 
Moreover, different methods (models) were used for index forecasting. \gls{mlp}, \gls{rnn}, \gls{lstm}, \gls{dnn} (most probably \gls{dfnn}, or \gls{dmlp}) methods were the most used methods for index forecasting. In Table~\ref{table:index_forecasting_1}, these various methods/models are also listed as four sub-groups: \gls{ann}, \gls{dnn}, \gls{mlp}, and \gls{fddr} models; \gls{rl} and \gls{dl} models; \gls{lstm} and \gls{rnn} models; novel methods. 

\input{tables/table2xx_1.tex}

\gls{ann}, \gls{dnn}, \gls{mlp}, and \gls{fddr} models were used in some of the studies. In \cite{Lachiheb_2018}, log returns of the index data was used with \gls{dnn} with hierarchical input for the prediction of the TUNINDEX data. The authors of \cite{Yong_2017} used deep \gls{ffnn} and \gls{ochl} of the last 10 days of index data for prediction. In addition, \gls{mlp} and \gls{ann} were used for the prediction of index data. In  \cite{Yumlu_2005}, the raw index data was used with \gls{mlp}, \gls{rnn}, \gls{moe} and \gls{egarch} for the forecast. In \cite{Yang_2017}, ensembles of \gls{ann} with \gls{ochlv} of the data were used for the prediction of the Shanghai composite index.

Furthermore, \gls{rl} and \gls{dl} methods were used together for the prediction of the index data in some of the studies. In \cite{Deng_2017}, \gls{fddr}, \gls{dnn} and \gls{rl} methods were used to predict  300 stocks from \gls{szse} index data and commodity prices. In \cite{Jeong_2019}, Deep Q-Learning and \gls{dnn} methods and 200-days stock price dataset were used together for the prediction of \gls{sp500} index. 

Most of the preferred methods for prediction of the index data using the raw time series data were based on \gls{lstm} and \gls{rnn}. In \cite{Bekiros_2013},  \gls{rnn} was used for prediction of the log returns of \gls{djia} index. In \cite{Fischer_2018}, \gls{lstm} was used to predict \gls{sp500} Index data. The authors of \cite{Althelaya_2018} used stacked \gls{lstm}, \gls{bi-lstm} methods for \gls{sp500} Index forecasting. The authors of \cite{Yan_2017} used \gls{lstm} network to predict the next day closing price of Shanghai stock Index. In their study, they used wavelet decomposition to reconstruct the financial time series for denoising and better learning. In \cite{Pang_2018}, \gls{lstm} was used for the prediction of Shanghai A-shares composite index. The authors of \cite{Namini_2018} used \gls{lstm} to predict \gls{nikkei}225,  IXIC,  HIS,  GSPC and \gls{djia} index data. In \cite{Takahashi_2017} and \cite{Baek_2018}, \gls{lstm} was also used for the prediction of \gls{sp500} and \gls{kospi}200 index. The authors of \cite{Baek_2018} developed an \gls{lstm} based stock index forecasting model called ModAugNet. The proposed method was able to beat \gls{b-h} in the long term with an overfitting prevention mechanism. The authors of \cite{Elliot_2017} compared different \gls{ml} models (linear model), \gls{gml} and several \gls{lstm}, \gls{rnn} models for stock index price prediction. In \cite{Hansson_2017}, \gls{lstm} and autoregressive part of the time series index data were used for prediction of \gls{sp500}, \gls{bovespa}50, \gls{omx}30 indices. 

Also, some studies adapted novel appraches. In \cite{Zhang_2015}, genetic \gls{dnn} was used for \gls{djia} index forecasting. The authors of  \cite{borovykh_2018} proposed a new \gls{dnn} model which is called Wavenet convolutional net for time series forecasting. The authors of \cite{Bildirici_2010} proposed a (\gls{tar}-\gls{vec}-\gls{rhe}) model for forex and stock index of return prediction and compared several models. The authors of \cite{Parida_2016} proposed a method that is called \gls{lrnfis} with \gls{fhso} \gls{ea} to predict \gls{sp500}, \gls{nikkei}225 indices and USD Exchange price data. The authors of \cite{Psaradellis_2016} proposed a \gls{har} with a \gls{gasvr} model that was called \gls{har}-\gls{gasvr} for prediction of \gls{vix}, \gls{vxn}, \gls{vxd} indices.

In the literature, some of the studies used various input data such as technical indicators, index data, social media news, news from Reuters, Bloomberg, the statistical features of data (standard deviation, skewness, kurtosis, omega ratio, fund alpha). Table~\ref{table:index_forecasting_2} tabulates the index forecasting papers using these aforementioned various data. \gls{dnn}, \gls{rnn}, \gls{lstm}, \gls{cnn} methods were the most commonly used models in index forecasting. In Table~\ref{table:index_forecasting_2}, different methods/models are also listed within four sub-groups: \gls{dnn} model; \gls{rnn} and \gls{lstm} models; \gls{cnn} model; novel methods.

\gls{dnn}  was used as the classification model in some of the papers. In \cite{Chen_2016}, \gls{dnn} and some of the feature of the data (Return, \gls{sr}, \gls{std}, Skewness, Kurtosis, Omega ratio, Fund alpha) were used for the prediction. In \cite{Widegren_2017}, \gls{dnn}, \gls{rnn} and technical indicators were used for the prediction of \gls{ftse}100, \gls{omx}30, \gls{sp500} indices. 

In addition, \gls{rnn} and \gls{lstm} models with various other data were also used for the prediction of the indices. The authors of \cite{Hsieh_2011} used \gls{rnn} and \gls{ochlv} of indices, technical indicators to predict \gls{djia}, \gls{ftse}, Nikkei, \gls{taiex} indices. The authors of \cite{Mourelatos_2018} used \gls{gasvr}, \gls{lstm} for the forecast. The authors of \cite{Chen_2018_f} used four \gls{lstm} models (technical analysis, attention mechanism and market vector embedded) for the prediction of the daily return ratio of \gls{hsi}300 index. In \cite{Li_2017a}, \gls{lstm} with wavelet denoising and index data, volume, technical indicators were used for the prediction of the \gls{hsi}, \gls{sse}, \gls{szse}, \gls{taiex}, \gls{nikkei}, \gls{kospi} indices. The authors of \cite{Si_2017} used MODRL+\gls{lstm} method to predict Chinese stock-IF-IH-IC contract indices. The authors of \cite{Bao_2017} used stacked \glspl{ae} to generate deep features using \gls{ochl} of the stock prices, technical indicators and macroeconomic conditions to feed to \gls{lstm} to predict the future stock prices.

\input{tables/table2xx_2.tex}

Besides, different \gls{cnn} implementations with various data (technical indicators, news, index data) were used in the literature. In \cite{Dingli_2017}, \gls{cnn} and index data, technical indicators were used for the \gls{sp500}, \gls{dow30}, \gls{nasdaq}100 indices and Commodity, Forex, Bitcoin prices. In \cite{Ding_2015}, \gls{cnn} model with news from Reuters and Bloomberg were used for the prediction of \gls{sp500} Index and 15 stocks' prices in \gls{sp500}. In \cite{Lee_2017_b}, \gls{cnn} + \gls{lstm} and technical indicators, index data, news were used for the forecasting of \gls{twse} index and 4 stocks' prices in \gls{twse}. 
 
In addition, there were some novel methods proposed for the index forecasting. The authors of \cite{Rout_2017} used \gls{rnn} models, \gls{rceflann}  and \gls{flann}, with their weights optimized using various \gls{ea} like \gls{pso}, HMRPSO and \gls{pso} for time series forecasting. The authors of \cite{Chen_2018_e} used social media news to predict the index price and index direction with \gls{rnn}-Boost with \gls{lda} features. 

\subsection{Commodity Price Forecasting}

There were a number of studies  particularly focused on the price prediction of any given commodity, such as gold, silver, oil, copper, etc. With increasing number of commodities that are available for public trading through online stock exchanges, interest in this topic will likely grow in the following years.

In the literature, there were different methods that were used for commodity price forecasting. \gls{dnn}, \gls{rnn}, \gls{fddr}, \gls{cnn} were the most used models to predict the commodity prices. Table~\ref{table:commodity_price_forecasting} provides the details about the commodity price forecasting studies with \gls{dl}.

In \cite{Dingli_2017}, the authors used \gls{cnn} for predicting the next week and next month price directional movement. Meanwhile, \gls{rnn} and \gls{lstm} models were used in some of the commodity forecasting studies. In \cite{Dixon_2016}, \gls{dnn} was used for Commodity forecasting. In \cite{Widegren_2017}, different datasets (Commodity, forex, index) were used as datasets. \gls{dnn} and \gls{rnn} were used to predict the prices of the time series data. Technical indicators were used as the feature set which consist of \gls{rsi}, \gls{williamr}, \gls{cci}, \gls{pposc}, momentum, \gls{ema}.  In \cite{S_nchez_Lasheras_2015}, the authors used Elman \gls{rnn} to predict COMEX copper spot price (through \gls{nymex}) from daily close prices.

\hl{Hybrid and novel models were adapted in some studies.} In \cite{Zhao_2017}, \gls{fnn} and \gls{sdae} deep models were compared against \gls{svr}, \gls{rw} and \gls{mrs} models for WTI oil price forecasting. As performance criteria, accuracy, \gls{mape}, \gls{rmse} were used.  In \cite{Chen_2017_d}, authors tried to predict WTI crude oil prices using several models including combinations of \gls{dbn}, \gls{lstm}, \gls{arma} and \gls{rw}. \gls{mse} was used as the  performance criteria. In \cite{Deng_2017}, the authors used \gls{fddr} for stock price prediction and trading signal generation. They combined \gls{dnn} and \gls{rl}. Profit, return, SR, profit-loss curves were used as the performance criteria. 

\input{tables/table3xx.tex}

\subsection{Volatility Forecasting}

Volatility is directly related with the price variations in a given time period and is mostly used for risk assesment and asset pricing. Some researchers implemented models for accurately forecasting the underlying volatility of any given asset. 

In the literature, there were different methods that were used for volatility forecasting. \gls{lstm}, \gls{rnn}, \gls{cnn}, MM, \gls{garch} models were shown as some of these methods. Table~\ref{table:volatility_forecasting} summarizes the studies that were focused on volatility forecasting.  In Table~\ref{table:volatility_forecasting}, different methods/models are also represented as three sub-groups: \gls{cnn} model; \gls{rnn} and \gls{lstm} models; hybrid and novel models. 

\gls{cnn}  model was used in one volatility forecasting study based on \gls{hft} data \cite{Doering_2017}.

Meanwhile, \gls{rnn} and \gls{lstm} models were used in some of the researches. In \cite{Tino_2001}, the authors used financial time series data to predict volatility changes with Markov Models and Elman \gls{rnn}  for profitable straddle options trading. The authors of \cite{Xiong_2015} used the price data and different types of Google Domestic trends with \gls{lstm}. The authors of \cite{Zhou_2018_b} used \gls{csi}300, 28 words of the daily search volume based on Baidu as the dataset with \gls{lstm} to predict the  index volatility. The authors of \cite{Kim_2018} developed several \gls{lstm} models integrated with \gls{garch} for the prediction of volatility. 

Hybrid and novel approaches were also adapted in some of the researches. In \cite{Nikolaev_2013}, \gls{rmdn-garch} model was proposed. In addition, several models including traditional forecasting models and \gls{dl} models were compared for the estimation of volatility. The authors of \cite{Psaradellis_2016} proposed a novel method that is called \gls{har-gasvr} for volatility index forecasting.

\input{tables/table4xx.tex}

\subsection{Bond Price Forecasting}

Some financial experts follow the changes in the bond prices to analyze the state of the economy, claiming bond prices represent the health of the economy better than the stock market \cite{Harvey_1989}. Historically, long term rates are higher than the short term rates under normal economic expansion times, whereas just before recessions short term rates pass the long term rates, i.e. the inverted yield curve. Hence, accurate bond price prediction is very useful. However, \gls{dl} implementations for bond price prediction is very scarce. In one study \cite{bianchi_2018}, excess bond return was predicted using several \gls{ml} models including \gls{rf}, \gls{ae} and \gls{pca} network and a 2-3-4-layer \gls{dfnn}. 4 layer \gls{nn} outperformed the other models.

\subsection{Forex Price Forecasting}

Foreign exchange market has the highest volume among all existing financial markets in the world. It is open 24/7 and trillions of dollars worth of foreign exhange transactions happen in a single day.  According to the Bank for International Settlements, foreign-exchange trading had a volume of more than 5 trillion USD a day \cite{Venketas_2019}. In addition, there are a large number of online forex trading platforms that provide leveraged transaction opportunities to their subscribers. As a result, there is a huge interest for profitable trading strategies by traders. Hence, there were a number of forex forecasting and trading studies that were based on \gls{dl} models.  Since most of the global financial transactions were based on US Dollar, almost all forex prediction research papers include USD in their analyses. However, depending on regional differences and intended research focus, various models were developed accordingly. 

In the literature, there were different methods that were used for forex price forecasting. \gls{rnn}, \gls{lstm}, \gls{cnn}, \gls{dbn}, \gls{dnn}, \gls{ae}, \gls{mlp} methods were shown as some of these methods. Table~\ref{table:forex_price_forecasting} provides details about these implementations.  In Table~\ref{table:forex_price_forecasting}, different methods/models are listed as four sub-groups: \gls{cdbn}, \gls{dbn}, \gls{dbn}+\gls{rbm}, and \gls{ae} models; \gls{dnn}, \gls{rnn}, \gls{psn}, and \gls{lstm} models; \gls{cnn} models; hybrid models. 

\gls{cdbn}, \gls{dbn}, \gls{dbn}+\gls{rbm}, and \gls{ae} models were used in some of the studies. In \cite{Zhang_2014}, Fuzzy information granulation integrated with \gls{cdbn} was applied for predicting EUR/USD and GBU/USD exchange rates. They extended \gls{dbn} with \gls{crbm} to improve the performance. In \cite{Chao_2011}, weekly GBP/USD and INR/USD prices were predicted, whereas in \cite{Zheng_2017}, CNY/USD and INR/USD was the main focus. In both cases, \gls{dbn} was compared with \gls{ffnn}. Similarly, the authors in \cite{Shen_2015} implemented several different \gls{dbn} networks to predict weekly GBP/USD, BRL/USD and INR/USD exchange rate returns. The researchers in \cite{Shen_2016} combined Stacked \gls{ae} and \gls{svr} for predicting 28 normalized currency pairs using the time series data of (USD, GBP, EUR, JPY, AUD, CAD, CHF). 

\gls{dnn}, \gls{rnn}, \gls{psn}, and \gls{lstm} models were preferred in some of the researches. In \cite{Dixon_2016}, multiple \gls{dmlp} models were developed for predicting AD and BP futures using 5-minute data in a 130 day period. The authors of \cite{Sermpinis_2012_a} used \gls{mlp}, \gls{rnn}, \gls{gp} and other \gls{ml} techniques along with traditional regression methods for also predicting EUR/USD time series. They also integrated Kalman filter, LASSO operator and other models to further improve the results in \cite{Sermpinis_2012}. They further extended their analyses by including \gls{psn} and providing comparisons along with traditional forecasters like \gls{arima}, RW and STAR \cite{Sermpinis_2014}. To improve the performance they also integrated hybrid time-varying volatility leverage. In \cite{SUN_2009}, the authors implemented RMB exchange rate forecasting against JPY, HKB, EUR and USD by comparing \gls{rw}, \gls{rnn} and \gls{ffnn} performances. In \cite{Maknickien__2013}, the authors predicted various Forex time series and created portfolios consisted of these investments. Each network used \gls{lstm} (\gls{rnn} EVOLINO) and different risk appetites for users have been tested. The authors of \cite{Maknickiene_2014} also used EVOLINO RNN + orthogonal input data for predicting USD/JPY and XAU/USD prices for different periods.

Different \gls{cnn} models were used in some of the studies. In \cite{persio_2016}, EUR/USD was once again forecasted using multiple \gls{dl} models including \gls{mlp}, \gls{cnn}, \gls{rnn} and Wavelet+\gls{cnn}. The authors of \cite{Korczak_2017} implemented forex trading (GBP/PLN) using several different input parameters on a multi-agent based trading environment. One of the agents was using \gls{ae}+\gls{cnn} as the prediction model and outperformed all other models.

Hybrid models were also adapted in some of the researches. The authors of \cite{Bildirici_2010} developed several (TAR-VEC-RHE) models for predicting monthly returns for TRY/USD and compared model performances.  In \cite{Nikolaev_2013}, the authors compared several models including traditional forecasting models and \gls{dl} models for DEM/GBP prediction. The authors in \cite{Parida_2016} predicted AUD, CHF, MAX and BRL against USD currency time series data using LRNFIS and compared it with different models. Meanwhile, instead of using LMS based error minimization during the learning, they used \gls{fhso}.

\input{tables/table5xx.tex}

\subsection{Cryptocurrency Price Forecasting}

Since cryptocurrencies became a hot topic for discussion in the finance world, lots of studies and implementations started emerging in recent years. Most of the cryptocurrency studies were focused on price forecasting.

The rise of bitcoin from 1000 USD in January 2017 to 20,000 USD in January 2018 has attracted a lot of attention not only from the financial world, but also from ordinary people on the street. Recently, some papers have been published for price prediction and trading strategy development for bitcoin and other cryptocurrencies. Given the attention that the underlying technology has attracted, there is a great chance that some new studies will start appearing in the near future.

In the literature, \gls{dnn}, \gls{lstm}, \gls{gru}, \gls{rnn}, Classical methods (\gls{arma}, \gls{arima}, \gls{arch}, \gls{garch}, etc) were used for cryptocurrency price forecasting.  Table~\ref{table:cryptocurrency_price_prediction} tabulates the studies that utilize these methods. In \cite{Lopes_2018_thesis}, the author combined the opinion market and price prediction for cryptocurrency trading. Text mining combined with 2 models \gls{cnn} and \gls{lstm} were used to extract the opinion. Bitcoin, Litecoin, StockTwits were used as the dataset. \gls{ochlv} of prices, technical indicators, and sentiment analysis were used as the feature set. In \cite{McNally_2018}, the authors compared Bayesian optimized \gls{rnn}, \gls{lstm} and \gls{arima} to predict bitcoin price direction. Sensitivity, specificity, precision, accuracy, RMSE were used as the performance metrics.

\input{tables/table6xx.tex}

\subsection{Trend Forecasting}

Even though trend forecasting and price forecasting share the same input characteristics, some researchers prefer to predict the price direction of the asset instead of the actual price. This alters the nature of the problem from regression to classification and the corresponding performance metrics also change. However, it is worth to mention that these two approaches are not really different, the difference is in the interpretation of the output. 

In the literature, there were different methods for trend forecasting.  In this survey, we grouped the articles according to their feature set such as studies using only the raw time series data (only price data, \gls{ochlv}); studies using technical indicators \& price data \& fundamental data at the same time; studies using text mining techniques and studies using other various data. Table~\ref{table:trend_forecasting_1} tabulates the trend forecasting using only the raw time series data. 

\input{tables/table7xx_1.tex}

Different methods and models were used for trend forecasting. In Table~\ref{table:trend_forecasting_1}, these are divided into three sub-groups: \gls{ann}, \gls{dnn}, and \gls{ffnn} models; \gls{lstm}, \gls{rnn}, and Probabilistic \gls{nn} models; novel methods. \gls{ann}, \gls{dnn}, \gls{dfnn}, and  \gls{ffnn} methods were used in some of the studies.  In \cite{Das_2018_a}, \gls{nn} with the price data were used for prediction of the trend of \gls{sp500} stock indices. The authors of  \cite{Navon_2017} combined deep \gls{fnn} with a selective trading strategy unit to predict the next price.  The authors of \cite{Yang_2017} created an ensemble network of several Backpropagation and \gls{adam} models for trend prediction.

In the literature, \gls{lstm}, \gls{rnn}, \gls{pnn} methods with the raw time series data were also used for trend forecasting. In \cite{Saad_1998}, the authors compared  \gls{tdnn}, \gls{rnn} and \gls{pnn} for trend detection using 10 stocks from \gls{sp500}. The authors of \cite{persio_2017} compared 3 different \gls{rnn} models (basic \gls{rnn}, \gls{lstm}, \gls{gru}) to predict the movement of Google stock price. The authors of \cite{Hansson_2017} used \gls{lstm} (and other classical forecasting techniques)  to predict the trend of the stocks prices. In \cite{Shen_2018}, \gls{gru} and \gls{gru}-\gls{svm} models were used for the trend of \gls{hsi}, \gls{dax}, \gls{sp500} indices.

There were also novel methods that used only the raw time series price/index data in the literature. The author of \cite{Chen_2016_d} proposed a method that used \gls{cnn} with \gls{gaf}, \gls{mam}, Candlestick with converted image data. In \cite{Sezer_2019}, a novel method, \gls{cnn} with feature imaging was proposed for the prediction of the buy/sell/hold positions of the \glspl{etf}' prices and Dow30 stocks' prices. The authors of \cite{Zhou_2019} proposed a method that uses  \gls{emd2fnn} models to forecast the stock close prices' direction accurately. In \cite{Ausmees_2017}, \gls{dbn} with the price data were used for the prediction of the trend of 23 large cap stocks from the \gls{omx}30 index.

\input{tables/table7xx_2.tex}

In the literature, some of the studies used technical indicators \& price data \& fundamental data at the same time. Table~\ref{table:trend_forecasting_2} tabulates the trend forecasting papers using technical indicators, price data, fundamental data. In addition, these studies are clustered into three sub-groups: \gls{ann}, \gls{mlp}, \gls{dbn}, and \gls{rbm} models; \gls{lstm} and \gls{gru} models; novel methods. \gls{ann}, \gls{mlp}, \gls{dbn}, and \gls{rbm} methods were used with technical indicators, price data and fundamental data in some of the studies. In  \cite{Raza_2017}, several classical, \gls{ml} models and \gls{dbn} were compared for trend forecasting. In \cite{Sezer_2017}, technical analysis indicator's (\gls{rsi}) buy \& sell limits were optimized with \gls{ga} which was used for buy-sell signals. After optimization, \gls{dmlp} was also used for function approximation. The authors of \cite{Liang_2017} used technical analysis parameters, \gls{ochlv} of prices and \gls{rbm} for stock trend prediction.

Besides, \gls{lstm} and \gls{gru} methods with technical indicators \& price data \& fundamental data were also used in some of the papers. In \cite{Troiano_2018}, the crossover and \gls{macd} signals were used to predict the trend of the Dow 30 stocks prices. The authors of \cite{Nelson_2017} used \gls{lstm} for stock price movement estimation. The author of \cite{song_2018} used stock prices, technical analysis features and four different \gls{ml} Models (\gls{lstm}, \gls{gru}, \gls{svm} and \gls{xgboost}) to predict the trend of the stocks prices.

In addition, there were also novel and new methods that used \gls{cnn} with the price data and technical indicators. The authors of \cite{Gudelek_2017} converted the time series of price data to 2-dimensional images using technical analysis and classified them with deep \gls{cnn}. Similarly, the authors of \cite{Sezer_2018} also proposed a novel technique that converted financial time series data that consisted of technical analysis indicator outputs to 2-dimensional images and classified these images using \gls{cnn} to determine the  trading signals. The authors of \cite{Gunduz_2017} proposed a method that used \gls{cnn} with correlated features combined together to predict the trend of the stocks prices.

Besides, there were also studies that used text mining techniques in the literature. Table~\ref{table:trend_forecasting_3} tabulates the trend forecasting papers using text mining techniques. Different methods/models are represented within four sub-groups in that table: \gls{dnn}, \gls{dmlp}, and \gls{cnn} with text mining models; \gls{gru} model; \gls{lstm}, \gls{cnn}, and \gls{lstm}+\gls{cnn} models; novel methods. 
In the first group of studies, \gls{dnn}, \gls{dmlp}, \gls{cnn} with text mining were used for trend forecasting. In \cite{Huang_2016}, the authors used different models that included \gls{hmm}, \gls{dmlp} and \gls{cnn} using Twitter moods to predict the next days' move. In \cite{Peng_2016}, the authors used the combination of text mining and word embeddings to extract information from financial news and \gls{dnn} model for prediction of the stock trends.

Moreover, \gls{gru} methods with text mining techniques were also used for trend forecasting. The authors of \cite{Huynh_2017} used financial news from Reuters, Bloomberg and stock prices data and \gls{bi-gru} model to predict the stock movements in the future. The authors of  \cite{Dang_2018} used Stock2Vec and \gls{tgru} models to generate input data from financial news and stock prices. Then, they used the sign difference between the previous close and next open for the classification of the stock prices. The results were better than the state-of-the-art models.

\gls{lstm}, \gls{cnn} and \gls{lstm}+\gls{cnn} models were also used for trend forecasting. The authors of \cite{Verma_2017} combined news data with financial data to classify the stock price movement and assessed them with certain factors.  They used \gls{lstm} model as the \gls{nn} architecture. The authors of \cite{Pinheiro_2017} proposed a novel method that used character-based neural language model using financial news and \gls{lstm} for trend prediction.  In \cite{Prosky_2017}, sentiment/mood prediction and price prediction based on sentiment, price prediction with text mining and \gls{dl} models (\gls{lstm}, \gls{nn}, \gls{cnn}) were used for trend forecasting. The authors of \cite{Liu_2018} proposed a method that used two separate \gls{lstm} networks to construct an ensemble network. One of the \gls{lstm} models was used for word embeddings with word2Vec to create a matrix information as input to \gls{cnn}. The other one was used for price prediction using technical analysis features and stock prices.

In the literature, there were also novel and different methods to predict the trend of the time series data. In \cite{Yoshihara_2014}, the authors proposed a novel method that uses a combination of \gls{rbm}, \gls{dbn} and word embedding to create word vectors for \gls{rnn}-\gls{rbm}-\gls{dbn} network to predict the trend of stock prices. The authors of \cite{Shi_2018} proposed a novel method (called DeepClue) that visually interpretted text-based \gls{dl} models in predicting stock price movements. In their proposed method, financial news, charts and social media tweets were used together to predict the stock price movement. The authors of \cite{Zhang_2018} proposed a method that performed information fusion from several news and social media sources to predict the trend of the stocks. The authors of \cite{Hu_2018} proposed a novel method that used text mining techniques and Hybrid Attention Networks based on financial news for the forecast of the trend of stocks. The authors of \cite{Wang_2018_a} combined technical analysis and sentiment analysis of social media (related financial topics) and created \gls{drse} method for classification. The authors of \cite{MATSUBARA_2018} proposed a method that used \gls{dgm} with news articles using Paragraph Vector algorithm to create the input vector for the prediction of the trend of stocks. The authors of  \cite{Li_2018} implemented intraday stock price direction classification using financial news and stocks prices.

\input{tables/table7xx_3.tex}

Moreover, there were also studies that used different data variations in the literature. Table~\ref{table:trend_forecasting_4} tabulates the trend forecasting papers using these various data clustered into two sub-groups: \gls{lstm}, \gls{rnn}, \gls{gru} models; \gls{cnn} model.

\gls{lstm}, \gls{rnn}, \gls{gru} methods with various data representations were used in some trend forecasting papers. In \cite{Tsantekidis_2017}, the authors used the limit order book time series data and \gls{lstm} method for trend prediction. The authors of \cite{Sirignano_2018} proposed a novel method that used limit order book flow and history information for the determination of the stock movements using \gls{lstm}. The results of the proposed method were remarkably stationary. The authors of \cite{Chen_2018_e} used social media news, \gls{lda} features and \gls{rnn} model to predict the trend of the index price. The authors of \cite{Buczkowski_2017} proposed a novel method that used expert recommendations (Buy, Hold or Sell), emsemble of \gls{gru} and \gls{lstm} to predict the trend of the stocks prices.

\gls{cnn} models with different data representations were also used for trend prediction. In \cite{Tsantekidis_2017_a}, the authors used the last 100 entries from the limit order book to create images for the stock price prediction using \gls{cnn}. Using the limit order book data to create 2D matrix-like format with \gls{cnn} for predicting directional movement was innovative. In \cite{Doering_2017}, \gls{hft} microstructures forecasting with \gls{cnn} was implemented.

\input{tables/table7xx_4.tex}

%% file: tables/table1xx_1.tex

\begingroup
\footnotesize
\fontsize{7}{9}\selectfont
\begin{longtable}{
                p{0.03\linewidth}
                p{0.15\linewidth}
                p{0.08\linewidth}
                p{0.13\linewidth}
                p{0.05\linewidth}
                p{0.05\linewidth}
                p{0.10\linewidth}
                p{0.13\linewidth}
                p{0.08\linewidth}
                }

\caption{Stock Price Forecasting Using Only Raw Time Series Data }

\\
\hline
\textbf{Art.}                                               &
\textbf{Data Set}                                           &
\textbf{Period}                                             &
\textbf{Feature Set}                                        &
\textbf{Lag}                                                &
\textbf{Horizon}                                            &
\textbf{Method}                                             &
\textbf{Performance Criteria}                               &
\textbf{Env.}                                                    \\
\hline
\endhead                                    

\cite{Chong_2017}   & 38 stocks in \acrshort{kospi} & 2010-2014 & Lagged stock returns                                        & 50min& 5min     & \acrshort{dnn} & \acrshort{nmse}, \acrshort{rmse}, \acrshort{mae}, \acrshort{mi} & - \\ \hline
\cite{Chen_2015}    & China stock market, 3049 Stocks & 1990-2015 & \acrshort{ochlv}                                          & 30d  & 3d       & \acrshort{lstm} & Accuracy & Theano, Keras \\ \hline
\cite{Dezsi_2016}   & Daily returns of ‘BRD’ stock in Romanian Market & 2001-2016 & \acrshort{ochlv}                          & -    & 1d       & \acrshort{lstm}  & \acrshort{rmse}, \acrshort{mae} & Python, Theano \\ \hline
\cite{Samarawickrama_2017} & 297 listed companies of \acrshort{cse} & 2012-2013 & \acrshort{ochlv}                            & 2d   & 1d       & \acrshort{lstm}, \acrshort{srnn}, \acrshort{gru} & \acrshort{mad}, \acrshort{mape} & Keras \\ \hline
\cite{M_2018}       & 5 stock in \acrshort{nse} & 1997-2016 & \acrshort{ochlv}, Price data, turnover and number of trades.    & 200d & 1..10d   & \acrshort{lstm}, \acrshort{rnn}, \acrshort{cnn}, \acrshort{mlp} & \acrshort{mape} & - \\ \hline
\cite{Selvin_2017}  & Stocks of Infosys, TCS and CIPLA from \acrshort{nse} & 2014 & Price data          & -    & -        & \acrshort{rnn}, \acrshort{lstm} and \acrshort{cnn} & Accuracy & - \\ \hline
\cite{Lee_2018}     & 10 stocks in \acrshort{sp500} & 1997-2016 & \acrshort{ochlv}, Price data                                     & 36m  & 1m       & \acrshort{rnn}, \acrshort{lstm}, \acrshort{gru} & Accuracy, Monthly return & Keras, Tensorflow \\ \hline
\cite{Li_2017}      & Stocks data from \acrshort{sp500} & 2011-2016 & \acrshort{ochlv}                                             & 1d   & 1d       & \acrshort{dbn} & \acrshort{mse}, \acrshort{norm-rmse}, \acrshort{mae} & - \\ \hline
\cite{Chen_2018}    & High-frequency transaction data of the \acrshort{csi}300 futures & 2017 & Price data                    & -    & 1min & \acrshort{dnn}, \acrshort{elm}, \acrshort{rbf} & \acrshort{rmse}, \acrshort{mape}, Accuracy & Matlab \\ \hline
\cite{Krauss_2017}  & Stocks in the \acrshort{sp500} & 1990-2015 & Price data                                                      & 240d    & 1d & \acrshort{dnn}, \acrshort{gbt}, \acrshort{rf} & Mean return, \acrshort{mdd}, Calmar ratio & H2O \\ \hline
\cite{Chandra_2016} & ACI Worldwide, Staples, and Seagate in \acrshort{nasdaq} & 2006-2010 & Daily closing prices  & 17d    & 1d  & \acrshort{rnn}, \acrshort{ann} & \acrshort{rmse} & - \\ \hline
\cite{Liu_2017}     & Chinese Stocks & 2007-2017 & \acrshort{ochlv}                                                           & 30d    & 1..5d & \acrshort{cnn} + \acrshort{lstm} & Annualized Return, Mxm Retracement & Python  \\ \hline
\cite{Heaton_2016}  & 20 stocks in \acrshort{sp500}  & 2010-2015 & Price data                                                      & -    & - & \acrshort{ae} + \acrshort{lstm} & Weekly Returns & - \\ \hline
\cite{Batres_2015}  & \acrshort{sp500} & 1985-2006 & Monthly and daily log-returns                                                 & *    & 1d & \acrshort{dbn}+\acrshort{mlp} & Validation, Test Error & Theano, Python, Matlab \\ \hline
\cite{Yuan_2018}    & 12 stocks from \acrshort{sse} Composite Index & 2000-2017 & \acrshort{ochlv}                            & 60d    & 1..7d & \acrshort{dwnn} & \acrshort{mse}  & Tensorflow \\ \hline
\cite{Zhang_2017}   & 50 stocks from \acrshort{nyse} & 2007-2016 & Price data                                                 & -    & 1d, 3d, 5d & \acrshort{sfm} & \acrshort{mse} & - \\ \hline

\label{table:stock_price_forecasting_1} 
\end{longtable}
\endgroup

%% file: tables/table1xx_2.tex
\begingroup
\footnotesize
\fontsize{7}{9}\selectfont
\begin{longtable}{
                p{0.03\linewidth}
                p{0.15\linewidth}
                p{0.08\linewidth}
                p{0.13\linewidth}
                p{0.05\linewidth}
                p{0.05\linewidth}
                p{0.10\linewidth}
                p{0.13\linewidth}
                p{0.08\linewidth}
                }

\caption{Stock Price Forecasting Using Various Data}

\\
\hline
\textbf{Art.}                                                 &
\textbf{Data Set}                                            &
\textbf{Period}                                               &
\textbf{Feature Set}                                             &
\textbf{Lag}                                                &
\textbf{Horizon}                                            &
\textbf{Method}                                                  &
\textbf{Performance Criteria}                                        &
\textbf{Env.}                                                    \\
\hline
\endhead                                    

\cite{Abe_2018} & Japan Index constituents from WorldScope & 1990-2016 & 25 Fundamental Features & 10d & 1d & \acrshort{dnn} & Correlation, Accuracy, \acrshort{mse} & Tensorflow \\ \hline
\cite{Feng_2018} & Return of \acrshort{sp500} & 1926-2016 & Fundamental Features: & - & 1s & \acrshort{dnn} & \acrshort{mspe} & Tensorflow \\ \hline
\cite{Fan_2014} & U.S. low-level disaggregated macroeconomic time series & 1959-2008 & GDP,  Unemployment rate, Inventories, etc. & - & - & \acrshort{dnn} & \acrshort{r-sq} & - \\ \hline
\cite{Kraus_2017} & \acrshort{cdax} stock market data & 2010-2013 & Financial news,  stock market data & 20d & 1d & \acrshort{lstm} & \acrshort{mse}, \acrshort{rmse}, \acrshort{mae}, Accuracy,  \acrshort{auc} & TensorFlow, Theano, Python, Scikit-Learn \\ \hline
\cite{Minami_2018} & Stock of Tsugami Corporation & 2013 & Price data & - & - & \acrshort{lstm} & \acrshort{rmse} & Keras, Tensorflow \\ \hline
\cite{Zhang_2018_a} & Stocks in China’s A-share & 2006-2007 & 11 technical indicators & - & 1d & \acrshort{lstm} & \acrshort{areturn}, \acrshort{ir}, \acrshort{ic} & - \\ \hline
\cite{Zhuge_2017} & SCI prices & 2008-2015 & \acrshort{ochl} of change rate, price & 7d & - & EmotionalAnalysis + \acrshort{lstm} & \acrshort{mse} & - \\ \hline
\cite{Akita_2016} & 10 stocks in Nikkei 225 and news & 2001-2008 & Textual information and Stock prices & 10d & - & Paragraph Vector + \acrshort{lstm} & Profit & - \\ \hline
\cite{Ozbayoglu_2007} & TKC stock in \acrshort{nyse} and QQQQ ETF & 1999-2006 & Technical indicators, Price & 50d & 1d & \acrshort{rnn} (Jordan-Elman) & Profit, \acrshort{mse} & Java \\ \hline
\cite{Khare_2017} & 10 Stocks in \acrshort{nyse} & - & Price data, Technical indicators & 20min & 1min & \acrshort{lstm}, \acrshort{mlp} & \acrshort{rmse} & - \\ \hline
\cite{Zhou_2018} & 42 stocks in China's \acrshort{sse} & 2016 & \acrshort{ochlv}, Technical Indicators & 242min & 1min & \acrshort{gan} (\acrshort{lstm}, \acrshort{cnn}) & \acrshort{rmsre}, \acrshort{dpa}, \acrshort{gan}-F, \acrshort{gan}-D & - \\ \hline
\cite{Singh_2016} & Google’s daily stock data & 2004-2015 & \acrshort{ochlv}, Technical indicators & 20d & 1d & $(2D)^2$ \acrshort{pca} + \acrshort{dnn} & \acrshort{smape}, \acrshort{pcd}, \acrshort{mape}, \acrshort{rmse}, \acrshort{hr}, \acrshort{tr}, \acrshort{r-sq} & R, Matlab \\ \hline
\cite{Karaoglu_2017} & GarantiBank in \acrshort{bist}, Turkey & 2016 & \acrshort{ochlv}, Volatility, etc. & - & - & \acrshort{plr}, Graves \acrshort{lstm} & \acrshort{mse}, \acrshort{rmse}, \acrshort{mae}, \acrshort{rse}, \acrshort{r-sq} & Spark \\ \hline
\cite{Zhou_2018_a} & Stocks in \acrshort{nyse}, \acrshort{amex}, \acrshort{nasdaq}, \acrshort{taq} intraday trade & 1993-2017 & Price, 15 firm characteristics & 80d & 1d & \acrshort{lstm}+\acrshort{mlp} & Monthly return, \acrshort{sr} & Python,Keras, Tensorflow in AWS \\ \hline
\cite{Abroyan_2017} & Private brokerage company’s real data of risky transactions & - & 250 features: order details, etc. & - & - & \acrshort{cnn}, \acrshort{lstm} & F1-Score & Keras, Tensorflow \\ \hline
\cite{GooglePatent} & Fundamental and Technical Data, Economic Data & - & Fundamental , technical and market information & - & - & \acrshort{cnn} & - & - \\ \hline
\cite{Tran_2017} & The LOB of 5 stocks of Finnish Stock Market & 2010 & FI-2010 dataset: bid/ask and volume & - & * & \acrshort{wmtr}, \acrshort{mda} & Accuracy, Precision, Recall, F1-Score & - \\ \hline
\cite{Feng_2018_a} & Returns in \acrshort{nyse}, \acrshort{amex}, \acrshort{nasdaq} & 1975-2017 & 57 firm characteristics & * & - & Fama-French n-factor model \acrshort{dl} & \acrshort{r-sq}, \acrshort{rmse} & Tensorflow \\ \hline

\label{table:stock_price_forecasting_2}
\end{longtable}
\endgroup

%% file: tables/table1xx_3.tex
\begingroup
\footnotesize
\fontsize{7}{9}\selectfont
\begin{longtable}{
                p{0.03\linewidth}
                p{0.15\linewidth}
                p{0.08\linewidth}
                p{0.13\linewidth}
                p{0.05\linewidth}
                p{0.05\linewidth}
                p{0.10\linewidth}
                p{0.13\linewidth}
                p{0.08\linewidth}
                }

\caption{Stock Price Forecasting Using Text Mining Techniques for Feature Extraction}
 
\\
\hline
\textbf{Art.}                                                 &
\textbf{Data Set}                                            &
\textbf{Period}                                               &
\textbf{Feature Set}                                             &
\textbf{Lag}                                                &
\textbf{Horizon}                                            &
\textbf{Method}                                                  &
\textbf{Performance Criteria}                                        &
\textbf{Env.}                                                    \\
\hline
\endhead

\cite{Ding_2015} & \acrshort{sp500} Index,  15 stocks in \acrshort{sp500} & 2006-2013 & News from Reuters and Bloomberg & - & - & \acrshort{cnn} & Accuracy, \acrshort{mcc} & - \\ \hline
\cite{Vargas_2017} & \acrshort{sp500} index news from Reuters & 2006-2013 & Financial news titles, Technical indicators & 1d & 1d & \acrshort{rcnn} & Accuracy & - \\ \hline
\cite{Lee_2017_b} & \acrshort{twse} index, 4 stocks in \acrshort{twse} & 2001-2017 & Technical indicators, Price data, News & 15d & - & \acrshort{cnn} + \acrshort{lstm} & \acrshort{rmse}, Profit & Keras, Python, TALIB \\ \hline
\cite{Iwasaki_2018} & Analyst reports on the TSE and Osaka Exchange & 2016-2018 & Text & - & - & \acrshort{lstm}, \acrshort{cnn}, \acrshort{bi-lstm} & Accuracy, R-squared & R, Python, MeCab \\ \hline
\cite{Das_2018} & Stocks of Google, Microsoft and Apple & 2016-2017 & Twitter sentiment and stock prices & - & - & \acrshort{rnn} & - & Spark, Flume, Twitter API, \\ \hline
\cite{Jiahong_Li_2017} & Stocks  of \acrshort{csi}300 index, \acrshort{ochlv} of \acrshort{csi}300 index & 2009-2014 & Sentiment Posts, Price data & 1d & 1d & Naive Bayes + \acrshort{lstm} & Precision, Recall, F1-score, Accuracy & Python, Keras \\ \hline
\cite{Zhongshengz_2018} & SCI prices & 2013-2016 & Text data and Price data & 7d & 1d & \acrshort{lstm} & Accuracy, F1-Measure & Python, Keras \\ \hline
\cite{Nascimento_2015} & Stocks from \acrshort{sp500} & 2006-2013 & Text (news) and Price data & 7d & 1d & \acrshort{lar}+News, \acrshort{rf}+News & \acrshort{mape}, \acrshort{rmse} & - \\ \hline
\cite{Han_2018} & News from Sina.com, ACE2005 Chinese corpus & 2012-2016 & A set of news text & - & - & Their unique algorithm & Precision, Recall, F1-score & - \\ \hline

\label{table:stock_price_forecasting_3}
\end{longtable}
\endgroup

%% file: tables/table2xx_1.tex
\begingroup
\footnotesize
\fontsize{7}{9}\selectfont
\begin{longtable}{
                p{0.03\linewidth}
                p{0.15\linewidth}
                p{0.08\linewidth}
                p{0.13\linewidth}
                p{0.05\linewidth}
                p{0.05\linewidth}
                p{0.10\linewidth}
                p{0.13\linewidth}
                p{0.08\linewidth}
                }
                
\caption{Index Forecasting Using Only Raw Time Series Data}\\

\\
\hline

\textbf{Art.}                                                          & 
\textbf{Data Set}                                                       & 
\textbf{Period}                                                    & 
\textbf{Feature Set}                                                    & 
\textbf{Lag}                                                &
\textbf{Horizon}                                            &
\textbf{Method}                                                 & 
\textbf{Performance Criteria}                                           & 
\textbf{Env.}                                                      \\ 
\hline
\endhead

\cite{Parida_2016} & \acrshort{sp500}, Nikkei225, USD Exchanges & 2011-2015 & Index data & - & 1d, 5d, 7d, 10d & \acrshort{lrnfis} with Firefly-Harmony Search & \acrshort{rmse}, \acrshort{mape}, \acrshort{mae} & - \\ \hline
\cite{Fischer_2018} & \acrshort{sp500} Index & 1989-2005 & Index data, Volume & 240d & 1d & \acrshort{lstm} & Return, \acrshort{std}, \acrshort{sr}, Accuracy & Python, TensorFlow, Keras, R, H2O \\ \hline
\cite{borovykh_2018} & \acrshort{sp500}, \acrshort{vix} & 2005-2016 & Index data & * & 1d & uWN, cWN & \acrshort{mase}, \acrshort{hit}, \acrshort{rmse} & - \\ \hline
\cite{Althelaya_2018} & \acrshort{sp500} Index & 2010-2017 & Index data & 10d & 1d, 30d & Stacked \acrshort{lstm}, \acrshort{bi-lstm} & \acrshort{mae}, \acrshort{rmse}, R-squared & Python, Keras, Tensorflow \\ \hline
\cite{Jeong_2019} & \acrshort{sp500}, \acrshort{kospi}, \acrshort{hsi}, and EuroStoxx50 & 1987-2017 & 200-days stock price & 200d & 1d & Deep Q-Learning and \acrshort{dnn} & Total profit,  Correlation & - \\ \hline
\cite{Baek_2018} & \acrshort{sp500}, \acrshort{kospi}200, 10-stocks & 2000-2017 & Index data & 20d & 1d & ModAugNet: \acrshort{lstm} & \acrshort{mse}, \acrshort{mape}, \acrshort{mae} & Keras \\ \hline
\cite{Hansson_2017} & \acrshort{sp500}, Bovespa50, \acrshort{omx}30 & 2009-2017 & Autoregressive part of the time series & - & 1d & \acrshort{lstm} & \acrshort{mse}, Accuracy & Tensorflow, Keras, R \\ \hline
\cite{Elliot_2017} & \acrshort{sp500} & 2000-2017 & Index data & - & 1..4d, 1w, 1..3m & \acrshort{glm}, \acrshort{lstm}+\acrshort{rnn} & \acrshort{mae}, \acrshort{rmse} & Python \\ \hline
\cite{Namini_2018} & Nikkei225,  \acrshort{ixic},  \acrshort{hsi},  \acrshort{gspc}, \acrshort{djia} & 1985-2018 & \acrshort{ochlv} & 5d & 1d & \acrshort{lstm} & \acrshort{rmse} & Python, Keras, Theano \\ \hline
\cite{Zhang_2015} & \acrshort{djia} & - & Index data & - & - & Genetic Deep Neural Network & \acrshort{mse} & Java \\ \hline
\cite{Bekiros_2013} & Log returns of the \acrshort{djia} & 1971-2002 & Index data & 20d & 1d & \acrshort{rnn} & \acrshort{tr}, sign rate, PT/HM test, \acrshort{msfe}, \acrshort{sr}, profit & - \\ \hline
\cite{Pang_2018} & Shanghai A-shares composite index, \acrshort{szse} & 2006-2016 & \acrshort{ochlv} & 10d & - & Embedded layer + \acrshort{lstm} & Accuracy, \acrshort{mse} & Python, Matlab, Theano \\ \hline
\cite{Deng_2017} & 300 stocks from \acrshort{szse}, Commodity & 2014-2015 & Index data & - & - & \acrshort{fddr}, \acrshort{dnn} + \acrshort{rl} & Profit, return, \acrshort{sr}, profit-loss curves & Keras \\ \hline
\cite{Yang_2017} & Shanghai composite index and \acrshort{szse} & 1990-2016 & \acrshort{ochlv} & 20d & 1d & Ensembles of \acrshort{ann} & Accuracy & - \\ \hline
\cite{Lachiheb_2018} & \acrshort{tunindex} & 2013-2017 & Log returns of index data & - & 5min & \acrshort{dnn} with hierarchical input & Accuracy, \acrshort{mse} & Java \\ \hline
\cite{Yong_2017} & Singapore Stock Market Index & 2010-2017 & \acrshort{ochl} of last 10 days of index & 10d & 3d & Feed-forward \acrshort{dnn} & \acrshort{rmse}, \acrshort{mape}, Profit, \acrshort{sr} & - \\ \hline
\cite{Yumlu_2005} & \acrshort{bist} & 1990-2002 & Index data & 7d & 1d & \acrshort{mlp}, \acrshort{rnn}, \acrshort{moe} & \acrshort{hit}, positive/negative \acrshort{hit},  \acrshort{mse}, \acrshort{mae} & - \\ \hline
\cite{Yan_2017} & SCI & 2012-2017 & \acrshort{ochlv}, Index data & - & 1..10d & Wavelet + \acrshort{lstm} & \acrshort{mape}, theil unequal coefficient & - \\ \hline
\cite{Takahashi_2017} & \acrshort{sp500} & 1950-2016 & Index data & 15d & 1d & \acrshort{lstm} & \acrshort{rmse} & Keras \\ \hline
\cite{Bildirici_2010} & \acrshort{ise}100 & 1987-2008 & Index data & - & 2d, 4d, 8d, 12d, 18d & \acrshort{tar}-\acrshort{vec}-\acrshort{mlp}, \acrshort{tar}-\acrshort{vec}-\acrshort{rbf}, \acrshort{tar}-\acrshort{vec}-\acrshort{rhe} & \acrshort{rmse} & - \\ \hline
\cite{Psaradellis_2016} & \acrshort{vix}, \acrshort{vxn}, \acrshort{vxd} & 2002-2014 & First five autoregressive lags & 5d & 1d,  22d & \acrshort{har-gasvr} & \acrshort{mae}, \acrshort{rmse} & - \\ \hline

\label{table:index_forecasting_1}
\end{longtable}
\endgroup

%% file: tables/table2xx_2.tex
\begingroup
\footnotesize
\fontsize{7}{9}\selectfont
\begin{longtable}{
                p{0.03\linewidth}
                p{0.15\linewidth}
                p{0.08\linewidth}
                p{0.13\linewidth}
                p{0.05\linewidth}
                p{0.05\linewidth}
                p{0.10\linewidth}
                p{0.13\linewidth}
                p{0.08\linewidth}
                }
                
\caption{Index Forecasting Using Various Data}\\

\\
\hline

\textbf{Art.}                                                          & 
\textbf{Data Set}                                                       & 
\textbf{Period}                                                    & 
\textbf{Feature Set}                                                    & 
\textbf{Lag}                                                &
\textbf{Horizon}                                            &
\textbf{Method}                                                 & 
\textbf{Performance Criteria}                                           & 
\textbf{Env.}                                                      \\ 
\hline
\endhead

\cite{Ding_2015} & \acrshort{sp500} Index,  15 stocks in \acrshort{sp500} & 2006-2013 & News from Reuters and Bloomberg & - & - & \acrshort{cnn} & Accuracy, \acrshort{mcc} & - \\ \hline
\cite{Lee_2017_b} & \acrshort{twse} index, 4 stocks in \acrshort{twse} & 2001-2017 & Technical indicators, Index data, News & 15d & - & \acrshort{cnn} + \acrshort{lstm} & \acrshort{rmse}, Profit & Keras, Python, \acrshort{talib} \\ \hline
\cite{Bao_2017} & \acrshort{csi}300, \acrshort{nifty}50, \acrshort{hsi}, \acrshort{nikkei}225, \acrshort{sp500}, \acrshort{djia} & 2010-2016 & \acrshort{ochlv}, Technical Indicators & - & 1d & \acrshort{wt}, Stacked autoencoders, \acrshort{lstm} & \acrshort{mape}, Correlation coefficient, \acrshort{theil-u} & - \\ \hline
\cite{Widegren_2017} & FTSE100, OMXS 30, SP500, Commodity, Forex & 1993-2017 & Technical indicators & 60d & 1d & \acrshort{dnn}, \acrshort{rnn} & Accuracy, p-value & - \\ \hline
\cite{Dingli_2017} & \acrshort{sp500}, \acrshort{dow30}, \acrshort{nasdaq}100, Commodity, Forex, Bitcoin & 2003-2016 & Index data, Technical indicators & - & 1w, 1m & \acrshort{cnn} & Accuracy & Tensorflow \\ \hline
\cite{Rout_2017} & \acrshort{bse}, \acrshort{sp500} & 2004-2012 & Index data, technical indicators & 5d & 1d..1m & \acrshort{pso}, \acrshort{hmrpso}, \acrshort{de}, \acrshort{rceflann} & \acrshort{rmse}, \acrshort{mape} & - \\ \hline
\cite{Li_2017a} & \acrshort{hsi}, \acrshort{sse}, \acrshort{szse}, \acrshort{taiex}, \acrshort{nikkei}, \acrshort{kospi} & 2010-2016 & Index data, volume, technical indicators & 2d..512d & 1d & \acrshort{lstm} with wavelet denoising & Accuracy, \acrshort{mape} & - \\ \hline
\cite{Hsieh_2011} & \acrshort{djia}, \acrshort{ftse}, \acrshort{nikkei}, \acrshort{taiex} & 1997-2008 & \acrshort{ochlv}, Technical indicators & 26d & 1d & \acrshort{rnn} & \acrshort{rmse}, \acrshort{mae}, \acrshort{mape}, \acrshort{theil-u} & C \\ \hline
\cite{Chen_2016} & Hedge fund monthly return data & 1996-2015 & Return, \acrshort{sr}, \acrshort{std}, Skewness, Kurtosis, Omega ratio, Fund alpha & 12m & 3m, 6m, 12m & \acrshort{dnn} & Sharpe ratio, Annual return, Cum. return & - \\ \hline
\cite{Mourelatos_2018} & Stock of National Bank of Greece (ETE). & 2009-2014 & \acrshort{ftse}100, \acrshort{djia}, \acrshort{gdax}, \acrshort{nikkei}225, EUR/USD, Gold & 1d, 2d, 5d, 10d & 1d & \acrshort{gasvr}, \acrshort{lstm} & Return, volatility, \acrshort{sr}, Accuracy & Tensorflow \\ \hline
\cite{Chen_2018_f} & Daily return ratio of \acrshort{hs}300 index & 2004-2018 & \acrshort{ochlv},  Technical indicators & - & - & Market Vector + Tech. ind. + \acrshort{lstm} + Attention & \acrshort{mse}, \acrshort{mae} & Python, Tensorflow \\ \hline
\cite{Si_2017} & Chinese stock-IF-IH-IC contract & 2016-2017 & Decisions for index change & 240min & 1min & \acrshort{modrl}+\acrshort{lstm} & Profit and loss, \acrshort{sr} & - \\ \hline
\cite{Chen_2018_e} & \acrshort{hs}300 & 2015-2017 & Social media news, Index data & 1d & 1d & \acrshort{rnn}-Boost with \acrshort{lda} & Accuracy, \acrshort{mae}, \acrshort{mape}, \acrshort{rmse} & Python, Scikit-learn \\ \hline

\label{table:index_forecasting_2}
\end{longtable}
\endgroup

%% file: tables/table3xx.tex
\begingroup
\footnotesize
\fontsize{7}{9}\selectfont
\begin{longtable}{
                p{0.03\linewidth}
                p{0.15\linewidth}
                p{0.08\linewidth}
                p{0.13\linewidth}
                p{0.05\linewidth}
                p{0.05\linewidth}
                p{0.10\linewidth}
                p{0.13\linewidth}
                p{0.08\linewidth}
                }
                
\caption{Commodity Price Forecasting}\\

\\
\hline

\textbf{Art.}   &
\textbf{Data Set}                                                       & 
\textbf{Period}                                                    & 
\textbf{Feature Set}                                                    & 
\textbf{Lag}                                                &
\textbf{Horizon}                                            &
\textbf{Method}                                                 & 
\textbf{Performance Criteria}                                           & 
\textbf{Env.}                                                      \\ 
\hline
\endhead

\cite{Dingli_2017} & \acrshort{sp500}, \acrshort{dow30}, \acrshort{nasdaq}100, Commodity, Forex, Bitcoin & 2003-2016 & Price data, Technical indicators & - & 1w, 1m & \acrshort{cnn} & Accuracy & Tensorflow \\ \hline
\cite{Dixon_2016} & Commodity, FX future, \acrshort{etf} & 1991-2014 & Price Data & 100*5min & 5min & \acrshort{dnn} & \acrshort{sr}, capability ratio, return & C++, Python \\ \hline
\cite{Widegren_2017} & \acrshort{ftse}100, \acrshort{omx}30, \acrshort{sp500}, Commodity, Forex & 1993-2017 & Technical indicators & 60d & 1d & \acrshort{dnn}, \acrshort{rnn} & Accuracy, p-value & - \\ \hline
\cite{S_nchez_Lasheras_2015} & Copper prices from \acrshort{nymex} & 2002-2014 & Price data & - & - & Elman \acrshort{rnn} & \acrshort{rmse} & R \\ \hline
\cite{Zhao_2017} & \acrshort{wti} crude oil price & 1986-2016 & Price data & 1m & 1m & \acrshort{sdae}, Bootstrap aggregation & Accuracy, \acrshort{mape}, \acrshort{rmse} & Matlab \\ \hline
\cite{Chen_2017_d} & \acrshort{wti} Crude Oil Prices & 2007-2017 & Price data & - & - & \acrshort{arma} + \acrshort{dbn}, \acrshort{rw} + \acrshort{lstm} & \acrshort{mse} & Python, Keras, Tensorflow \\ \hline
\cite{Deng_2017} & 300 stocks from \acrshort{szse}, Commodity & 2014-2015 & Price data & - & - & \acrshort{fddr}, \acrshort{dnn} + \acrshort{rl} & Profit, return, \acrshort{sr}, profit-loss curves & Keras \\ \hline

\label{table:commodity_price_forecasting}
\end{longtable}
\endgroup

%% file: tables/table4xx.tex
\begingroup
\fontsize{7}{9}\selectfont
\begin{longtable}{
                p{0.03\linewidth}
                p{0.15\linewidth}
                p{0.08\linewidth}
                p{0.13\linewidth}
                p{0.05\linewidth}
                p{0.05\linewidth}
                p{0.10\linewidth}
                p{0.13\linewidth}
                p{0.08\linewidth}
                }
                
\caption{Volatility Forecasting}\\

\\
\hline

\textbf{Art.}                                                          & 
\textbf{Data Set}                                                       & 
\textbf{Period}                                                    & 
\textbf{Feature Set}                                                    & 
\textbf{Lag}                                                &
\textbf{Horizon}                                            &
\textbf{Method}                                                 & 
\textbf{Performance Criteria}                                           & 
\textbf{Env.}                                                      \\ 
\hline
\endhead

\cite{Doering_2017} & London Stock Exchange & 2007-2008 & Limit order book state, trades, buy/sell orders, order deletions & - & - & \acrshort{cnn} & Accuracy, kappa & Caffe \\ \hline
\cite{Tino_2001} & \acrshort{dax}, \acrshort{ftse}100, call/put options & 1991-1998 & Price data & * & * & \acrshort{mm}, \acrshort{rnn} & Ewa-measure, iv, daily profits' mean and std & - \\ \hline
\cite{Xiong_2015} & \acrshort{sp500} & 2004-2015 & Price data, 25 Google Domestic trend dimensions & - & 1d & \acrshort{lstm} & \acrshort{mape}, \acrshort{rmse} & - \\ \hline
\cite{Zhou_2018_b} & \acrshort{csi} 300, 28 words of the daily search volume based on Baidu & 2006-2017 & Price data and text & 5d & 5d & \acrshort{lstm} & \acrshort{mse}, \acrshort{mape} & Python, Keras \\ \hline
\cite{Kim_2018} & \acrshort{kospi}200, Korea Treasury Bond interest rate, AA-grade corporate bond interest rate, gold, crude oil & 2001-2011 & Price data & 22d & 1d & \acrshort{lstm} + \acrshort{garch} & \acrshort{mae}, \acrshort{mse}, \acrshort{hmae}, \acrshort{hmse} & - \\ \hline
\cite{Nikolaev_2013} & DEM/GBP exchange rate & - & Returns & - & - & \acrshort{rmdn-garch} & \acrshort{nmse}, \acrshort{nmae}, \acrshort{hr}, \acrshort{whr} & - \\ \hline
\cite{Psaradellis_2016} & \acrshort{vix}, \acrshort{vxn}, \acrshort{vxd} & 2002-2014 & First five autoregressive lags & 5d & 1d,  22d & \acrshort{har-gasvr} & \acrshort{mae}, \acrshort{rmse} & - \\ \hline

\label{table:volatility_forecasting}
\end{longtable}
\endgroup

%% file: tables/table5xx.tex
\begingroup
\footnotesize
\fontsize{7}{9}\selectfont
\begin{longtable}{
                p{0.03\linewidth}
                p{0.15\linewidth}
                p{0.08\linewidth}
                p{0.13\linewidth}
                p{0.05\linewidth}
                p{0.05\linewidth}
                p{0.10\linewidth}
                p{0.13\linewidth}
                p{0.08\linewidth}
                }
                
\caption{Forex Price Forecasting}\\

\\
\hline

\textbf{Art.}                                                          & 
\textbf{Data Set}                                                       & 
\textbf{Period}                                                    & 
\textbf{Feature Set}                                                    &
\textbf{Lag}                                                &
\textbf{Horizon}                                            &
\textbf{Method}                                                 & 
\textbf{Performance Criteria}                                           & 
\textbf{Env.}                                                      \\ 
\hline
\endhead

\cite{Zhang_2014} & EUR/USD, GBP/USD & 2009-2012 & Price data & * & 1d & \acrshort{cdbn-fg} & Profit & - \\ \hline
\cite{Chao_2011} & GBP/USD, INR/USD & 1976-2003 & Price data & 10w & 1w & \acrshort{dbn} & \acrshort{rmse}, \acrshort{mae}, \acrshort{mape}, \acrshort{da}, \acrshort{pcc} & - \\ \hline
\cite{Zheng_2017} & CNY/USD,INR/USD & 1997-2016 & Price data & - & 1w & \acrshort{dbn} & \acrshort{mape}, R-squared & - \\ \hline
\cite{Shen_2015} & GBP/USD, BRL/USD, INR/USD & 1976-2003 & Price data & 10w & 1w & \acrshort{dbn} + \acrshort{rbm} & \acrshort{rmse}, \acrshort{mae}, \acrshort{mape}, accuracy, \acrshort{pcc} & - \\ \hline
\cite{Shen_2016} & Combination of USD, GBP, EUR, JPY, AUD, CAD, CHF & 2009-2016 & Price data & - & - & Stacked \acrshort{ae} + \acrshort{svr} & \acrshort{mae}, \acrshort{mse}, \acrshort{rmse} & Matlab \\ \hline
\cite{Dixon_2016} & Commodity, FX future, \acrshort{etf} & 1991-2014 & Price Data & 100*5min & 5min & \acrshort{dnn} & \acrshort{sr}, capability ratio, return & C++, Python \\ \hline
\cite{Widegren_2017} & \acrshort{ftse}100, \acrshort{omx}30, \acrshort{sp500}, Commodity, Forex & 1993-2017 & Technical indicators & 60d & 1d & \acrshort{dnn}, \acrshort{rnn} & Accuracy, p-value & - \\ \hline
\cite{Sermpinis_2012_a} & EUR/USD & 2001-2010 & Close data & 11d & 1d & \acrshort{rnn} and more & \acrshort{mae}, \acrshort{mape}, \acrshort{rmse}, \acrshort{theil-u} & - \\ \hline
\cite{Sermpinis_2012} & EUR/USD & 2002-2010 & Price data & 13d & 1d & \acrshort{rnn}, \acrshort{mlp}, \acrshort{psn} & \acrshort{mae}, \acrshort{mape}, \acrshort{rmse}, \acrshort{theil-u} & - \\ \hline
\cite{Sermpinis_2014} & EUR/USD, EUR/GBP, EUR/JPY,  EUR/CHF & 1999-2012 & Price data & 12d & 1d & \acrshort{rnn}, \acrshort{mlp}, \acrshort{psn} & \acrshort{mae}, \acrshort{mape}, \acrshort{rmse}, \acrshort{theil-u} & - \\ \hline
\cite{SUN_2009} & RMB against USD, EUR, JPY, HKD & 2006-2008 & Price data & 10d & 1d & \acrshort{rnn}, \acrshort{ann} & \acrshort{rmse}, \acrshort{mae}, \acrshort{mse} & - \\ \hline
\cite{Maknickien__2013} & EUR/USD, EUR/JPY, USD/JPY, EUR/CHF, XAU/USD, XAG/USD, QM, QG & 2011-2012 & Price data & - & - & Evolino \acrshort{rnn} & Correlation between predicted, real values & - \\ \hline
\cite{Maknickiene_2014} & USD/JPY & 2009-2010 & Price data, Gold & - & 5d & EVOLINO \acrshort{rnn} + orthogonal input data & \acrshort{rmse} & - \\ \hline
\cite{persio_2016} & \acrshort{sp500}, EUR/USD & 1950-2016 & Price data & 30d, 30d*min & 1d, 1min & Wavelet+\acrshort{cnn} & Accuracy, log-loss & Keras \\ \hline
\cite{Korczak_2017} & USD/GBP, \acrshort{sp500}, \acrshort{ftse}100, oil, gold & 2016 & Price data & - & 5min & \acrshort{ae} + \acrshort{cnn} & \acrshort{sr}, \% volatility, avg return/trans, rate of return & H2O \\ \hline
\cite{Bildirici_2010} & \acrshort{ise}100, TRY/USD & 1987-2008 & Price data & - & 2d, 4d, 8d, 12d, 18d & \acrshort{tar}-\acrshort{vec}-\acrshort{mlp}, \acrshort{tar}-\acrshort{vec}-\acrshort{rbf}, \acrshort{tar}-\acrshort{vec}-\acrshort{rhe} & \acrshort{rmse} & - \\ \hline
\cite{Nikolaev_2013} & DEM/GBP exchange rate & - & Returns & - & - & \acrshort{rmdn}-\acrshort{garch} & \acrshort{nmse}, \acrshort{nmae}, \acrshort{hr}, \acrshort{whr} & - \\ \hline
\cite{Parida_2016} & \acrshort{sp500}, \acrshort{nikkei}225, USD Exchanges & 2011-2015 & Price data & - & 1d, 5d, 7d, 10d & \acrshort{lrnfis} with \acrshort{fhso} & \acrshort{rmse}, \acrshort{mape}, \acrshort{mae} & - \\ \hline

\label{table:forex_price_forecasting}
\end{longtable}
\endgroup

%% file: tables/table6xx.tex
\begingroup
\footnotesize
\fontsize{7}{9}\selectfont
\begin{longtable}{
                p{0.03\linewidth}
                p{0.15\linewidth}
                p{0.08\linewidth}
                p{0.13\linewidth}
                p{0.05\linewidth}
                p{0.05\linewidth}
                p{0.10\linewidth}
                p{0.13\linewidth}
                p{0.08\linewidth}
                }
                
\caption{Cryptocurrency Price Prediction}\\

\\
\hline
\textbf{Art.}                                                          & 
\textbf{Data Set}                                                       & 
\textbf{Period}                                                    & 
\textbf{Feature Set}                                                    & 
\textbf{Lag}                                                &
\textbf{Horizon}                                            &
\textbf{Method}                                                 & 
\textbf{Performance Criteria}                                           & 
\textbf{Env.}                                                      \\ 
\hline
\endhead

\cite{Lopes_2018_thesis} & Bitcoin, Litecoin, StockTwits & 2015-2018 & \acrshort{ochlv}, technical indicators, sentiment analysis & - & 30min, 4h, 1d & \acrshort{cnn}, \acrshort{lstm}, State Frequency Model & \acrshort{mse} & Keras, Tensorflow \\ \hline
\cite{McNally_2018} & Bitcoin & 2013-2016 & Price data & 100d & 30d & Bayesian optimized \acrshort{rnn}, \acrshort{lstm} & Sensitivity, specificity, precision, accuracy, \acrshort{rmse} & Keras, Python, Hyperas \\ \hline

\label{table:cryptocurrency_price_prediction}
\end{longtable}
\endgroup

%% file: tables/table7xx_1.tex
\begingroup
\footnotesize
\fontsize{7}{9}\selectfont
\begin{longtable}{
                p{0.03\linewidth}
                p{0.15\linewidth}
                p{0.08\linewidth}
                p{0.13\linewidth}
                p{0.05\linewidth}
                p{0.05\linewidth}
                p{0.10\linewidth}
                p{0.13\linewidth}
                p{0.08\linewidth}
                }

\caption{Trend Forecasting Using Only Raw Time Series Data}\\

\\
\hline

\textbf{Art.}                                                          & 
\textbf{Data Set}                                                       & 
\textbf{Period}                                                    & 
\textbf{Feature Set}                                                    & 
\textbf{Lag}                                                &
\textbf{Horizon}                                            &
\textbf{Method}                                                 & 
\textbf{Performance Criteria}                                           & 
\textbf{Env.}                                                      \\ 
\hline
\endhead


\cite{Das_2018_a} & \acrshort{sp500} stock indexes & 1963-2016 & Price data & 30d & 1d & \acrshort{nn} & Accuracy, precision, recall, F1-score, \acrshort{auroc} & R, H2o, Python, Tensorflow \\ \hline
\cite{Navon_2017} & \acrshort{spy} \acrshort{etf}, 10 stocks from \acrshort{sp500} & 2014-2016 & Price data & 60min & 30min & \acrshort{fnn} & Cumulative gain & MatConvNet, Matlab \\ \hline
\cite{Yang_2017} & Shanghai composite index and \acrshort{szse} & 1990-2016 & \acrshort{ochlv} & 20d & 1d & Ensembles of \acrshort{ann} & Accuracy & - \\ \hline
\cite{Saad_1998} & 10 stocks from \acrshort{sp500} & - & Price data &  &  & \acrshort{tdnn}, \acrshort{rnn}, \acrshort{pnn} & Missed opportunities, false alarms ratio & - \\ \hline
\cite{persio_2017} & GOOGL stock daily price data & 2012-2016 & Time window of 30 days of \acrshort{ochlv} & 22d, 50d, 70d & * & \acrshort{lstm}, \acrshort{gru}, \acrshort{rnn} & Accuracy, Logloss & Python, Keras \\ \hline
\cite{Hansson_2017} & \acrshort{sp500}, Bovespa50, \acrshort{omx}30 & 2009-2017 & Autoregressive part of the price data & 30d & 1..15d & \acrshort{lstm} & \acrshort{mse}, Accuracy & Tensorflow, Keras, R \\ \hline
\cite{Shen_2018} & \acrshort{hsi}, \acrshort{dax}, \acrshort{sp500} & 1991-2017 & Price data & - & 1d & \acrshort{gru}, \acrshort{gru}-\acrshort{svm} & Daily return \% & Python, Tensorflow \\ \hline
\cite{Chen_2016_d} & Taiwan Stock Index Futures & 2001-2015 & \acrshort{ochlv} & 240d & 1..2d & \acrshort{cnn} with \acrshort{gaf}, \acrshort{mam}, Candlestick & Accuracy & Matlab \\ \hline
\cite{Sezer_2019} & \acrshort{etf} and Dow30 & 1997-2007 & Price data &  &  & \acrshort{cnn} with feature imaging & Annualized return & Keras, Tensorflow \\ \hline
\cite{Zhou_2019} & \acrshort{ssec}, \acrshort{nasdaq}, \acrshort{sp500} & 2007-2016 & Price data & 20min & 7min & \acrshort{emd2fnn} & \acrshort{mae}, \acrshort{rmse}, \acrshort{mape} & - \\ \hline
\cite{Ausmees_2017} & 23 cap stocks from the \acrshort{omx}30 index  in Nasdaq Stockholm & 2000-2017 & Price data and returns & 30d & * & \acrshort{dbn} & \acrshort{mae} & Python, Theano \\ \hline

\label{table:trend_forecasting_1}
\end{longtable}
\endgroup

%% file: tables/table7xx_2.tex
\begingroup
\footnotesize
\fontsize{7}{9}\selectfont
\begin{longtable}{
                p{0.03\linewidth}
                p{0.15\linewidth}
                p{0.08\linewidth}
                p{0.13\linewidth}
                p{0.05\linewidth}
                p{0.05\linewidth}
                p{0.10\linewidth}
                p{0.13\linewidth}
                p{0.08\linewidth}
                }

\caption{Trend Forecasting Using Technical Indicators \& Price Data \& Fundamental Data}\\

\\
\hline

\textbf{Art.}                                                          & 
\textbf{Data Set}                                                       & 
\textbf{Period}                                                    & 
\textbf{Feature Set}                                                    & 
\textbf{Lag}                                                &
\textbf{Horizon}                                            &
\textbf{Method}                                                 & 
\textbf{Performance Criteria}                                           & 
\textbf{Env.}                                                      \\ 
\hline
\endhead


\cite{Raza_2017} & \acrshort{kse}100 index & - & Price data, several fundamental data & - & - & \acrshort{ann}, \acrshort{slp}, \acrshort{mlp}, \acrshort{rbf}, \acrshort{dbn}, \acrshort{svm} & Accuracy & - \\ \hline
\cite{Sezer_2017} & Stocks in Dow30 & 1997-2017 & \acrshort{rsi} (Technical Indicators) & 200d & 1d & \acrshort{dmlp} with genetic algorithm & Annualized return & Spark MLlib, Java \\ \hline
\cite{Liang_2017} & \acrshort{sse} Composite Index, \acrshort{ftse}100, PingAnBank & 1999-2016 & Technical indicators, \acrshort{ochlv} price & 24d & 1d & \acrshort{rbm} & Accuracy & - \\ \hline
\cite{Troiano_2018} & Dow30 stocks & 2012-2016 & Price data, several technical indicators & 40d & - & \acrshort{lstm} & Accuracy & Python, Keras, Tensorflow, \acrshort{talib} \\ \hline
\cite{Nelson_2017} & Stock price from \acrshort{ibovespa} index & 2008-2015 & Technical indicators, \acrshort{ochlv} of price & - & 15min & \acrshort{lstm} & Accuracy, Precision, Recall, F1-score, \% return, Maximum drawdown & Keras \\ \hline
\cite{song_2018} & 20 stocks  from \acrshort{nasdaq} and \acrshort{nyse} & 2010-2017 & Price data, technical indicators & 5d & 1d & \acrshort{lstm}, \acrshort{gru}, \acrshort{svm}, \acrshort{xgboost} & Accuracy & Keras, Tensorflow, Python \\ \hline
\cite{Gudelek_2017} & 17 \acrshort{etf} & 2000-2016 & Price data, technical indicators & 28d & 1d & \acrshort{cnn} & Accuracy, \acrshort{mse}, Profit, \acrshort{auroc} & Keras, Tensorflow \\ \hline
\cite{Sezer_2018} & Stocks in Dow30 and 9 Top Volume \acrshort{etf} & 1997-2017 & Price data, technical indicators & 20d & 1d & \acrshort{cnn} with feature imaging & Recall, precision, F1-score, annualized return & Python, Keras, Tensorflow, Java \\ \hline
\cite{Gunduz_2017} & Borsa Istanbul 100 Stocks & 2011-2015 & 75 technical indicators, \acrshort{ochlv} of price & - & 1h & \acrshort{cnn} & Accuracy & Keras \\ \hline


\label{table:trend_forecasting_2}
\end{longtable}
\endgroup

%% file: tables/table7xx_3.tex
\begingroup
\footnotesize
\fontsize{7}{9}\selectfont
\begin{longtable}{
                p{0.03\linewidth}
                p{0.15\linewidth}
                p{0.08\linewidth}
                p{0.13\linewidth}
                p{0.05\linewidth}
                p{0.05\linewidth}
                p{0.10\linewidth}
                p{0.13\linewidth}
                p{0.08\linewidth}
                }

\caption{Trend Forecasting Using Text Mining Techniques}\\

\\
\hline

\textbf{Art.}                                                          & 
\textbf{Data Set}                                                       & 
\textbf{Period}                                                    & 
\textbf{Feature Set}                                                    & 
\textbf{Lag}                                                &
\textbf{Horizon}                                            &
\textbf{Method}                                                 & 
\textbf{Performance Criteria}                                           & 
\textbf{Env.}                                                      \\ 
\hline
\endhead


\cite{Huang_2016} & \acrshort{sp500}, \acrshort{nyse} Composite, \acrshort{djia}, \acrshort{nasdaq} Composite & 2009-2011 & Twitter moods, index data & 7d & 1d & \acrshort{dnn}, \acrshort{cnn} & Error rate & Keras, Theano \\ \hline
\cite{Peng_2016} & News from Reuters and Bloomberg, Historical stock security data & 2006-2013 & News, price data & 5d & 1d & \acrshort{dnn} & Accuracy & - \\ \hline
\cite{Huynh_2017} & News from Reuters, Bloomberg & 2006-2013 & Financial news, price data & - & 1d, 2d, 5d, 7d & \acrshort{bi-gru} & Accuracy & Python, Keras \\ \hline
\cite{Dang_2018} & News about Apple, Airbus, Amazon from Reuters, Bloomberg, \acrshort{sp500} stock prices & 2006-2013 & Price data, news, technical indicators & - & - & Two-stream \acrshort{gru}, stock2vec & Accuracy, precision, \acrshort{auroc} & Keras, Python \\ \hline
\cite{Verma_2017} & \acrshort{nifty}50 Index, \acrshort{nifty} Bank/Auto/IT/Energy Index, News & 2013-2017 & Index data, news & 1d, 2d, 5d & 1d & \acrshort{lstm} & \acrshort{mcc}, Accuracy & - \\ \hline
\cite{Pinheiro_2017} & News from Reuters, Bloomberg, stock price/index data from \acrshort{sp500} & 2006-2013 & News and sentences & - & 1h, 1d & \acrshort{lstm} & Accuracy & - \\ \hline
\cite{Prosky_2017} & 30 \acrshort{djia} stocks, \acrshort{sp500}, \acrshort{dji}, news from Reuters & 2002-2016 & Price data and features from news articles & 1m & 1d & \acrshort{lstm}, \acrshort{nn}, \acrshort{cnn} and word2vec & Accuracy & VADER \\ \hline
\cite{Liu_2018} & APPL from \acrshort{sp500} and news from Reuters & 2011-2017 & News, \acrshort{ochlv}, Technical indicators & - & 1d & \acrshort{cnn} + \acrshort{lstm}, \acrshort{cnn}+\acrshort{svm} & Accuracy, F1-score & Tensorflow \\ \hline
\cite{Yoshihara_2014} & News, Nikkei Stock Average and 10-Nikkei companies & 1999-2008 & News, \acrshort{macd} & - & 1d & \acrshort{rnn}, \acrshort{rbm}+\acrshort{dbn} & Accuracy, P-value & - \\ \hline
\cite{Shi_2018} & News from Reuters and Bloomberg for \acrshort{sp500} stocks & 2006-2015 & Financial news, price data & 1d & 1d & DeepClue & Accuracy & Dynet software \\ \hline
\cite{Zhang_2018} & Price data, index data, news, social media data & 2015 & Price data, news from articles and social media & 1d & 1d & Coupled matrix and tensor & Accuracy, \acrshort{mcc} & Jieba \\ \hline
\cite{Hu_2018} & News and Chinese stock data & 2014-2017 & Selected words in a news & 10d & 1d & \acrshort{han} & Accuracy, Annual return & - \\ \hline
\cite{Wang_2018_a} & Sina Weibo, Stock market records & 2012-2015 & Technical indicators, sentences & - & - & DRSE & F1-score, precision, recall, accuracy, \acrshort{auroc} & Python \\ \hline
\cite{MATSUBARA_2018} & Nikkei225, \acrshort{sp500}, news from Reuters and Bloomberg & 2001-2013 & Price data and news & 1d & 1d & \acrshort{dgm} & Accuracy, \acrshort{mcc}, \%profit & - \\ \hline
\cite{Li_2018} & News, stock prices from Hong Kong Stock Exchange & 2001 & Price data and \acrshort{tfidf} from news & 60min & (1..6)*5min & \acrshort{elm}, \acrshort{dlr}, \acrshort{pca}, \acrshort{belm}, \acrshort{kelm}, \acrshort{nn} & Accuracy & Matlab \\ \hline

\label{table:trend_forecasting_3}
\end{longtable}
\endgroup

%% file: tables/table7xx_4.tex
\begingroup
\footnotesize
\fontsize{7}{9}\selectfont
\begin{longtable}{
                p{0.03\linewidth}
                p{0.15\linewidth}
                p{0.08\linewidth}
                p{0.13\linewidth}
                p{0.05\linewidth}
                p{0.05\linewidth}
                p{0.10\linewidth}
                p{0.13\linewidth}
                p{0.08\linewidth}
                }

\caption{Trend Forecasting Using Various Data}\\

\\
\hline

\textbf{Art.}                  & 
\textbf{Data Set}              & 
\textbf{Period}                & 
\textbf{Feature Set}           & 
\textbf{Lag}                   &
\textbf{Horizon}               &
\textbf{Method}                & 
\textbf{Performance Criteria}  & 
\textbf{Env.}                  \\ 
\hline
\endhead


\cite{Tsantekidis_2017} & Nasdaq Nordic (Kesko Oyj, Outokumpu Oyj, Sampo, Rautaruukki, Wartsila Oyj) & 2010 & Price and volume data in  \acrshort{lob} & 100s & 10s, 20s, 50s & \acrshort{lstm} & Precision, Recall, F1-score, Cohen's k & - \\ \hline
\cite{Sirignano_2018} & High-frequency record of all orders & 2014-2017 & Price data, record of all orders, transactions & 2h & - & \acrshort{lstm} & Accuracy & - \\ \hline
\cite{Chen_2018_e} & Chinese, The Shanghai-Shenzhen 300 Stock Index (\acrshort{hs}300 & 2015-2017 & Social media news (Sina Weibo), price data & 1d & 1d & \acrshort{rnn}-Boost with \acrshort{lda} & Accuracy, \acrshort{mae}, \acrshort{mape}, \acrshort{rmse} & Python, Scikit learn \\ \hline
\cite{Buczkowski_2017} & ISMIS 2017 Data Mining Competition dataset & - & Expert identifier, class predicted by expert & - & - & \acrshort{lstm} + \acrshort{gru} + \acrshort{fcnn} & Accuracy & - \\ \hline
\cite{Tsantekidis_2017_a} & Nasdaq Nordic (Kesko Oyj, Outokumpu Oyj, Sampo, Rautaruukki, Wartsila Oyj) & 2010 & Price, Volume data, 10 orders of the \acrshort{lob} & - & - & \acrshort{cnn} & Precision, Recall, F1-score, Cohen's k & Theano, Scikit learn, Python \\ \hline
\cite{Doering_2017} & London Stock Exchange & 2007-2008 & Limit order book state, trades, buy/sell orders, order deletions & - & - & \acrshort{cnn} & Accuracy, kappa & Caffe \\ \hline

\label{table:trend_forecasting_4}
\end{longtable}
\endgroup

%% file: 6_Snapshot_of_Field.tex
\section{Current Snaphot of The Field}
\label{sec:snapshot}

\begin{figure*}[!htb]
\centering
\includegraphics[width=4.5in]{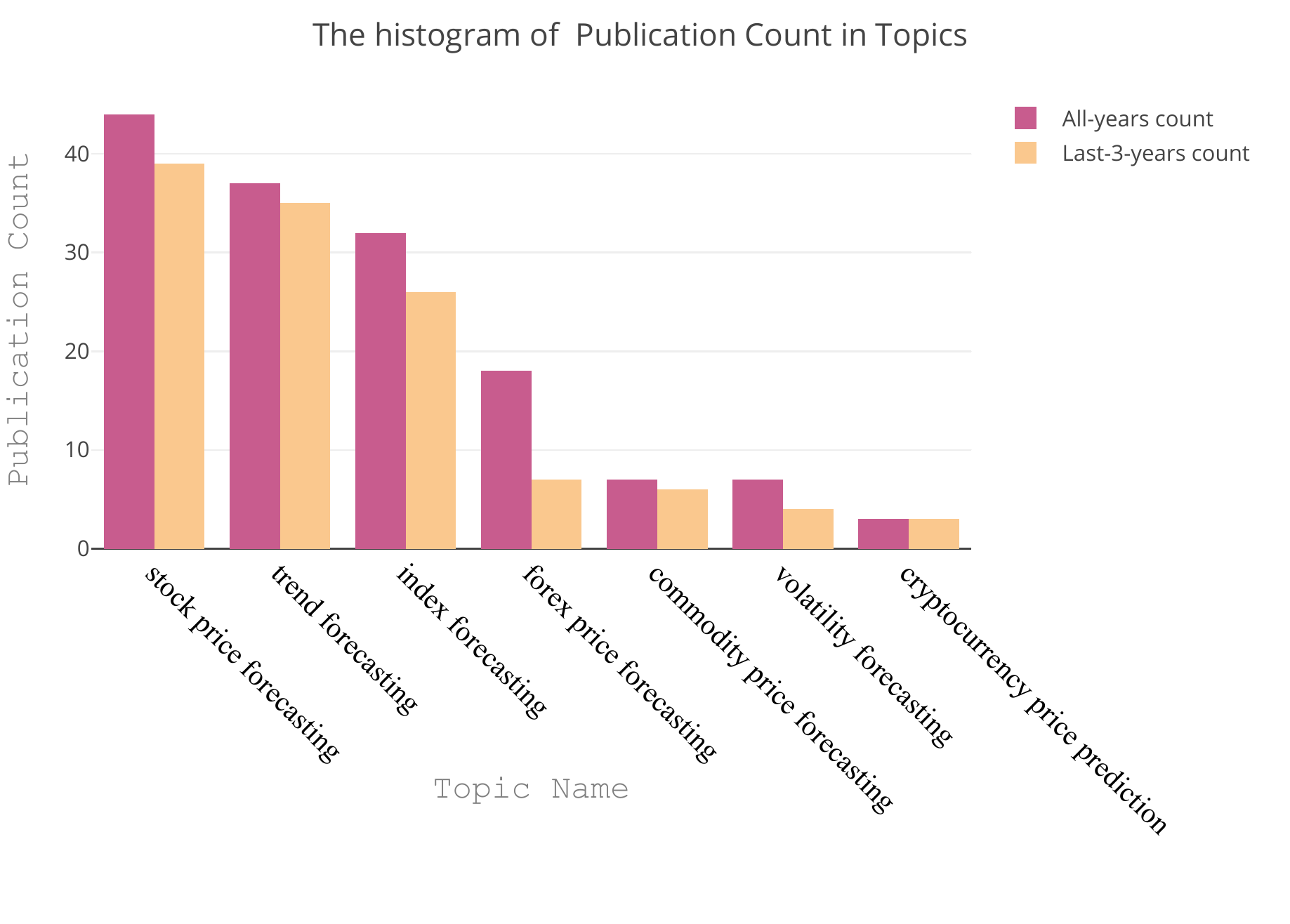}
\caption{The histogram of  Publication Count in Topics}
\label{fig:histogram_of_topic}
\end{figure*}

After reviewing through all the research papers specifically targeted for financial time series forecasting implementations using \gls{dl} models, we are now ready to provide some overall statistics about the current state of the studies. The number of papers that we were able to locate to be included in our survey was 140. We categorized the papers according to their forecasted asset type. Furthermore, we also analyzed the studies through their \gls{dl} model choices, frameworks for the development environment, data sets, comparable benchmarks, and some other differentiating criteria like feature sets, number of citations, etc. which we were not able to include in the paper due to space constraints. We will now summarize our notable observations to provide important highlights for the interested researchers within the field.  

\begin{figure*}[!htb]
\centering
\includegraphics[width=4.5in]{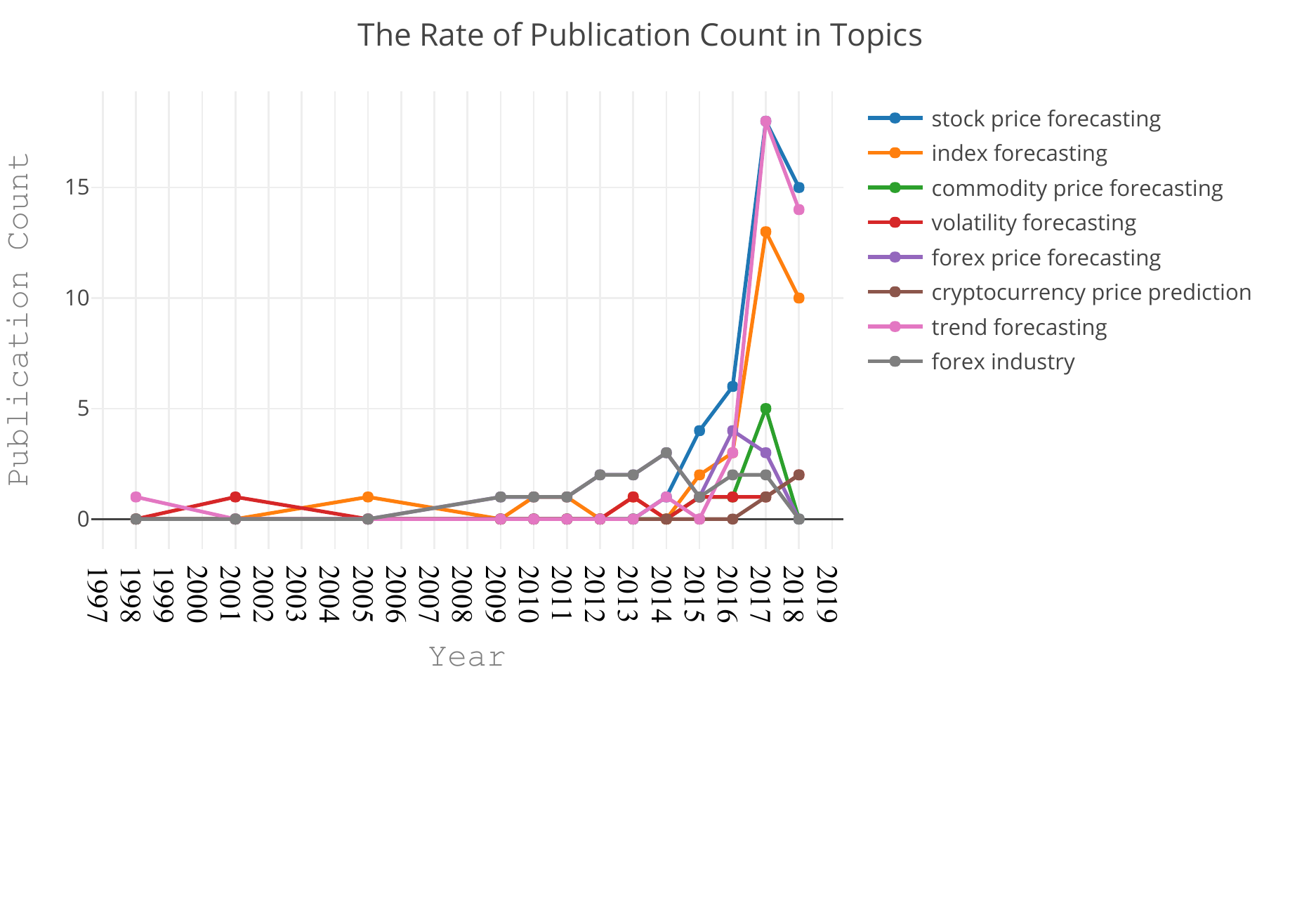}
\caption{The rate of  Publication Count in Topics}
\label{fig:scatter_topic}
\end{figure*}

\begin{figure*}[!htb]
\centering
\includegraphics[width=4.5in]{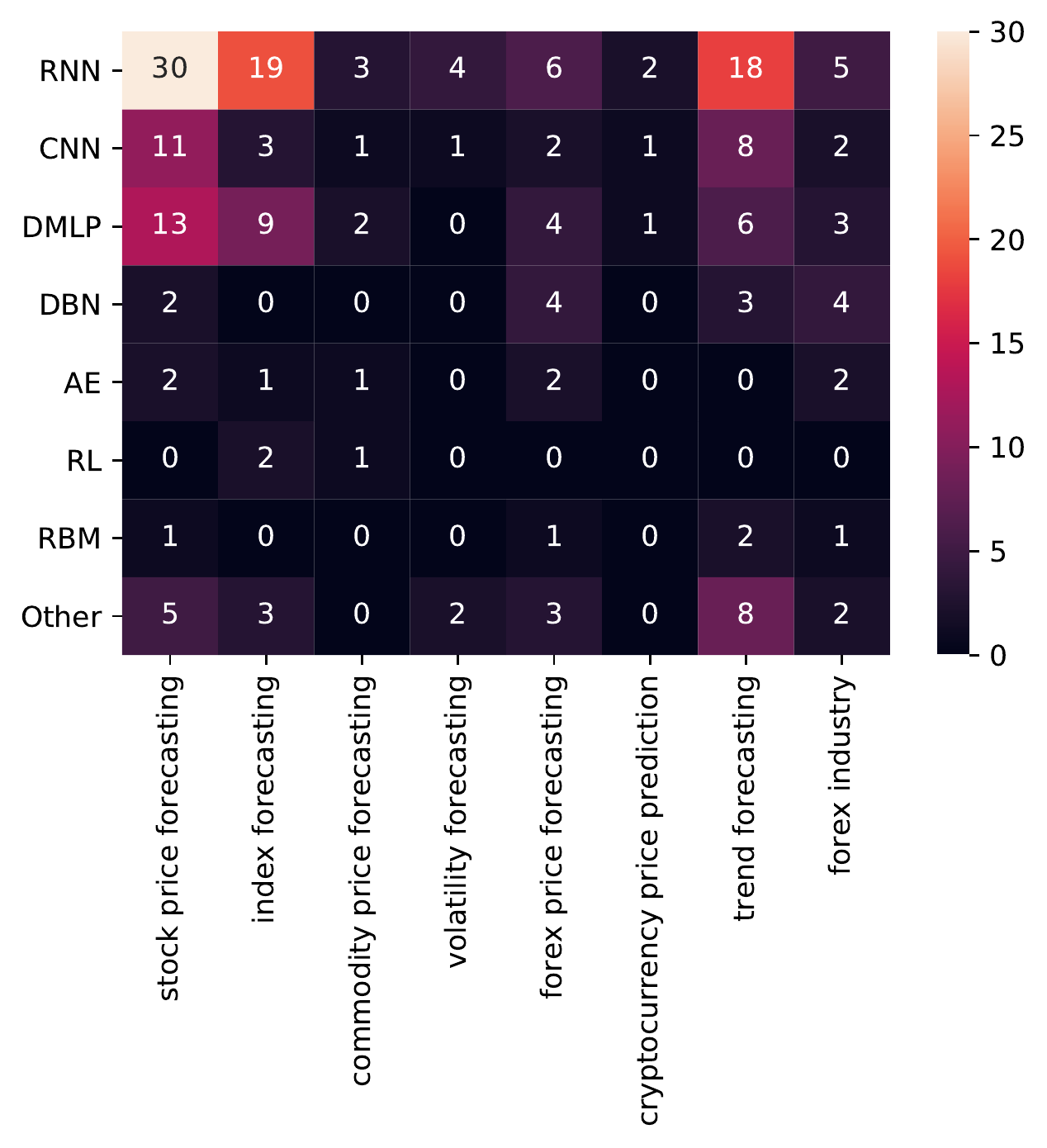}
\caption{Topic-Model Heatmap}
\label{fig:topic_method_heatmap}
\end{figure*}

\begin{figure*}[!htb]
\centering
\includegraphics[width=4.5in]{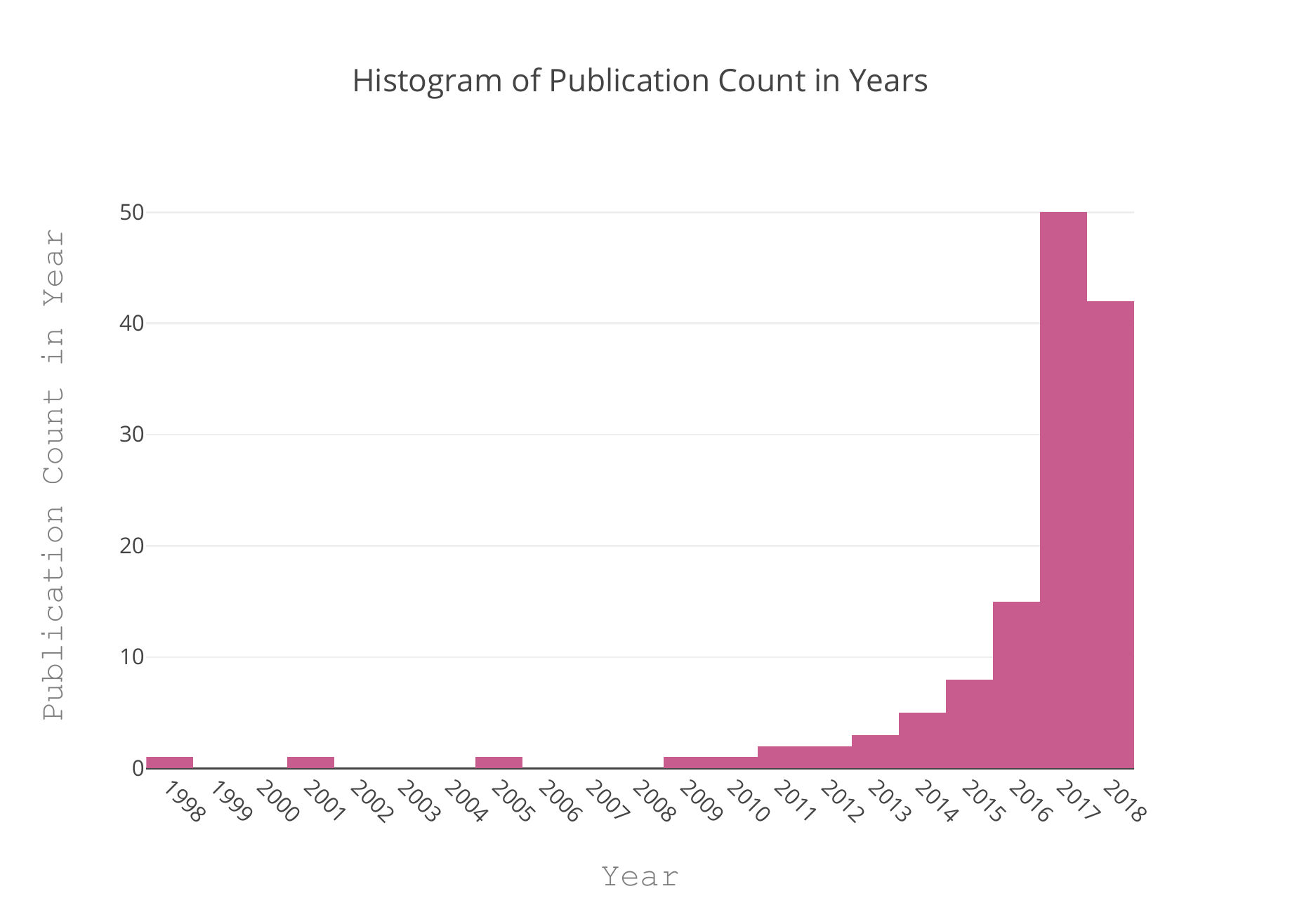}
\caption{The histogram of  Publication Count in Years}
\label{fig:histogram_of_year}
\end{figure*}

Figure \ref{fig:histogram_of_topic} presents the various asset types that the researchers decided to develop their corresponding forecasting models for. As expected, stock market-related prediction studies dominate the field. Stock price forecasting, trend forecasting and index forecasting were the top three picks for the financial time series forecasting research. So far, 46 papers were published for stock price forecasting, 38 for trend forecasting and 33 for index forecasting, respectively. These studies constitute more than 70\% of all studies indicating high interest. Following those include 19 papers for forex prediction and 7 papers for volatility forecasting. Meanwhile cryptocurrency forecasting has started attracting researchers, however, there were just 3 papers published yet, but this number is expected to increase in coming years \cite{Fischer_2019}.   Figure  \ref{fig:scatter_topic}  highlights the rate of publication counts for various implementation areas throughout the years.  Meanwhile Figure  \ref{fig:topic_method_heatmap}  provides more details about the choice of DL models over various implementation areas.

\begin{figure*}[!htb]
\centering
\includegraphics[width=4.5in]{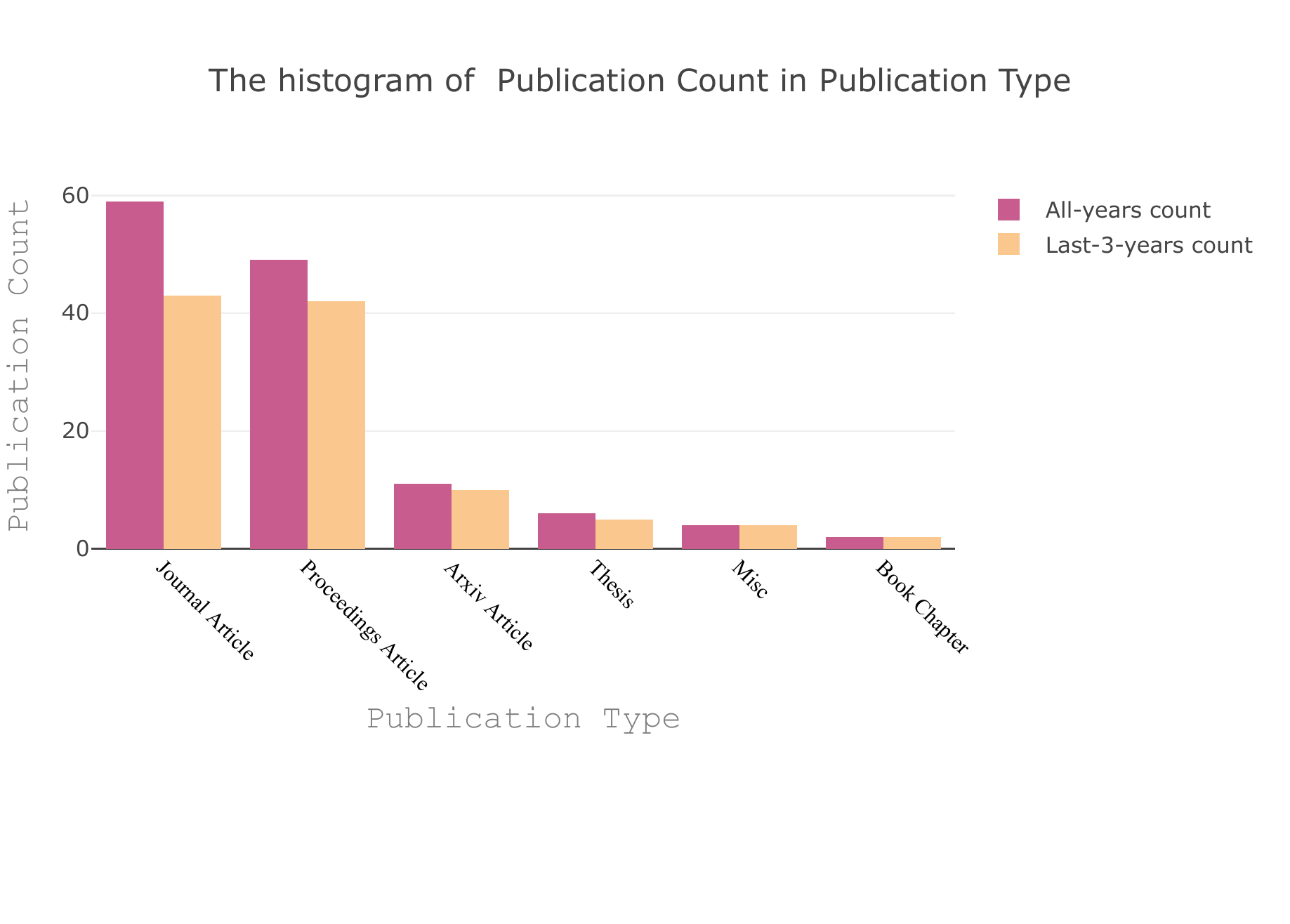}
\caption{The histogram of  Publication Count in Publication Types}
\label{fig:histogram_of_pub_type}
\end{figure*}

Figure \ref{fig:histogram_of_year} illustrates the accelerating appetite in the last 3 years by researchers for developing \gls{dl} models for the financial time series implementations. Meanwhile, as Figure \ref{fig:histogram_of_pub_type} indicates, most of the studies were published in journals (57 of them) and conferences (49 papers) even though a considerable amount of arXiv papers (11) and graduate theses (6) also exist.

\begin{figure*}[!htb]
\centering
\includegraphics[width=\linewidth]{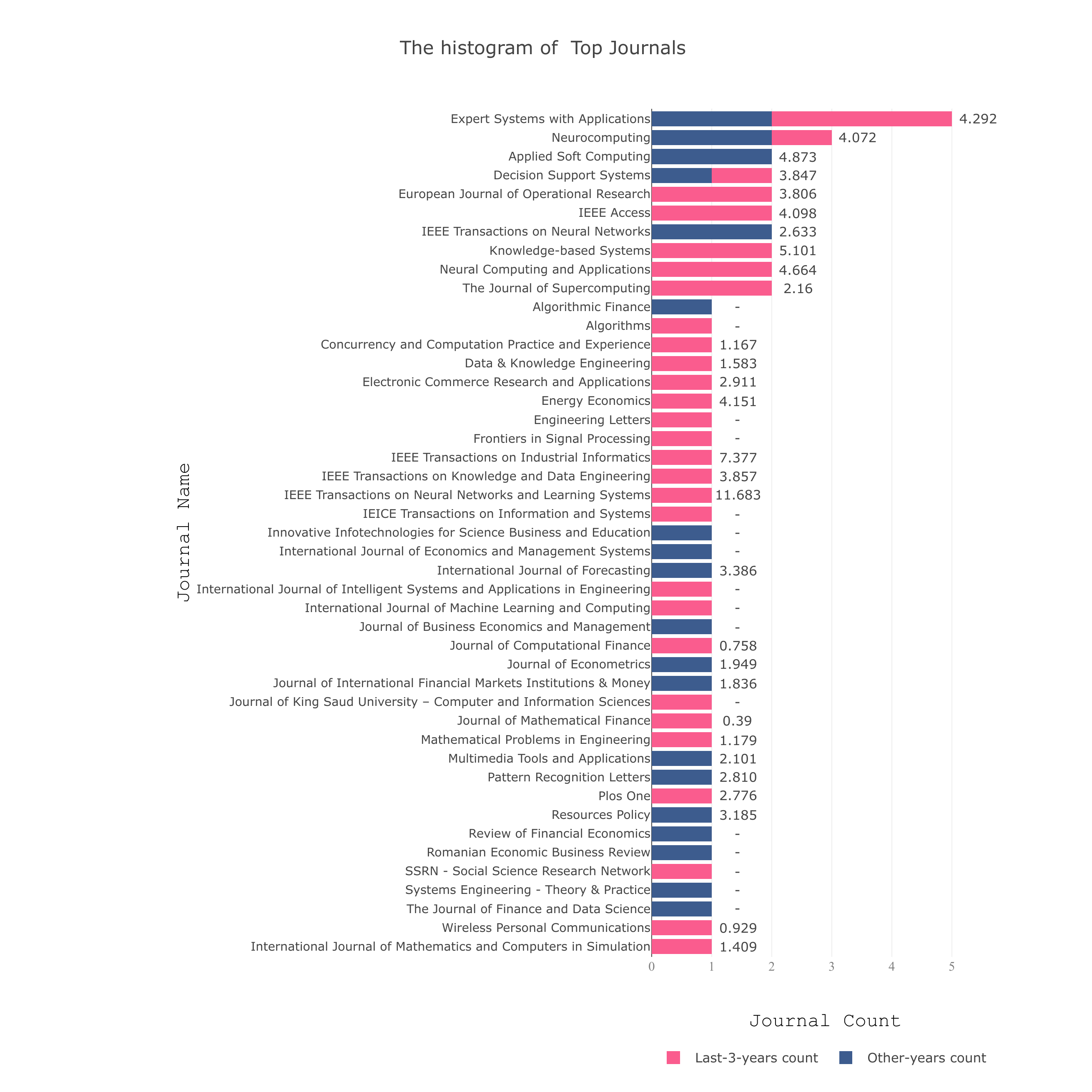}
\caption{Top Journals - corresponding numbers next to the bar graph are representing the impact factor of the journals}
\label{fig:top_journals}
\end{figure*}

One of the most important questions for a researcher is where he/she can publish their research findings. During our review of the papers, we also carefully investigated where each paper was published. We tabulated our results for the top journals for financial time series forecasting in Fig~\ref{fig:top_journals}. According to these results, the journals with the most published papers include Expert Systems with Applications, Neurocomputing, Applied Soft Computing, The Journal of Supercomputing, Decision Support Systems, Knowledge-based Systems, European Journal of Operational Research and IEEE Access. The interested researchers should also consider the trend within the last 3 years, as tendencies can be slightly varying depending on the particular implementation areas.

\begin{figure*}[!htb]
\centering
\includegraphics[width=4.5in]{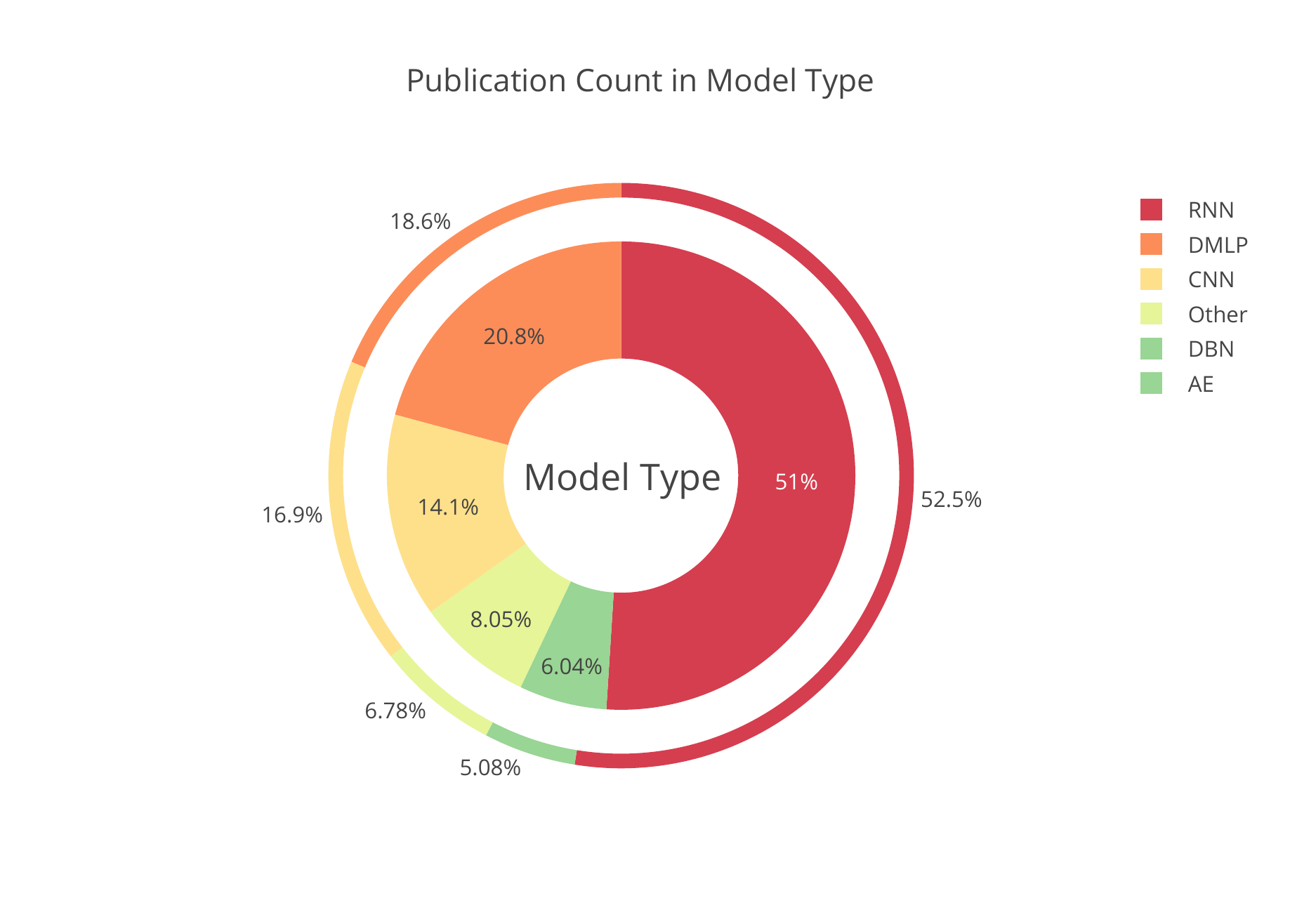}
\caption{The Piechart of  Publication Count in Model Types}
\label{fig:piechart_model_type}
\end{figure*}

\begin{figure*}[!htb]
\centering
\includegraphics[width=4.5in]{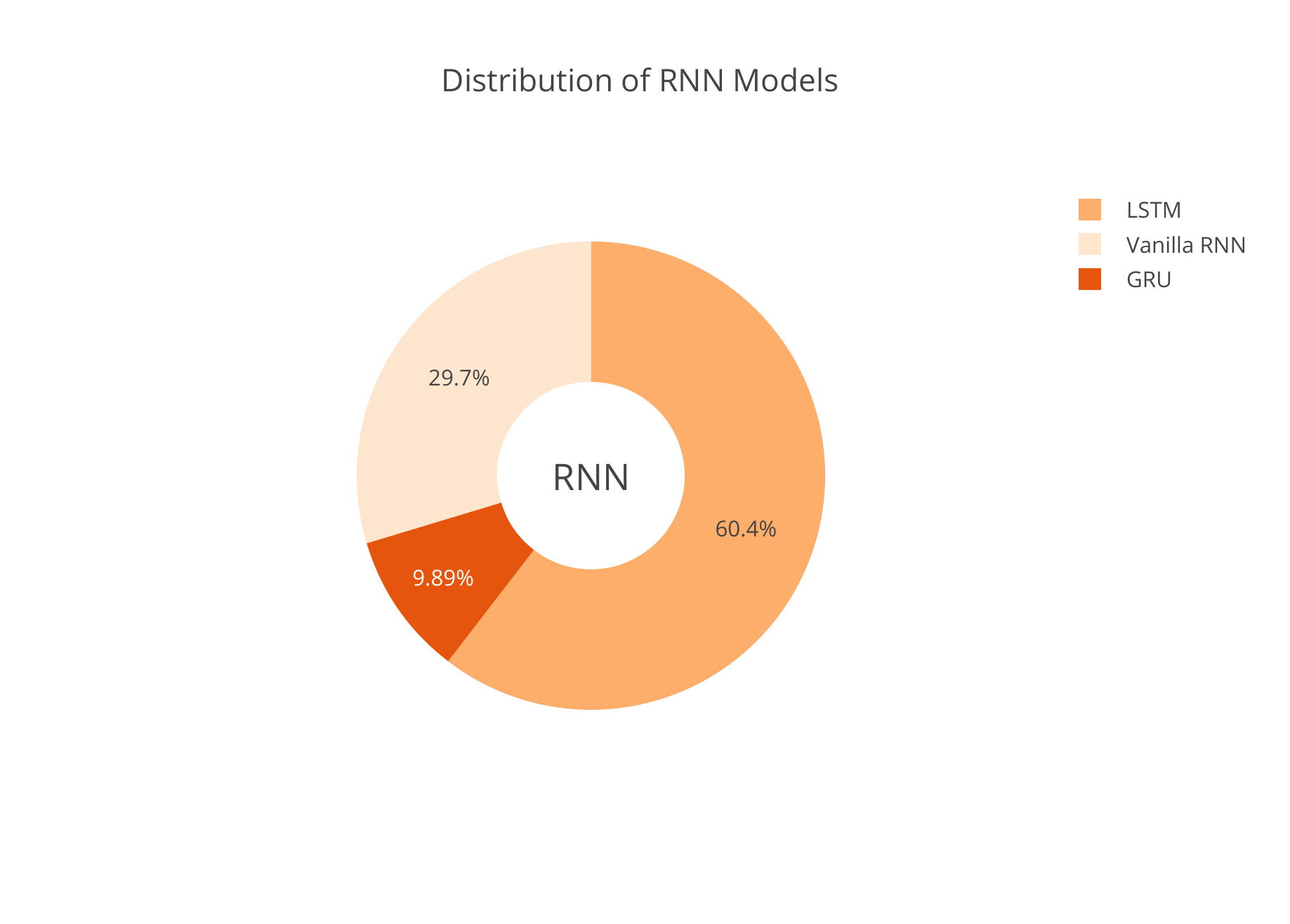}
\caption{Distribution of RNN Models}
\label{fig:piechart_wrt_rnn_model_type}
\end{figure*}

Carefully analyzing Figure \ref{fig:piechart_model_type} clearly validates the dominance of \gls{rnn} based models (65 papers) among all others for \gls{dl} model choices, followed by \gls{dmlp} (23 papers) and \gls{cnn} (20 papers). The inner-circle represents all years considered, meanwhile the outer circle just provides the studies within the last 3 years.  We should note that \gls{rnn} is a general model with several versions including \gls{lstm}, \gls{gru}, etc.  Within \gls{rnn}, the researchers mostly prefer \gls{lstm} due to its relative easiness of model development phase, however, other types of \gls{rnn} are also common. Figure \ref{fig:piechart_wrt_rnn_model_type} provides a snapshot of the \gls{rnn} model distribution. As mentioned above, \gls{lstm} had the highest interest among all with 58 papers, while Vanilla \gls{rnn} and \gls{gru} had 27 and 10 papers respectively. Hence, it is clear that \gls{lstm} was the most popular \gls{dl} model for financial time series forecasting or regression studies. 

Meanwhile, \gls{dmlp} and \gls{cnn} generally were preferred for classification problems. Since the time series data generally consists of temporal components, some data preprocessing might be required before the actual classification can occur. Hence, a lot of these implementations utilize feature extraction, selection techniques along with possible dimensionality reduction methods. A lot of researchers decided to use \gls{dmlp} mostly due to the fact that its shallow version \gls{mlp} has been used extensively before and has a proven successful track record for many different financial applications including financial time series forecasting. Consistent with our observations, \gls{dmlp} was also mostly preferred in the stock, index or in particular trend forecasting, since it is by definition, a classification problem with two (uptrend or downtrend) and three (uptrend, stationary or downtrend) class instances.

In addition to \gls{dmlp}, \gls{cnn} was also a popular choice for classification type financial time series forecasting implementations. Most of these studies appeared within the last 3 years. As mentioned before, in order to convert the temporal time-varying sequential data into a more stationary classifiable form, some preprocessing might be necessary. Even though some 1-D representations exist, the 2-D implementation for \gls{cnn} was more common, mostly inherited through image recognition applications of \gls{cnn} from computer vision implementations. In some studies \cite{Chen_2016_d, Sezer_2019, Sezer_2017, Sezer_2018, Tsantekidis_2017_a}, innovative transformations of financial time series data into an image-like representation has been adapted and impressive performance results have been achieved. As a result, \gls{cnn} might increase its share of interest for financial time series forecasting in the next few years.

\begin{figure*}[!htb]
\centering
\includegraphics[width=4.5in]{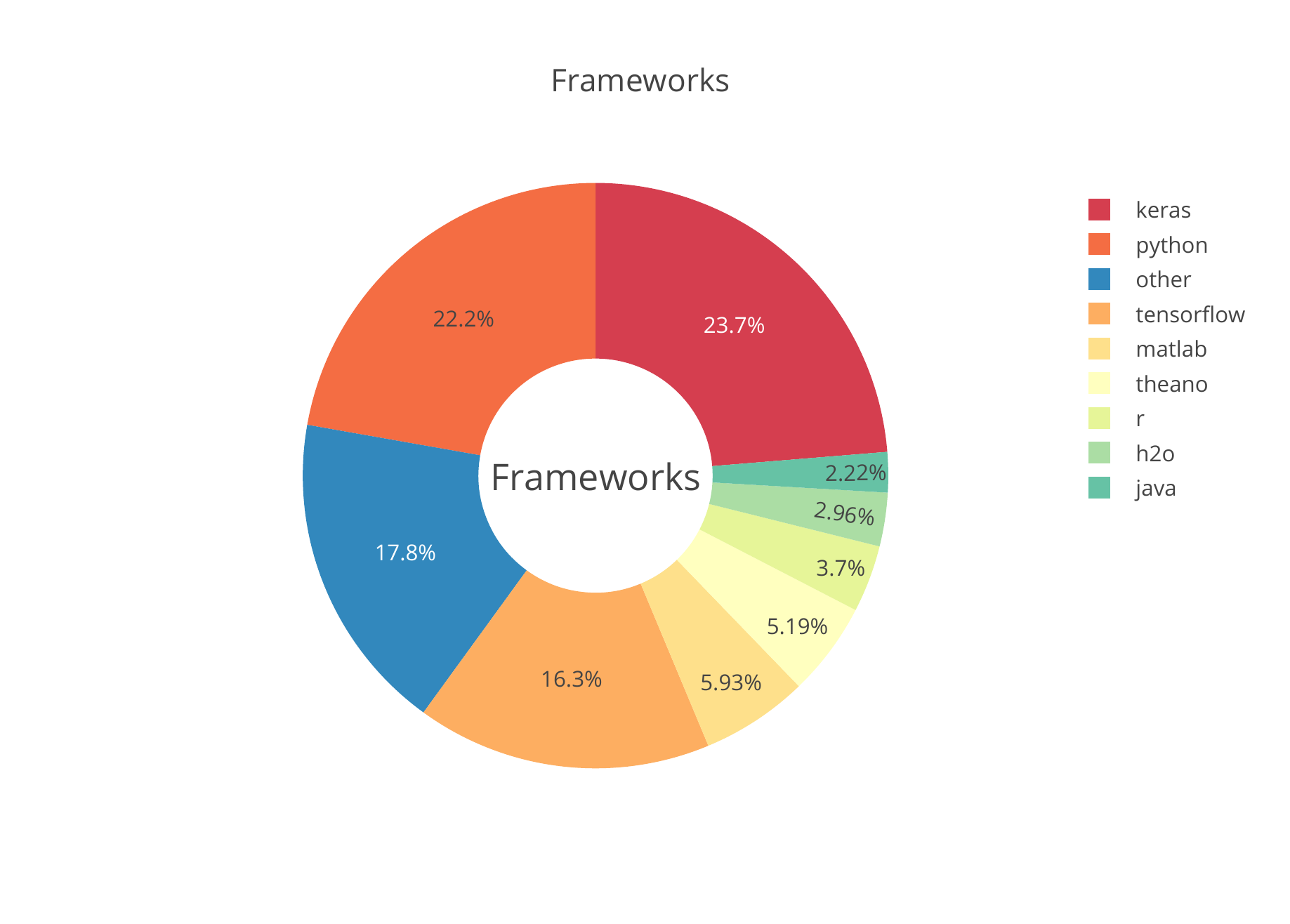}
\caption{The Preferred Development Environments}
\label{fig:platform_piechart}
\end{figure*}

As one final note, Figure  \ref{fig:platform_piechart} shows which frameworks and platforms the researchers and developers used while implementing their work. We tried to extract this information from the papers to the best of our effort. However, we need to keep in mind that not every publication provided their development environment. Also in most of the papers, generally, the details were not given preventing us from a more thorough comparison chart, i.e. some researchers claimed they used Python, but no further information was given, while some others mentioned the use of Keras or TensorFlow providing more details.  Also, within the ``Other" section the usage of Pytorch is on the rise in the last year or so, even though it is not visible from the chart.  Regardless, Python-related tools were the most influential technologies behind the implementations covered in this survey.


%% file: 7_Discussion.tex
\section{Discussion and Open Issues}
\label{sec:discussion}

From an application perspective, even though financial time series forecasting has a relatively narrow focus, i.e. the implementations were mainly based on price or trend prediction, depending on the underlying \gls{dl} model, very different and versatile models exist in literature. We need to keep in mind that, even though financial time series forecasting is a subset of time-series studies, due to the embedded profit-making expectations through successful prediction models, some differences exist, such that higher prediction accuracy sometimes might not reflect a profitable model. Hence, the risk and reward structure must also be taken into consideration. At this point, we will try to elaborate on our observations about these differences in various model designs and implementations. 

\subsection{DL Models for financial time series forecasting}

According to the publication statistics, \gls{lstm} was the preferred choice of most researchers for financial time series forecasting. \gls{lstm} and its variations utilized the time-varying data with feedback embedded representations, resulting in higher performances for time series prediction implementations. Since most of the financial data, one way or another, included time-dependent components, \gls{lstm} was the natural choice in financial time series forecasting problems. Meanwhile, \gls{lstm} is a special \gls{dl} model deriven from a more general classifier family, namely \gls{rnn}.

Careful analysis of Figure \ref{fig:piechart_model_type} illustrates the dominance of \gls{rnn} (which is highly consisted of \gls{lstm}). As a matter of fact, more than half of the published papers for time series forecasting studies fall into the \gls{rnn} model category. Regardless of its problem type, price or trend prediction, the ordinal nature of the data representation forced the researchers to consider \gls{rnn}, \gls{gru} and \gls{lstm} as viable preferences for their model choices. Hence, \gls{rnn} models were chosen, at least for benchmarking, in a lot of studies for performance comparison against other developed models.

Meanwhile, other models were also used for time series forecasting problems. Among those, \gls{dmlp} had the most interest due to the market dominance of its shallow cousin, \gls{mlp} and its wide acceptance and long history within \gls{ml} society. However, there is a fundamental difference in how \gls{dmlp} and \gls{rnn} based models were used for financial time series prediction problems.

\gls{dmlp} fits well for both regression and classification problems. However, in general, data order independence must be preserved for better utilizing the internal working dynamics of such networks, even though through the learning algorithm configuration, some adjustments can be performed. In most cases, either trend components of the data need to be removed from the underlying time series, or some data transformations might be needed so that the resulting data becomes stationary. Regardless, some careful preprocessing might be necessary for the \gls{dmlp} model to be successful. In contrast, \gls{rnn} based models can directly work with time-varying data, making it easier for researchers to develop \gls{dl} models.

As a result, most of the \gls{dmlp} implementations had embedded data preprocessing before the learning stage. However, this inconvenience did not prevent the researchers to use \gls{dmlp} and its variations during their model development process. Instead, a lot of versatile data representations were attempted in order to achieve higher overall prediction performances. A combination of fundamental and/or technical analysis parameters along with other features like financial sentiment through text mining was embedded into such models. In most of the \gls{dmlp} studies, the corresponding problem was treated as classification, especially in trend prediction models, whereas \gls{rnn} based models directly predicted the next value of the time series. Both approaches had some success in beating the underlying benchmark; hence it is not possible to claim victory of one model type over the other. However, for the general rule of thumb, researchers prefer \gls{rnn} based models for time series regression and \gls{dmlp} for trend classification (or buy-sell point identification)

Another model that started becoming popular recently is \gls{cnn}. \gls{cnn} also works better for classification problems and unlike \gls{rnn} based models, it is more suitable for either non-time varying or static data representations. 
The comments for \gls{dmlp} are also mostly valid for \gls{cnn}. Furthermore, unlike \gls{dmlp}, \gls{cnn} mostly requires locality within the data representation for better-performing classification results. One particular implementation area of \gls{cnn} is image-based object recognition problems. In recent years, \gls{cnn} based models dominated this field, handily outperforming all other models. 
Meanwhile, most financial data is time-varying and it might not be easy to implement \gls{cnn} directly for financial applications. However, in some recent studies, various independent research groups followed an innovative transformation of 1-D time-varying financial data into 2-D mostly stationary image-like data so that they could utilize the power of \gls{cnn} through adaptive filtering and implicit dimensionality reduction. Hence, with that approach, they were able to come up with successful models.  

There is also a rising trend to use deep \gls{rl} based financial algorithmic trading implementations; these are mostly associated with various agent-based models where different agents interact and learn from their interactions. This field even has more opportunities to offer with advancements in financial sentiment analysis through text mining to capture investor psychology; as a result, behavioral finance can benefit from these particular studies associated with \gls{rl} based learning models coupled with agent-based studies. 

Other models including \gls{dbn}, \gls{ae} and \gls{rbm} also were used by several researchers and superior performances were reported in some of their work; but the interested readers need to check these studies case by case to see how they were modelled both from the data representation and learning point of view.

\subsection{Discussions on Selected Features}

Regardless of the underlying forecasting problem, somehow the raw time series data was almost always embedded directly or indirectly within the feature vector, which is particularly valid for \gls{rnn}-based models. However, in most of the other model types, other features were also included. Fundamental analysis and technical analysis features were among the most favorable choices for stock/index forecasting studies.

Meanwhile, in recent years, financial text mining is particularly getting more attention, mostly for extracting the investor/trader sentiment. The streaming flow of financial news, tweets, statements, blogs allowed the researchers to build better and more versatile prediction and evaluation models integrating numerical and textual data. The general methodology involves in extracting financial sentiment analysis through text mining and combining that information with fundamental/technical analysis data to achieve better overall performance. It is logical to assume that this trend will continue with the integration of more advanced text and \gls{nlp} techniques. 

\subsection{Discussions on Forecasted Asset Types}

Even though forex price forecasting is always popular among the researchers and practitioners, stock/index forecasting has always had the most interest among all asset groups. Regardless, price/trend prediction and algo-trading models were mostly embedded with these prediction studies.  

These days, one other hot area to financial time series forecasting research is involved with cryptocurrencies. Cryptocurrency price prediction has an increasing demand from the financial community. Since the topic is fairly new, we might see more studies and implementations coming in due to high expectations and promising rewards.

There were also a number of publications in commodity price forecasting research, in particular, the price of oil. Oil price prediction is crucial due to its tremendous effect on world economic activities and planning. Meanwhile, gold is considered a safe investment and almost every investor, at one time, considers allocating some portion of their portfolios for gold-related investments. In times of political uncertainties, a lot of people turn to gold for protecting their savings. Even though we have not encountered a noteworthy study for gold price forecasting, due to its historical importance, there might be opportunities in this area for the years to come.

\subsection{Open Issues and Future Work}

Despite the general motivation for financial time series forecasting remaining fairly unchanged, the means of achieving the financial goals vary depending on the choices and trade-off between the traditional techniques and newly developed models. Since our fundamental focus is on the application of \gls{dl} for financial time series studies, we will try to asses the current state of the research and extrapolate that into the future. 
 
\subsubsection{Model Choices for the Future}

The dominance of \gls{rnn}-based models for price/trend prediction will probably not disappear anytime soon, mainly due to their easy adaptation to most asset forecasting problems. Meanwhile, some enhanced versions of the original \gls{lstm} or \gls{rnn} models, generally integrated with hybrid learning systems started becoming more common. Readers need to check individual studies and assess their performances to see which one fits the best for their particular needs and domain requirements.  

We have observed the increasing interest in 2-D \gls{cnn} implementations of financial forecasting problems through converting the time series into an image-like data type. This innovative methodology seems to work quite satisfactorily and provides promising opportunities. More studies of this kind will probably continue in the near future.

Nowadays, new models are generated through older models via modifying or enhancing the existing models so that better performances can be achieved. Such topologies include \gls{gan}, Capsule networks, etc. They have been used in various non-financial studies, however, financial time series forecasting has not been investigated for those models yet. As such, there can be exciting opportunities both from research and practical point of view.

Another \gls{dl} model that is not investigated thoroughly is Graph \gls{cnn}. Graphs can be used to represent portfolios, social networks of financial communities, fundamental analysis data, etc. Even though graph algorithms can directly be applied to such configurations, different graph representations can also be implemented for the time series forecasting problems. Not much has been done on this particular topic, however, through graph representations of the time series data and implementing graph analysis algorithms, or implementing \gls{cnn} through these graphs are among the possibilities that the researchers can choose.

As a final note for the future models, we believe deep \gls{rl} and agent-based models offer great opportunities for the researchers. \gls{hft} algorithms, robo-advisory systems highly depend on automated algorithmic trading systems that can decide what to buy and when to buy without any human intervention. These aforementioned models can fit very well in such challenging environments. The rise of the machines will also lead to a technological (and algorithmic) arms race between Fintech companies and quant funds to be the best in their neverending search for ``achieving alpha". New research in these areas can be just what the doctor ordered.

\subsubsection{Future Projections for Financial Time Series Forecasting}

Most probably, for the foreseeable future, the financial time series forecasting will have a close research cooperation with the other financial application areas like algorithmic trading and portfolio management, as it was the case before. However, changes in the available data characteristics and introduction of new asset classes might not only alter the forecasting strategies of the developers, but also force the developers to look for new or alternative techniques to better adapt to these new challenging working conditions. In addition, metrics like \gls{crps} for evaluating probability distributions might be included for more thorough analysis.

One rising trend, not only for financial time series forecasting, but for all intelligent decision support systems, is the human-computer interaction and \gls{nlp} research. Within that field, text mining and financial sentiment analysis areas are of particular importance to financial time series forecasting. Behavioral finance may benefit from the new advancements in these fields. 

In order to utilize the power of text mining, researchers started developing new data representations like Stock2Vec \cite{Dang_2018} that can be useful for combining textual and numerical data for better prediction models. Furthermore, \gls{nlp} based ensemble models that integrate data semantics with time-series data might increase the accuracy of the existing models.

One area that can benefit a lot from the interconnected financial markets is the automated statistical arbitrage trading model development. It has been used in forex and commodity markets before. In addition, a lot of practitioners currently seek arbitrage opportunities in the cryptocurrency markets \cite{Fischer_2019}, due to the existence of the huge number of coins available on various marketplaces. Price disruptions, high volatility, bid-ask spread variations cause arbitrage opportunities across different platforms. Some opportunists develop software models that can track these price anomalies for the instant materialization of profits. Also, it is possible to construct pairs trading portfolios across different asset classes using appropriate models. It is possible that \gls{dl} models can learn (or predict) these opportunities faster and more efficient than classical rule-based systems. This will also benefit \gls{hft} studies that are constantly looking for faster and more efficient trading algorithms and embedded systems with minimum latency. In order to achieve that, Graphics Processing Unit (GPU) or Field Programmable Gate Array (FPGA) based hardware solutions embedded with \gls{dl} models can be utilized. There is a lack of research accomplished on this hardware aspect of financial time series forecasting and algorithmic trading. As long as there is enough computing power available, it is worth investigating the possibilities for better algorithms, since the rewards are high. 

\subsection{Responses to our Initial Research Questions}

We are now ready to go back to our initially stated research questions. Our question and answer pairs, through our observations, are as follows: 

\begin{itemize}
\item {Which DL models are used for financial time series forecasting ?}

Response: \gls{rnn} based models (in particular \gls{lstm}) are the most commonly used models. Meanwhile, \gls{cnn} and \gls{dmlp} have been used extensively in classification type implementations (like trend classification) as long as appropriate data processing is applied to the raw data. 

\item {How is the performance of \gls{dl} models compared with traditional machine learning counterparts ?}

Response: In the majority of the studies, \gls{dl} models were better than \gls{ml}. However, there were also many cases where their performances were comparable. There were even two particular studies  (\cite{Dezsi_2016,Sermpinis_2014}   where \gls{ml}  models performed better than  \gls{dl} models. Meanwhile, appetite for preferrance of DL implementations over ML models is growing. Advances in computing power, availability of big data, superior performance, implicit feature learning capabilities and user friendly model development environment for DL models are among the main reasons for this migration.

\item {What is the future direction for \gls{dl} research for financial time series forecasting ?}

Response: \gls{nlp}, semantics and text mining-based hybrid models ensembled with time-series data might be more common in the near future.

\end{itemize}

%% file: 8_Conclusion_Futurework.tex
\section{Conclusions}
\label{sec:conclusions}

Financial time series forecasting has been very popular among \gls{ml} researchers for more than 40 years. The financial community got a new boost lately with the introduction of \gls{dl} implementations for financial prediction research and a lot of new publications appeared accordingly. In our survey, we wanted to review the existing studies to provide a snapshot of the current research status of \gls{dl} implementations for financial time series forecasting. We grouped the studies according to their intended asset class along with the preferred \gls{dl} model associated with the problem. Our findings indicate, even though financial forecasting has a long research history, overall interest within the \gls{dl} community is on the rise through utilizing new \gls{dl} models; hence, a lot of opportunities exist for researchers.

\section{Acknowledgement}

This work is supported by Scientific and Technological Research Council of Turkey (TUBITAK) grant no 215E248.

